\RequirePackage{fix-cm}
\documentclass[twocolumn]{svjour3}          
\smartqed  
\usepackage{hyphenat}
\usepackage[english]{babel}
\usepackage[olditem, oldenum]{paralist} 
\usepackage{graphics, graphicx}
\usepackage{wrapfig}
\usepackage[caption=false]{subfig}
\usepackage{tabulary, tabularx, booktabs, multirow, makecell, array, threeparttable} 
  \newcolumntype{K}[1]{>{\centering\arraybackslash}p{#1}}
\usepackage{amsmath, amssymb, amstext, bm, mathtools}  
\usepackage[prependcaption]{todonotes}
\usepackage{soul}
\usepackage{afterpage}
\usepackage[nocenter]{qtree}
\newcommand{\poseyv}{\mathbf{y}_{\negthinspace v}}
\newcommand{\poseyva}{\mathbf{y}_{\negthinspace v{_{1}}}}

\newcommand{\poseyh}{\mathbf{y}_{\negthinspace h}}
\newcommand{\poseyha}{\mathbf{y}_{\negthinspace h{_{1}}}}
\newcommand{\poseyhb}{\mathbf{y}_{\negthinspace h{_{2}}}}

\makeatletter
\if@todonotes@disabled

\else

\fi
\makeatother

\newcommand\ignore[1]{}

\DeclareMathOperator*{\kl}{KL}

\newcommand{\figref}[1]{Fig.~\ref{#1}}

\newcommand{\eg}{e.g.~}
\newcommand{\ie}{i.e.~}
\newcommand{\etal}{et al.}

\def \path{\bp C}

\newcommand{\bfx}{{\mathbf{x}}}
\newcommand{\bfy}{{\mathbf{y}}}
\newcommand{\bfz}{{\mathbf{z}}}

\newcommand{\calD}{{\mathcal{D}}}

\newcommand{\calL}{{\mathcal{L}}}

\newcommand{\bbE}{{\mathbb{E}}}

\begin{document}

\title{DGPose: Deep Generative Models for Human Body Analysis
}

\titlerunning{DGPose: Deep Generative Models for Human Body Analysis}        

\author{Rodrigo de Bem$^{1,2}$ \and Arnab Ghosh$^{1}$ \and Thalaiyasingam Ajanthan$^{1,3}$ \and\\Ondrej Miksik$^{1}$ \and Adnane Boukhayma$^{1}$ \and N. Siddharth$^{1}$ \and Philip Torr$^{1}$}

\authorrunning{Rodrigo de Bem \and Arnab Ghosh \and Thalaiyasingam Ajanthan \and\\Ondrej Miksik \and Adnane Boukhayma \and N. Siddharth \and Philip Torr} 

\institute{$^{1}$Department of Engineering Science, University of Oxford, Oxford, UK.\\
\email{\\\{rodrigo, arnabg, ajanthan,\\omiksik, adnane, nsid, phst\}@robots.ox.ac.uk.}
\\ \\
$^{2}$Center of Computational Sciences, Federal University of Rio Grande, Rio Grande, Brazil.\\
\\
$^{3}$Thalaiyasingam Ajanthan is now at Australian National University, Canberra, Australia.
}

\date{Received: date / Accepted: date}

\maketitle
\begin{abstract}
Deep generative modelling for human body analysis is an emerging problem with many interesting applications. 
However, the latent space learned by such approaches is typically not interpretable, resulting in less flexibility.
In this work, we present deep generative models for human body analysis in which the body pose and the visual appearance are disentangled.
Such a disentanglement allows independent manipu\-lation of pose and appearance, and hence enables applications such as pose-transfer without specific training for such a task.
Our proposed models, the Conditional-DGPose and the Semi-DGPose, have different characteristics. 
In the first, body pose labels are taken as conditioners, from a fully-supervised training set. 
In the second, our structured semi-supervised approach allows for pose estimation to be performed by the model itself and relaxes the need for labelled data.
Therefore, the Semi-DGPose aims for the joint \emph{understanding} and \emph{generation} of people in images.
It is not only capable of mapping images to interpretable latent representations but also able to map these representations back to the image space.
We compare our models with relevant baselines, the ClothNet-Body and the Pose Guided Person Generation networks, demonstrating their merits on the Human3.6M, ChictopiaPlus and DeepFashion benchmarks.

\keywords{Deep generative models, Semi-supervised learning, Human pose estimation, Variational autoencoders, Generative adversarial networks}
\end{abstract}

\section{Introduction}
\label{sec:intro}

\begin{figure*}[ht]
\centering
\renewcommand{\thesubfigure}{a}
\subfloat[Generating different appearances]{\includegraphics[scale=0.65]{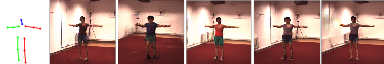}}\hspace{2em}
\renewcommand{\thesubfigure}{c}
\captionsetup{justification=centering}
\subfloat[Pose estimation and pose-transfer]{
\subfloat{\includegraphics[trim={0 0 0 0},clip,scale=0.145]{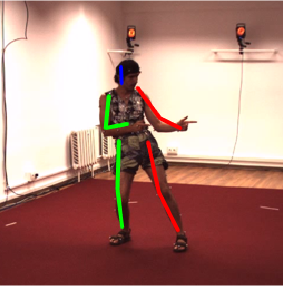}}
\subfloat{\includegraphics[trim={0 .1cm 0 0},clip,scale=0.3625]{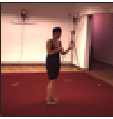}}
\subfloat{\includegraphics[trim={0 .1cm 0 .1cm},clip,scale=0.365]{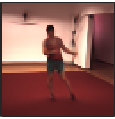}}
}\\
\vspace{-1em}
\renewcommand{\thesubfigure}{b}
\subfloat[Generating different poses]{\includegraphics[scale=0.65]{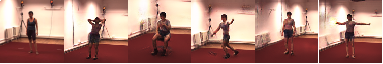}}\hspace{2em}
\renewcommand{\thesubfigure}{d}
\subfloat[Direct manipulation]{
{\includegraphics[scale=0.365]{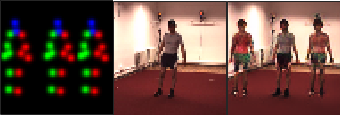}}
}
\caption{\textbf{Sampled results from our deep generative models for images of people.}
(a) For a given pose (first image), we show some samples of appearance. 
(b) For a given appearance (first image), samples of different poses. 
(c) For an estimated pose (first image) and a given appearance (second image), we show a generated sample combining the pose of the first image with the appearance of the second. 
(d) For manipulated poses (first image) and a given appearance (second image), it can \textit{hallucinate} people in the scene.
}
\label{fig:teaser}
\end{figure*}

Human body analysis has been a long-standing goal in computer vision, with many applications in human-machine interaction, health-care, shopping, sports, entertainment and gaming~\cite{patient_mocap,moeslund2011visual,seemann2004head,shotton2011cvpr,SIP}.
Popular approaches to this problem have focused on supervised learning of discriminative models~\cite{bulat2016human,cao2017realtime,chu2017multi,wei2016convolutional}, which map visual inputs (images or videos) to suitable abstract representations (\eg human body pose).
While these approaches do exceptionally well on their prescribed task, as evidenced by their performance on pose estimation benchmarks~\cite{andriluka14cvpr,h36m_pami,Johnson10}, they fall short due to:
\begin{inparaenum}[a)]
\item reliance on fully-labelled data, and
\item the inability to generate novel data from the abstractions.
\end{inparaenum}

The former is a fairly onerous shortcoming, particularly when one is dealing with real-world visual data, as it requires a substantial amount of human time and effort to annotate.
Thus, being able to relax the reliance on labelled data is a highly desirable goal.
The latter, states a rather significant limitation, the incapacity to manipulate abstractions directly with the aim of generating novel visual data.
For instance, changes in the pose of an arm cannot be used for the generation of images or videos in which that arm is correspondingly displaced.

Generative models, in contrast to discriminative ones, enable the \emph{analysis-by-synthesis} of the human body.
With them, ideally, one could generate images of humans in diverse combinations of body poses and appearances, \ie clothing, skin colours, hairstyles, and scenarios.
This has many potential applications.
For instance, it can be used for performance capture and reenactment of RGB videos, as already showcased for faces~\cite{thies2016face2face}, and still incipient for human bodies~\cite{balakrishnansynthesizing,chan2018everybody}.
It can also be used to generate images in user-specified poses, to enhance and augment datasets with minimal annotation effort.

Recently, such approaches have been commonly formulated as deep generative models (DGMs)~\cite{Goodfellow2014,kingma2013auto,rezende2014stochastic} -- an extension of standard generative models that incorporate neural networks as flexible function approximators.
These models are particularly effective in complex perceptual domains such as computer vision~\cite{kulkarni2015deep}, language~\cite{flipdial}, and robotics~\cite{wang2017nips}, effectively delegating bottom-up feature learning to neural networks, while simultaneously incorporating top-down probabilistic semantics into the model.
They solve both the deficiencies of discriminative methods discussed above by
\begin{inparaenum}[a)]
\item employing unsupervised learning, thereby removing the need for labels, and
\item embracing a fully generative modelling.
\end{inparaenum}

However, DGMs introduce a new problem -- the learnt abstractions, or latent variables, are not human-interpretable.
This lack of interpretability is a by-product of the unsupervised learning of representations from data.
The learnt latent variables, usually represented as a smooth high-dimensional mani\-fold, do not have the consistent semantic meaning as different sub-spaces in this manifold can encode arbitrary variations in the data.
This is particularly unsuitable for our purposes as we would like to view and manipulate the latent variables, \eg the body pose.

In order to ameliorate the issue mentioned above, while still eschewing reliance on fully-labelled data, we rely on a structured semi-supervised variational autoencoder (VAE) framework~\cite{kingma2014semi,siddharth2016learning}.
Here, the model structure is assumed to be partially specified, with consistent semantics imposed on some interpretable subset of the latent variables (\eg pose), and the rest is left to be non-interpretable, although referred by us here as appearance.
Weak (semi) supervision acts as a means to constrain the pose latent variables to actually encode the pose.
This gives us the full complement of desirable features, allowing
\begin{inparaenum}[a)]
\item semi-supervised learning, relaxing the need for labelled data,
\item generative modelling through stochastic computation graphs~\cite{schulman2015gradient}, and
\item interpretable subset of latent variables defined through the model structure.
\end{inparaenum}

In this context, we present a structured semi-supervised VAEGAN architecture, the Semi-DGPose, in which we further extend structured semi-supervised models~\cite{kingma2014semi,siddharth2016learning} with a discriminator-based loss function from generative adversarial networks (GANs)~\cite{Goodfellow2014,larsen2015autoencoding},
formulating it as a principled and unified probabilistic framework. 
To our knowledge, it is the first structured semi-supervised deep generative model of people directly learned in the \textit{natural image} (or \textit{natural scene}) space.
This allows the method to directly learn the intricacies in the formation of natural (\ie real) images.
However, it is important to mention that natural images, in contrast to artificial visual stimuli (e.g. segmentation masks, binary masks, or pose vectors), 
have complex statistical structure and are much more challenging to parameterised~\cite{geisler2008visual,kay2008identifying,simoncelli2001natural}. 
Consequently, methods that work well with the latter may not succeed when tackling the former~\cite{fei2005bayesian,krizhevsky2010factored}. 
In contrast to previous work~\cite{LassnerPG17,ma2017,ma2017disentangled,deformable_gans_for_pose_generation,Walker2017}, our model directly enables: 
i) \textit{semi-supervised pose estimation}; and 
ii) {\it indirect pose-transfer} without specific training for such a task,
both of which are tested and verified by experimental evidence.

Additionally, as an intermediate step in the investigation towards our main contribution, we propose a conditional-VAEGAN model, dubbed Conditional-DGPose.
It is less distinct from previous art~\cite{LassnerPG17,ma2017}, however, still differently from earlier work in the literature, it has:
i) allowed pose manipulation on extreme cases, \eg by performing \textit{cross-domain pose-transfer} and by \emph{hallucinating} multiple people, in a variety of unseen or even unrealistic poses; and
ii) achieved state-of-the-art results on image reconstruction conditioned on pose, outperforming the closest related comparable baseline~\cite{LassnerPG17}.
%
We illustrate some capabilities of our models in Fig.~\ref{fig:teaser}.

The present paper builds upon our previous approaches~\cite{debem2018a,debem2019} with further theoretical and technical details, evaluation, and discussion. 
Here, we present in full our comprehensive deep generative model framework for human body analysis in images.
Along with an overview of VAEGAN models, this enables us to shed light on differences and similarities between conditional-VAEGANs and structured semi-supervised VAEGANs.
More precisely, we provide additional evaluations of our Conditional-DGPose and Semi-DGPose models on the most relevant benchmarks in the literature, the Human3.6M~\cite{h36m_pami}, the ChictopiaPlus~\cite{LassnerPG17}, and the DeepFashion~\cite{liu2016deepfashion} datasets.
We also provide new qualitative and quantitative comparisons with the Pose Guided Person Generation ($\text{PG}^2$) baseline~\cite{ma2017}.
The application of our models to real images and the results obtained are essential to show the relevance of interpretable and structured modelling.
This emphasise the effectiveness of the proposals, despite the significant challenge of jointly aim for \emph{understanding} and \emph{generating} people in images.
In summary, our main contributions are:

\begin{inparaenum}[i)]

\item a comprehensive framework for the joint \emph{understanding} and \emph{generation} of people in images, not only capable of mapping images to interpretable latent representations but also capable of mapping these representations back to the image space;

\item a real-world application of structured deep gene\-rative models of images, disentangling pose from appearance in the analysis of the human body;

\item a thorough quantitative and qualitative evaluation of the capabilities of our models; and

\item a demonstration of its principal utilities by performing semi-supervised pose estimation, pose-transfer and pose manipulation.
\end{inparaenum}
%

\section{Related Work}
\label{sec:related_work}

\subsection{Analysing Humans in Images: Overview}

The analysis of people in visual data has been actively investigated as a computer vision and machine learning topic lately~\cite{balakrishnansynthesizing,chan2018everybody,hattori2018synthesizing,hattori2015learning,rogez2016mocap,rogez2018image,thies2016face2face,varol17,Wang2018}.
Historically, the process of synthesising \textit{virtual humans}~\cite{Hilton,Magnenat-Thalmann2005,Magnenat-Thalmann} is a computer graphics undertake since its origins in the '60s, with Boeing's ``first man''~\cite{Boeing2018,Fetter1982}.
Therefore, the geometric and photometric intricacies in the formation of digital images depicting people are well-known in computer graphics, as demonstrated by the existence of many commercial and academic specialised engines~\cite{3Lateral2018,makehuman,MassiveSoftware2017,muller2018sim4cv,poser,unreal2018}.
Nonetheless, the unconstrained creation of truly realistic RGB images is still reasonably dependent upon manual intervention~\cite{IanSpriggs2018}.
Moreover, to produce accurate images of people is harder since humans seem to be very familiarised to corporal traits (\eg faces) even since their early ages~\cite{macdorman2009too,Valenza1996}.

Over time, the generation of humans in images was also embraced by the computer vision community. 
Aiming for less manual intervention, image-based techniques were successfully adopted on matters like rendering and modelling~\cite{blanz1999morphable,Borshukov2005,ezzat1996facial,Kanade1997,Starck2007}.
For instance, a large body of work has relied on geometric 3D models for generating synthetic images of faces~\cite{ichim2015dynamic,pighin2006synthesizing}, bodies~\cite{bogo2014faust,starck2005video}, and hands~\cite{de2011model,romero2017embodied,rosales20013d}.
Despite that, to automatically synthesise artificial images indistinguishable from real ones may be considered as equivalent to succeed in a \textit{visual Turing test}~\cite{Shan2013}.
Hence, a substantially complicated and consequently yet unsolved challenge~\cite{Fan2018}.

Another line of approaches, following the machine learning methodologies closely, had modelled the image formation by designing and learning probabilistic generative models~\cite{enzweiler2008mixed,fleuret2007multicamera,fossati2007bridging,franco2005,lee2005facial,wang2004high,yuille2006vision}.
However, it is highly complex and constrained due to intractable probability distributions and the high variability of latent factors.
Often, simplifying assumptions are made in practice, such as independence between different factors of variation, leading to weak generative models that fail to capture statistical subtleties.

Recently, the advent of the deep generative models (DGMs)~\cite{Goodfellow2014,kingma2013auto,rezende2014stochastic} somehow gathers the three lines of methods mentioned above.
Bringing together characteristics from computer graphics, computer vision, and machine learning makes the DGMs a powerful \textit{analysis-by-synthesis} framework.
We discuss the DGM-based approaches related to our work in the following section.

\subsection{Analysing Humans in Images with DGMs}
Generally, in \emph{classical DGMs}, such as standard VAEs and GANs, pose representation is non-interpretable and unsupervised, entangled with the visual appearance in the latent space.
This is similarly employed by some \emph{image-to-image translation networks}, however, in contrast to the relatively low-dimensional manifolds learned by the DGMs, in the latter case high-dimensional abstractions are learned and used strictly for direct mapping from and to the image space. 
On the other hand, \emph{conditional DGMs} usually define part of the abstract data representation, \ie body pose, to be an interpretable and observable random variable, while the rest of the representation (visual appearance) is kept non-interpretable and latent, still subjected to unsupervised learning.
Finally, in \emph{structured DGMs} approaches, as the Semi-DGPose, the latent space can be simultaneously composed by interpretable and non-interpretable random variables.
In the former case, the variables may be fully or semi-supervised, while in the latter group they are still maintained unsupervised. 
Below, we describe related literature gathering the methods according to their adopted type of approach.

\paragraph{Image-to-image networks.} Ma~\etal~\cite{ma2017} introduce the Pose Guided Person Generation Network (PG$^2$), a two stage image-to-image translation model which is trained on pairs of images of the same person in different poses, scales and points of view.
The authors admit the difficulty of generating poses and detailed appearance simultaneously in an end-to-end fashion.
Their model, which is conditioned on images rather than poses, does not allow sampling, thus in its essence, it is not a generative model, which is again in contrast to our single-stage approaches.
In a second proposal, Ma~\etal~\cite{ma2017disentangled} present a GAN-based model for learning image embeddings of foreground, background and pose variables encoded as interpretable variables.
The method is still limited to training and testing with cross-pose/scale pairs for pose-transfer, however, it allows sampling, differently from the PG$^2$.
In contrast to our Semi-DGPose model, it is not capable of performing either pose estimation or semi-supervised learning, relying on off-the-shelf pose estimators to perform pose-transfer.

Recently, Esser~\etal~\cite{esser2018variational} present a conditional image-to-image translation network based on the U-Net~\cite{ronneberger2015u}.
The model is conditioned on an appearance encoding obtained using a VAE architecture. 
It is more versatile than~\cite{ma2017,ma2017disentangled}, although still not capable of producing either an interpretable encoding of pose (pose estimation) or performing semi-supervised learning.
Similarly, Balakrishnan~\etal~\cite{balakrishnansynthesizing} also propose a U-Net-based approach.
In this case, the authors make use of three U-Nets which tackle foreground segmentation and synthesis, as well as background synthesis.
The model is trained with video sequences of the same person performing a limited set of activities.
Therefore, it is limited to translating images of the same person to different poses.
Other very recent approaches~\cite{chan2018everybody,neverova2018dense} have to be explicitly trained for pose-transfer, \ie using images pairs, and do not have the capability of predicting pose.
This is in sharp contrast to our Semi-DGPose approach, in which we learn pose estimation, while pose-transfer is achieved as a by-product.
In the method by Trumble~\etal~\cite{trumble2018deep}, pose is estimated from multiple views, although it does not allow semi-supervised learning.

Rhodin~\etal~\cite{rhodin2018unsupervised} learn 3D pose estimation from multi-view images of the same person acquired from synchronised and calibrated cameras. 
In contrast to our approach, their method explicitly uses the rotation matrix between cameras during training for the unsupervised learning of a geometry-aware latent representation.
From such representation, the 3D pose is estimated posteriorly with a shallow network.
The authors do not define their method as a generative model, but as a 3D pose estimator, although it can perform novel viewpoint synthesis.
Another work by Zanfir~\etal~\cite{zanfir2018human} focus uniquely on the specific task of appearance transfer, also based on 3D pose.
In contrast, our closely related task of pose-transfer is just one among all the tasks our DGMs can perform (\eg~sampling, pose estimation, direct manipulation) employing only 2D pose representations.
Lastly, Zhang~\etal~\cite{zhang2018unsupervised} focus on a slightly different task.
They propose the unsupervised discovery of 2D landmarks using optical flows from Human3.6M videos as a short-term self-supervision.
Such landmarks are an intermediate representation of pose since they do not correspond explicitly to specific body parts.
In contrast, we employ single still images using directly and explicitly interpretable pose representations.

Finally, it is essential to differentiate such image-to-image translation methods from our DGMs.
The former depends upon input images at test time, while the latter effectively allow sampling from the latent structured representations learned during training.
This subtle difference means that such structured representations are responsible for learning the underlying factors of variations in image generation, without relying on information from input images for generating outputs at test time.

\paragraph{Classical DGMs.} Lassner~\etal~\cite{LassnerPG17} have proposed the ClothNet-full model, in which a VAE model is used to learn a latent representation of segmentation masks of people in given poses.
The reconstructed masks are mapped back to the image space by an image-to-image translation module based on~\cite{isola2016image}.
In contrast, we learn our generative models directly on the raw image data without the need for body parts segmentation.
Moreover, pose is interpretable in both of our methods.
Siarohin~\etal~\cite{deformable_gans_for_pose_generation} propose a GAN model with skip connection in the generator and a discriminator conditioned on pose. 
Similarly to~\cite{ma2017}, the model is restricted to pose-transfer on pairs of images of the same person.
The body pose is always given to the model and non-interpretable in the learned latent encoding.
Apart from this, Walker~\etal~\cite{Walker2017} proposed a hybrid architecture, associating a VAE and a GAN for forecasting future poses in a video.
Here, a low-dimensional pose representation is learned using a VAE, and once the future poses are predicted, they are mapped to images using a GAN generator.
Considering GAN based generative models, Tulyakov~\etal~\cite{Tulyakov2017} present a GAN network that learns motion and content in two separate latent spaces in an unsupervised manner.
However, it does not allow explicit manipulation over the human pose.

\paragraph{Conditional DGMs.} Lassner~\etal~\cite{LassnerPG17} present a second model, the ClothNet-Body, which is a CVAE conditioned on human pose.
This model is closely related to our Conditional-DGPose, but it also uses low-dimensional segmentation masks and an auxiliary image-to-image transfer network, based on~\cite{isola2016image}, to generate realistic images.
Pumarola~\etal~\cite{pumarola2018unsupervised} propose an unsupervised image synthesis based on a conditional GAN method, yet it is also not capable of performing pose prediction.
\newline

In summary, there are methods in the literature closely related to our Conditional-DGPose, mainly due to its conditional nature.
Although, to our knowledge, no other method gathers the capabilities of our Semi-DGPose as a \emph{structured DGM}.
The novelty in the Semi-DGPose largely relies on how the body pose is handled, differing it from related work. 
Moreover, the capacity for performing pose estimation, indirect pose-transfer, and semi-supervised learning, while aiming for joint \emph{understanding} and \emph{generation} of people in images is peculiar to our model.
Following Larsen~\etal~\cite{larsen2015autoencoding}, we use a discriminator in our training to improve the quality of the generated images.
However, in contrast to~\cite{larsen2015autoencoding}, the latent space of our approach is interpretable, which enables us to sample different poses and appearances.

\section{Preliminaries}
\label{sec:preliminaries}
Deep generative models (DGMs) come in two broad flavours -- Variational Autoencoders (VAEs)~\cite{kingma2013auto,rezende2014stochastic}, and Generative Adversarial Networks (GANs)~\cite{Goodfellow2014}.
In both cases, the goal is to learn a generative model~\(p_{\theta}(\bfx, \bfz)\) over data~\(\bfx\) and latent variables~\(\bfz\), with parameters~\(\theta\).\@
Typically the model parameters~\(\theta\) are represented in the form of a neural network.

VAEs express an objective to learn the parameters~\(\theta\) that maximise the marginal likelihood (or evidence) of the model denoted as \( p_{\theta}(\bfx) = \int p_{\theta}(\bfx | \bfz) p_{\theta}(\bfz) dz\).
They introduce a conditional probability density~\(q_{\phi}(\bfz | \bfx)\) as an approximation to the unknown and intractable model posterior~\(p_{\theta}(\bfz | \bfx)\), employing the variational principle in order to optimise a surrogate objective~\(\calL(\phi, \theta; \bfx)\), called the evidence lower bound (ELBO), as
\begin{align}
\qquad\qquad \log\, p_{\theta}(\bfx) & \ge \calL_{\text{VAE}}(\phi, \theta; \bfx ) \nonumber \\
  & =  \bbE_{q_{\phi}(\bfz|\bfx)} \left[\log \frac{p_{\theta}(\bfx, \bfz)}{q_{\phi}(\bfz|\bfx)}\right].
\label{eq:vae}
\end{align}

The conditional density~\(q_{\phi}(\bfz | \bfx)\) is called the recognition or inference distribution, with parameters~\(\phi\) also represented in the form of a neural network.
Lastly, VAEs also admit an extension to \emph{conditional} generative models (CVAEs)~\cite{sohn2015learning}, simply by incorporating a conditioning variable~\(\bfy\), 
to derive
\begin{align}
\quad\qquad \log\, p_{\theta}(\bfx | \bfy) & \ge \calL_{\text{CVAE}}(\phi, \theta; \bfx | \bfy ) \nonumber \\
& = \bbE_{q_{\phi}(\bfz|\bfx, \bfy)} \left[\log \frac{p_{\theta}(\bfx, \bfz | \bfy)}{q_{\phi}(\bfz | \bfx, \bfy)}\right].
\label{eq:cvae}
\end{align}

On the other hand, in the context of structured semi-supervised learning, one can factor the latent variables into unstructured or non-interpretable variables~\(\bfz\) and structured or interpretable variables~\(\bfy\) without loss of generality~\cite{kingma2014semi,siddharth2016learning}.
For learning in this framework, the objective can be expressed as the combination of supervised and unsupervised objectives.
Let~\(\calD_u\) and~\(\calD_s\) denote the unlabelled and labelled subset of the dataset~\(\calD\), and let the joint recognition network factorise as~\(q_{\phi}(\bfy, \bfz | \bfx) = q_{\phi}(\bfy | \bfx) q_{\phi}(\bfz | \bfx, \bfy )\).
Then, the combined objective summed over the entire dataset corresponds to 
\begin{align}
\calL_{\text{SS}}(\theta, \phi; \calD) &= \sum_{\bfx_u \in \calD_u} \calL_u(\theta, \phi; \bfx_u) \nonumber \\
&  + \gamma \!\!\!\!\!\!\! \sum_{(\bfx_s, \bfy_s) \in \calD_s} \!\!\!\!\!\!\! \calL_s(\theta, \phi; \bfx_s, \bfy_s)
\label{eq:ssvae} \\
\intertext{where~\(\calL_u\) and~\(\calL_s\) are defined as}
\calL_u(\theta, \phi; \bfx_u) &= \calL_{\text{VAE}}(\theta, \phi; \bfx_u) \text{, and} \label{eq:ss:unsup}\\
\calL_s(\theta, \phi; \bfx_s, \bfy_s) &= \bbE_{q_{\phi}(\bfz | \bfx_s, \bfy_s)}
  \left[
  \log \frac{p_{\theta}(\bfx_s, \bfz | \bfy_s)}{q_{\phi}(\bfz | \bfx_s, \bfy_s)}
  \right] \nonumber \\
  & + \alpha \log q_{\phi}(\bfy_s | \bfx_s), \label{eq:ss:sup}
\end{align}
respectively. Here, the hyper-parameter~\(\gamma\) (Eq.~\ref{eq:ssvae}) controls the relative weight between the supervised and unsupervised dataset sizes, and~\(\alpha\) (Eq.~\ref{eq:ss:sup}) controls the relative weight between generative and discriminative learning.

Note that by the factorisation of the generative model, VAEs necessitate the specification of an explicit likelihood function~\(p_{\theta}(\bfx | \bfz)\), which can often be difficult.
GANs, on the other hand, attempt to sidestep this requirement by learning a surrogate to the likelihood function, while avoiding the learning of a recognition distribution.
Here, the generative model~\(p_{\theta}(\bfx, \bfz)\), viewed as a mapping~\(G: \bfz \mapsto \bfx\), is setup in a two-player minimax game with a ``discriminator''~\(D: \bfx \mapsto \{0,1\}\), whose goal is to correctly identify if a data point~\(\bfx\) came from the generative model~\(p_{\theta}(\bfx, \bfz)\) or the true data distribution~\(p(\bfx)\). Such objective is defined as
\begin{align}
\qquad\quad  \calL_{\text{GAN}}(D,G) & = \bbE_{p(\bfx)}\left[\log D(\bfx)\right] \nonumber \\
  & + \bbE_{p_{\theta}(\bfz)}\left[1 - \log D(G(\bfz))\right].
  \label{eq:gan}
\end{align}
In fact, in our structured model, generation is defined as a function of pose and appearance as $G(\bfy,\bfz)$.
Crucially, learning a customised approximation to the likelihood can result in a much higher quality of generated data, particularly for the visual domain~\cite{improving_gans}.

A more recent family of DGMs, VAEGANs~\cite{larsen2015autoencoding}, bring together these two different approaches into a single objective that combines both the VAE and GAN objectives directly as
\begin{align}
\qquad\qquad\qquad\qquad \calL = \calL_{\text{VAE}} + \calL_{\text{GAN}}.
\label{eq:vaegan}
\end{align}
This marries better the likelihood learning with the inference-distribution learning, providing a more flexible family of models.

\section{Our Approach \label{sec:approach}}
As set out in the preliminaries (Sec.~\ref{sec:preliminaries}), we use the VAEGAN framework as the basis for our generative models~\cite{larsen2015autoencoding}.
Note that, in incorporating semi-supervised learning, the semi-supervised VAEGAN includes two distinct tasks.
First, it involves learning a recognition network that can estimate pose $\bfy$ and \textit{appearance} $\bfz$ for any given RGB image $\bfx$.
Second, it involves learning a generative network that combines a given pose with an appearance to generate visual data (RGB image) corresponding to those variables.

From discriminative modelling, we know that the first task, \ie predicting pose, is eminently plausible up to learning an appearance model.
However, learning the full generative model is something that can be fraught with difficulties.
For one, pose and appearance can exhibit a large degree of information imbalance -- pose can be distilled into a set of~\((x, y)\) coordinates, whereas appearance can encode a vast swathe of information (e.g. texture, colour, shapes) about the given input.

Given a generative model that takes both appearance~\(\bfz\) and pose~\(\bfy\) as inputs to produce an RGB image~\(\bfx\), a reasonable first step can be just to evaluate the performance of a conditional generative model, where the conditioning variable is taken to be the interpretable pose~\(\bfy\). 
We refer to this setup as Conditional-DGPose, with reference to the fact that it is a conditional-VAEGAN model. 
Its lower bound is given by Eq.~\ref{eq:cvae},
and its final objective function is defined as 
\begin{align}
\qquad\qquad\qquad\quad \calL = \calL_{\text{CVAE}} + \calL_{\text{GAN}}, \label{eq:cvaegan_obj}
\end{align}
in contrast to the standard VAEGAN objective (Eq.~\ref{eq:vaegan}).
Here, all data is ``labelled'' with pose, but the goals were: i) primarily, to verify qualitatively if a low\hyp{}dimensional conditioning variable would affect the conditional generative model;
ii) secondly, to evaluate the accuracy of the reconstructed images quantitatively w.r.t. the human body poses and the image quality.

Once verified through experiments that the conditional approach works, we could then proceed towards our structured semi-supervised VAEGAN, referred to as Semi-DGPose, as its main difference from the previous setup is that the encoding distribution is no longer conditioned on the pose, but instead predicts it as per Eq.~\ref{eq:ssvae}--\ref{eq:gan}.
In contrast to the standard VAEGAN objective (Eq.~\ref{eq:vaegan}), the structured semi-supervised VAEGAN final objective function is given by, 
\begin{align}
\qquad\qquad\qquad\qquad \calL = \calL_{\text{SS}} + \calL_{\text{GAN}}. \label{eq:vaegan_obj}
\end{align}

We describe the details and implementations of our models in the rest of this section.
Next, we start defining the adopted pose representations, which are common for both, the Conditional-DGPose and the Semi-DGPose architectures.

\subsection{Pose Representation\label{sec:pose_rep}}
\begin{figure}[ht]
  \centering
  \includegraphics[width=0.5\linewidth]{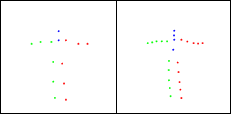}
  \subfloat[\label{fig:2d_vec}]{\hspace{-24em}}
  \subfloat[\label{fig:2d_vec_ext}]{\hspace{-9em}}
  \caption{\small \textbf{Vector representation.} (a) $J=14$ joints which compose a 2D pose vector $\poseyv \in \mathcal{R}^{2J}$.
  (b) An extended 2D vector composed by 24 body parts ($J=14$ annotated joints, $R=9$ intermediate points between joints and $B=1$ central point), such that $\poseyv \in \mathcal{R}^{2(J+R+B)}$.}
  \label{fig:pose_rep_vec}
\end{figure}

In our DGMs, the random variable $\bfy$ corresponds to an abstraction of the human body pose.
Therefore a suitable concrete representation must be adopted in the implementation of the models.
As mentioned in our literature review, many methods which define a generative model in the \textit{pose space} would simply encode $J$ joints defining the body as a vector $\poseyv$, such that $\poseyv \in \mathcal{R}^{2J}$.
Others employ extended versions of it, in which positions of $R$ rigid parts and $B$ whole body are derived from the annotated joints~\cite{yang2011articulated}, such that $\poseyv \in \mathcal{R}^{2(J+R+B)}$. 
Both cases are illustrated in Fig.~\ref{fig:pose_rep_vec}.

\begin{figure}[ht]
  \centering
  \includegraphics[width=0.75\linewidth]{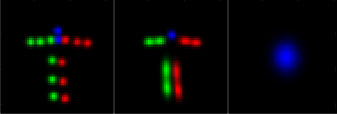}
  \subfloat[\label{fig:hm_joints}]{\hspace{-33em}}
  \subfloat[\label{fig:hm_rigid}]{\hspace{-20em}}
  \subfloat[\label{fig:hm_body}]{\hspace{-8em}}
  \caption{\small \textbf{Heatmap representation.} Heatmaps superimposed corresponding to (a) $J=14$ annotated joints, (b) $R=9$ rigid parts, and (c) $B=1$ whole body;
  such that $\poseyh \in \mathcal{R}^{P\times H\times W}$.
  Right, left and central body parts are denoted by the colours green, blue and red, respectively, in the person-centric representation.}
  \label{fig:pose_rep_hms}
\end{figure}

On the other hand, the mapping of 2D joints positions to heatmaps has shown to be very effective in several pose estimation approaches~\cite{chu2017multi,newell2016,Tompson2014,wei2016convolutional}.
The Gaussian heatmaps represent the underlying probability distribution of body parts' locations.
In our method, the heatmap representation $\poseyh$ consists of $P$ body elements, in a way that $\poseyh \in \mathcal{R}^{P\times H\times W}$, where $H$ and $W$ are the heatmap height and width, respectively. 
In the simplest case $P = J$, however, as the set of joints is reasonably sparse, to cover the entire area of the bodies, joints, rigid parts and the whole body might be used as an extended case, in which $P = J+R+B$~\cite{debem2018b}, as illustrated in Fig.~\ref{fig:pose_rep_hms}.
In this way, each body element $p$ is represented using a 2D Gaussian around its centre $\boldsymbol{\mu}_p = (i_p, j_p)$, 
with diagonal covariance matrix $\Sigma_p = R_p \left[ \begin{smallmatrix} \sigma_{p,i}^2 & 0 \\ 0 & \sigma_{p,j}^2 \end{smallmatrix}\right] R_p^\top$, computed as follows:

\paragraph{\textbf{Joints.}} Since joints have a limited spatial extent, we follow previous approaches~\cite{chu2017multi,newell2016,Tompson2014,wei2016convolutional} in modelling them as isotropic
Gaussians that are centred at the ground-truth joint location and have a small standard deviation (e.g. $\sigma_{p,i} = \sigma_{p,j}$ = 1.5 pixel for a $64 \times 64$ heatmap).

\paragraph{\textbf{Rigid Parts.}} The centre $\boldsymbol{\mu}_p$ of a rigid part $p$ is defined as the mean point of the centres $\boldsymbol{\mu}_{k}$ and $\boldsymbol{\mu}_{l}$ of the joints it connects.
We orient the Gaussian representing the rigid part to align its $i$ axis with the line connecting $\boldsymbol{\mu}_k$ and $\boldsymbol{\mu}_l$.
We define $\sigma_{p,i}$ to be proportional to $|\boldsymbol{\mu}_k - \boldsymbol{\mu}_l|$, and set $\sigma_{p,j}=\kappa_p \sigma_{p,i}$, where $\kappa_p$ is a part-specific ratio, inspired by anthropometric measurements~\cite{msisNasa1995}.

\paragraph{\textbf{Body.}} The body centre is defined to be the mean of the annotated joint centres. Principal component analysis (PCA) of the joint centres is used to obtain the orientation of the body in the image plane. 
We define $\sigma_{p,i}$ and $\sigma_{p,j}$ to be proportional to the distance between the extreme projections of the joint centres onto, respectively, the principal and secondary axes of variation.

In our both models, as detailed in the next sections, we make use of both forms of pose representation, taking advantage of their particular characteristics in each case.
In the Conditional-DGPose, only the heatmap representation $\poseyh$ is employed, since, as shown later in our experiments, it can be seamlessly concatenated to feature maps, helping on the generation of accurate output images.
On the other hand, in the Semi-DGPose model, we additionally employ the vector-based form $\poseyv$, as a way of maintaining a low-dimensional latent representation of pose.
\begin{figure*}[ht]
  \centering
  \includegraphics[width=.95\linewidth]{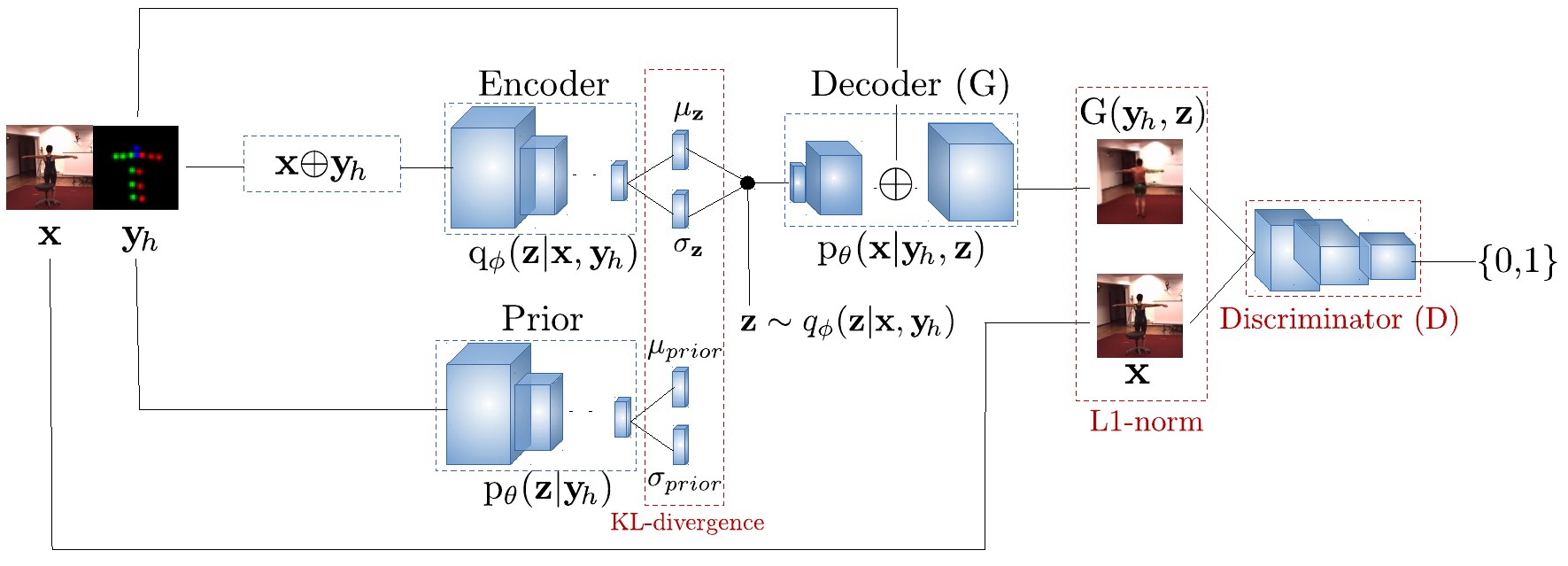}
  \caption{\small \textbf{Conditional-DGPose architecture.} 
  At the training, the Encoder receives $\bfx \oplus \poseyh$ as input and learns the posterior $q_{\phi}(\bfz | \bfx, \poseyh)$.
  The Prior module receives $\poseyh$ alone and learns the distribution $p_{\theta}(\bfz | \poseyh)$.
  Appearance is sampled $\mathbf{z} \sim q_{\phi}(\mathbf{z}|\mathbf{x},\poseyh)$, using the reparametrization trick~\cite{kingma2013auto}, and passed to the Decoder, as well as the conditioning pose $\poseyh$, which is concatenated to the Decoder feature maps.
  The Decoder then generates a reconstructed image $G(\poseyh,\bfz)$.
  The loss function (see Eq.~\ref{eq:cvaegan_obj}, Sec.~\ref{sec:approach}) is composed by the following terms, highlighted in red:
  the L1-norm $L1(\bfx, G(\poseyh,\bfz))$ which is computed between the original and the reconstructed image;
  the KL-divergence $\text{KL}[q_{\phi}(\bfz|\bfx,\poseyh)||p_{\theta}(\bfz|\poseyh)]$, which is used to regularise the posterior distribution;
  and the GAN Discriminator cross-entropy loss used to learn how to discern between real and generated images.
}
\label{fig:cvae_arch}
\end{figure*}
\subsection{DGPose Architectures\label{sec:dgpose_arch}}
We have tested several variations of deep CNN architectures for implementing our models, culminating in our best performing ones, which are described here.
All its modules are deep CNNs, and full implementation definitions are given in the appendix (Sec.~\ref{sec:dgpose_arch_details}) and referred adequately in the text.
Due to the generality of generative models, the architectures may be employed in different ways according to the aimed tasks. 
Thus, we describe separately training and test phases, dividing the latter into \textit{reconstruction}, \textit{pose-transfer}, \textit{sampling} and \textit{pose-estimation}, for both models.
Thus, the Conditional-DGPose and the Semi-DGPose are described following.
\subsubsection{Conditional-DGPose}
Our conditional-VAEGAN model learns the parameters of four deep CNN networks simultaneously:
i) a recognition network (Encoder), which estimates appearance $\bfz$ conditioned to pose $\poseyh$ and to a given RGB image $\bfx$;
ii) a Prior network, which estimates appearance $\bfz$ conditioned to pose $\poseyh$ alone;
iii) a generative network (Decoder), which combines appearance $\bfz$ and the conditioning pose $\poseyh$, to generate corresponding RGB images $G(\poseyh,\bfz)$; and
iv) a Discriminator network, which differentiates between real images $\bfx$ and generated images $G(\poseyh,\bfz)$.
Learning is pursued by the minimisation of the loss function \(\calL = \calL_{\text{CVAE}} + \calL_{\text{GAN}}\) (Eq.~\ref{eq:cvaegan_obj}, Sec.~\ref{sec:approach}), composed by the CVAE evidence lower bound (ELBO) $\calL_{\text{CVAE}}$ and by the GAN cross-entropy discriminator loss $\calL_{\text{GAN}}$.
An overview of our model is shown in Fig.~\ref{fig:cvae_arch} and implementation details are provided in Tab.~\ref{table:dgpose_arch} (appendix).
Below, we describe further the training and the test phases, dividing the latter into \textit{reconstruction}, \textit{pose-transfer} and \textit{sampling}.
\paragraph{\textbf{Training.}} 
Given an image $\bfx$, the corresponding heatmap labels (conditioning pose) are concatenated to it as per $\bfx \oplus \poseyh$ (Encoder, Layer 1, Tab.~\ref{table:dgpose_arch}).
Then, the Encoder estimates the conditional posterior distribution $q_{\phi}(\bfz|\bfx,\poseyh)$.
The heatmap labels $\poseyh$ alone are the input of the Prior module, which estimates the distribution $p_{\theta}(\bfz | \poseyh)$.
Appearance is sampled from the posterior $\bfz \sim q_{\phi}(\bfz|\bfx,\poseyh)$, using the reparametrisation trick~\cite{kingma2013auto}.
The sample $\bfz$, along with the conditioning pose $\poseyh$ (Decoder, Layer 7, Tab.~\ref{table:dgpose_arch}), are passed through the Decoder which generates a reconstructed image $G(\poseyh,\bfz)$.
This reconstructed image, along with the real image $\bfx$, are still used as inputs for the Discriminator module, which learns how to discern between them.
Finally, the overall loss function minimised during training is composed of the L1-norm reconstruction loss $L1(\bfx, G(\poseyh,\bfz))$;  
the KL-divergence, which acts as a regulariser, between the posterior and the prior distributions, $\kl[q_{\phi}(\bfz|\bfx,\poseyh)|p_{\theta}(\bfz | \poseyh)]$; 
and the cross-entropy Discriminator loss (Eq.~\ref{eq:gan}, Sec.~\ref{sec:preliminaries}).

\begin{figure}[h]
\centering
\includegraphics[scale=.4]{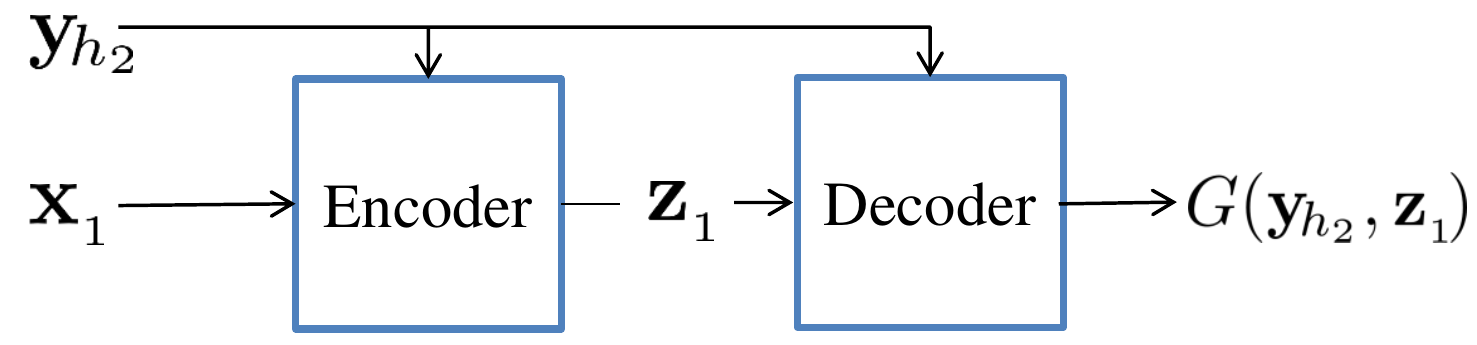}
  \captionsetup{aboveskip=-1em,justification=centering}
\caption{Conditional-DGPose direct pose-transfer and manipulation at test time.\label{fig:reconstruction_diag_cond}}
\end{figure}

\paragraph{\textbf{Reconstruction and Direct Pose-transfer.}} 
At test time, when an image $\bfx_{_1}$ and its corresponding pose $\poseyha$ are given as input, the reconstructed image $G(\poseyha, \bfz_{_1})$ is obtained as the Decoder output. 
However, if $\bfx_{_1}$ is used as input along with a different pose $\poseyhb$, the person in the reconstructed image $G(\poseyhb, \bfz_{_1})$ will keep the appearance of $\bfx_{_1}$, with the body pose defined by $\poseyhb$, as illustrated in Fig.~\ref{fig:reconstruction_diag_cond}.
Similarly, as shown later in our experiments, the same procedure may be adopted to \emph{directly manipulate} the reconstructed image, such as changing body size and aspect ratio, moving or suppressing body parts or even hallucinating multiple people.

\begin{figure}[h]
\centering
\includegraphics[scale=.4]{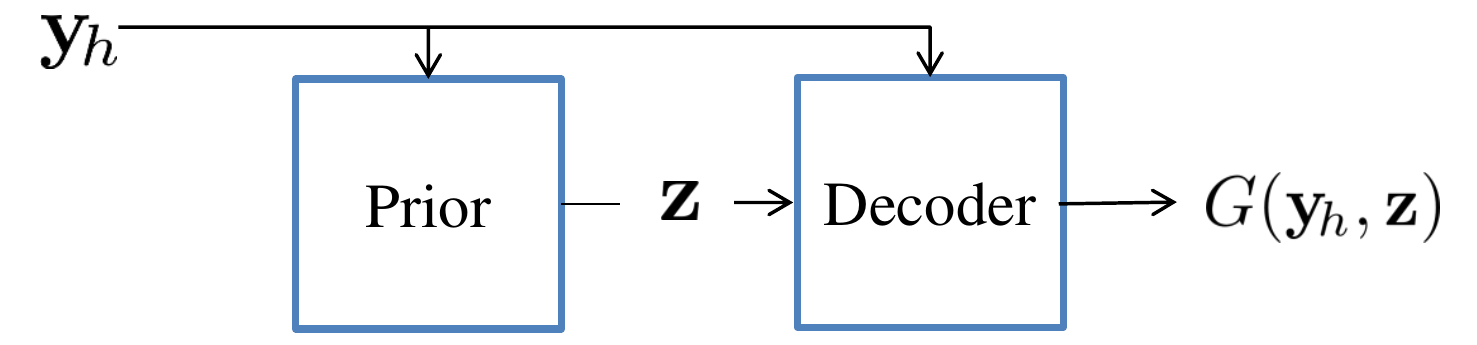}
  \captionsetup{aboveskip=-1em,justification=centering}
\caption{Conditional-DGPose sampling at test time.\label{fig:sampling_diag_cond}}
\end{figure}

\paragraph{\textbf{Sampling.}} 
At test time, sampling is obtained when no RGB image is given as input. 
In this case, as illustrated in Fig.~\ref{fig:sampling_diag_cond}, only a conditioning pose $\poseyh$ is given as the input of the Prior module, which defines $p_{\theta}(\bfz|\poseyh)$.
From this Prior distribution, the sampled appearance $\bfz$ and the conditioning pose $\poseyh$ are passed to the Decoder network. 
In this manner, for a given pose, different appearances can be randomly created from the learned generative model.

\subsubsection{Semi-DGPose \label{sec:exp_det_semi}}
\begin{figure*}[ht]
  \centering
  \includegraphics[width=.875\linewidth]{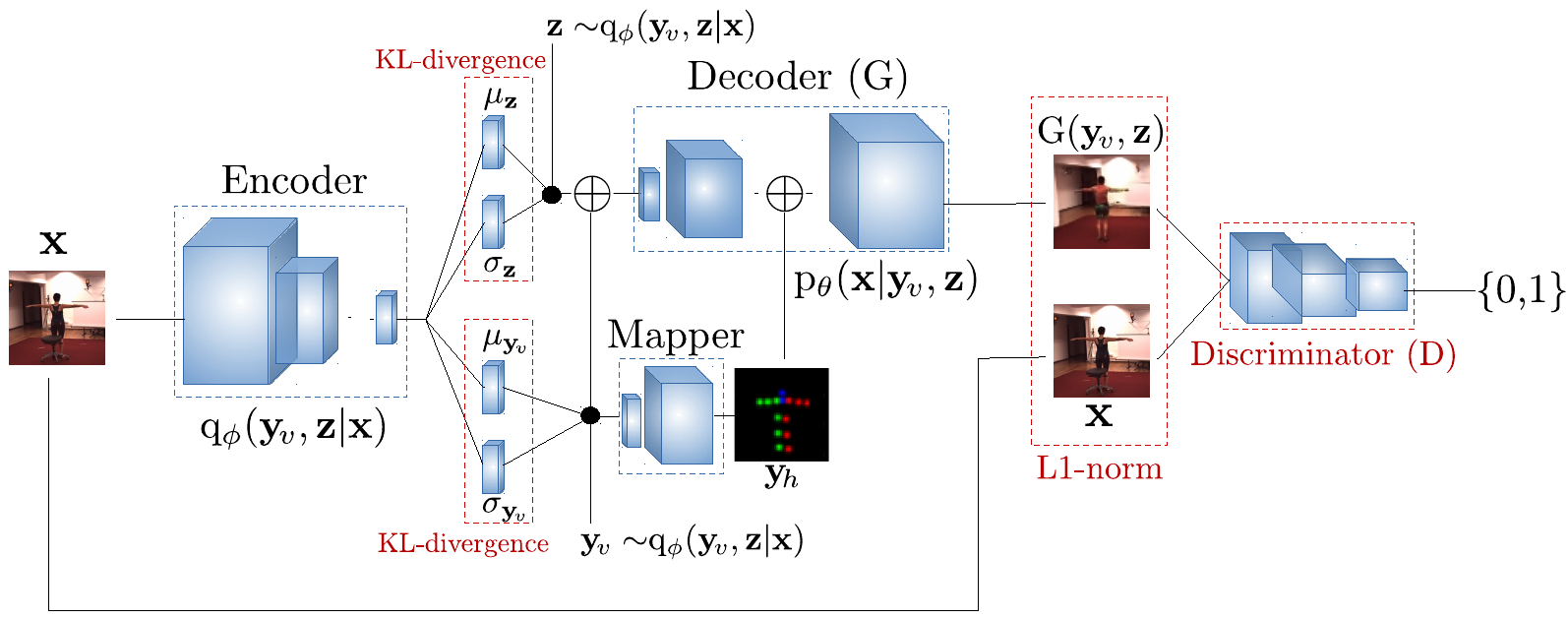}
  \caption{\textbf{Semi-DGPose architecture.} At the training, the Encoder receives $\bfx$ as input and learns the posterior distribution $q_{\phi}(\poseyv, \bfz | \bfx)$. 
  In the \emph{unsupervised} routine, samples of appearance $\bfz$ and pose $\poseyv$ are obtained using the reparametrisation trick~\cite{kingma2013auto}.
  These samples are passed to the Decoder, which generates a reconstructed image $G(\poseyv,\bfz)$.
  The unsupervised loss function is composed by the following terms, highlighted in red:
  the L1-norm $L1(\bfx, G(\poseyv,\bfz))$ between the original and the reconstructed images;
  the KL-divergence losses between the posterior distribution $q_{\phi}(\poseyv, \bfz | \bfx)$ and the weak priors $p(\poseyv)$ and $p(\bfz)$, which work as regularisers (see Eq.~\ref{eq:ss:unsup}, Sec.~\ref{sec:preliminaries});
  and the cross-entropy Discriminator loss (Eq.~\ref{eq:gan}, Sec.~\ref{sec:preliminaries}).
  In the \emph{supervised} routine (not shown above for simplicity), the only difference is that a regression loss between the estimated pose and the pose ground-truth label substitutes the KL-divergence over the pose posterior distribution (see Eq.~\ref{eq:ss:sup}, Sec.~\ref{sec:preliminaries}).
  In both, supervised and unsupervised training routines, the low-dimensional pose vector $\poseyv$ is mapped to a heatmap representation $\poseyh$ by the Mapper module and concatenated to the Decoder.
  Eq.~\ref{eq:ssvae} (Sec.~\ref{sec:preliminaries}) shows the overall loss function.
  }
  \label{fig:semi-vae_arch}
\end{figure*}

Differently from the Conditional-DGPose, our structured semi-supervised VAEGAN model (Fig.~\ref{fig:semi-vae_arch}) learns the parameters of three deep CNN networks simultaneously:
i) a recognition network (Encoder), which estimates appearance $\bfz$ and pose $\poseyv$ from a given RGB image $\bfx$;
ii) a generative network (Decoder), which combines appearance $\bfz$ and pose $\poseyv$, to generate corresponding RGB images $G(\poseyv,\bfz)$; and
iii) a Discriminator network, which differentiates between real images $\bfx$ and generated images $G(\poseyv,\bfz)$.
Learning is pursued by the minimisation of the loss function \(\calL = \calL_{\text{SS}} + \calL_{\text{GAN}}\) (Eq.~\ref{eq:vaegan_obj}, Sec.~\ref{sec:approach}), composed by the structured semi-supervised VAE evidence lower bound (ELBO) $\calL_{\text{SS}}$ and by the GAN cross-entropy discriminator loss $\calL_{\text{GAN}}$.
A fourth module, called Mapper, is introduced by us to overcome a peculiarity caused by the inclusion of pose in the latent space. 
Such a module, trained separately, is described next.

\paragraph{\textbf{The Mapper Module.}} 
Our preliminary experiments with the Conditional-DGPose showed that heatmaps led to better quality reconstructions, in contrast to the vector-based representation. 
On the other hand, a low-dimensional representation is more suitable and desirable as a latent variable, since human pose lies in a low-dimensional manifold embedded in the high-dimensional image space~\cite{elgammal2004inferring,goodfellow2016deep}.
To cope with this mismatch, we introduce the Mapper module, which maps pose-vectors $\poseyv$ to heatmaps $\poseyh$.
Ground-truth heatmaps are constructed from manually annotated 2D joints labels, using a simple weak annotation strategy~\cite{debem2018b}. 
The Mapper module is then trained to map 2D joints to heatmaps, minimising the L2-norm between predicted and ground-truth heatmaps.
This module is trained separately with the same training hyper-parameters used for our full architecture, described later in Sec.~\ref{sec:hyper-param}.
In the training of the full Semi-DGPose architecture, the Mapper module is integrated to it with its weights kept fixed, since the mapping function has been learned already.
The Mapper allows us to keep a low-dimensional representation $\poseyv$ in the latent space, at the same time that a dense high-dimensional ``spatial'' heatmap representation $\poseyh$ facilitates the generation of accurate images by the Decoder.
As it is fully differentiable, the module allows the gradients to be backpropagated normally from the Decoder to the Encoder, when it is required during the training of the full architecture.

In the rest of this section, we describe further the training and the test phases, dividing the latter into \textit{reconstruction}, \textit{indirect pose-transfer}, \textit{sampling} and \textit{pose estimation}.
An overview of our model is shown in Fig.~\ref{fig:semi-vae_arch} and implementation details are provided in Tab.~\ref{table:pose_model_semi_sup_arch} (appendix).

\paragraph{\textbf{Training.}}
The terms of Eq.~\ref{eq:ssvae} (Sec.~\ref{sec:preliminaries}) correspond to two training routines which are alternately employed, according to the presence or absence of ground-truth labels.

In the \textit{unsupervised case}, when no label is available, it is similar to the standard VAE (see Eq.~\ref{eq:ss:unsup}, Sec.~\ref{sec:preliminaries}). 
Accurately, given the image $\bfx$, the Encoder estimates the posterior distribution $q_{\phi}(\poseyv, \bfz|\bfx)$, where both appearance $\bfz$ and pose $\poseyv$ are assumed to be independent given the image $\bfx$. 
Then, pose $\poseyv$ and appearance $\bfz$ are sampled from the posterior, using the reparametrization trick~\cite{kingma2013auto}, and passed to the Decoder to generate a reconstructed image. 
Finally, the unsupervised loss function minimised during training is composed of the L1-norm reconstruction loss $L1(\bfx, G(\poseyv,\bfz))$;
the KL-divergences, which act as regularisers, between the posterior and the prior distributions, $\kl[q_{\phi}(\poseyv|\bfx)|p(\poseyv)]$ and $\kl[q_{\phi}(\bfz|\bfx)|p(\bfz)]$; 
and the cross-entropy Discriminator loss (Eq.~\ref{eq:gan}, Sec.~\ref{sec:preliminaries}).

In the \textit{supervised case}, when the pose label is available, the KL-divergence between the posterior pose distribution and the pose prior, $\kl[q_{\phi}(\poseyv|\bfx)|p(\poseyv)]$, is replaced with a regression loss between the estimated pose and the given label (see Eq.~\ref{eq:ss:sup}, Sec.~\ref{sec:preliminaries}). 
Now, only the appearance $\bfz$ is sampled from the posterior distribution and passed to the Decoder, along with the ground-truth pose label.
Finally, the supervised loss function minimised during training is composed of the L1-norm reconstruction loss, the KL-divergence over the appearance distribution, the regression loss over the pose vector, and the cross-entropy Discriminator loss.
In this case, gradients are not backpropagated from the Decoder to the Encoder, through the pose posterior distribution, since pose was not estimated.

In both \textit{unsupervised} and \textit{supervised} cases, the Mapper module, which is trained \textit{offline}, is used to map the pose-vector $\poseyv$ in the latent space to a dense heatmap representation $\poseyh$, as illustrated in Fig.~\ref{fig:semi-vae_arch}.

\begin{figure}[h]
\vspace{-1em}
\centering
\includegraphics[scale=.4]{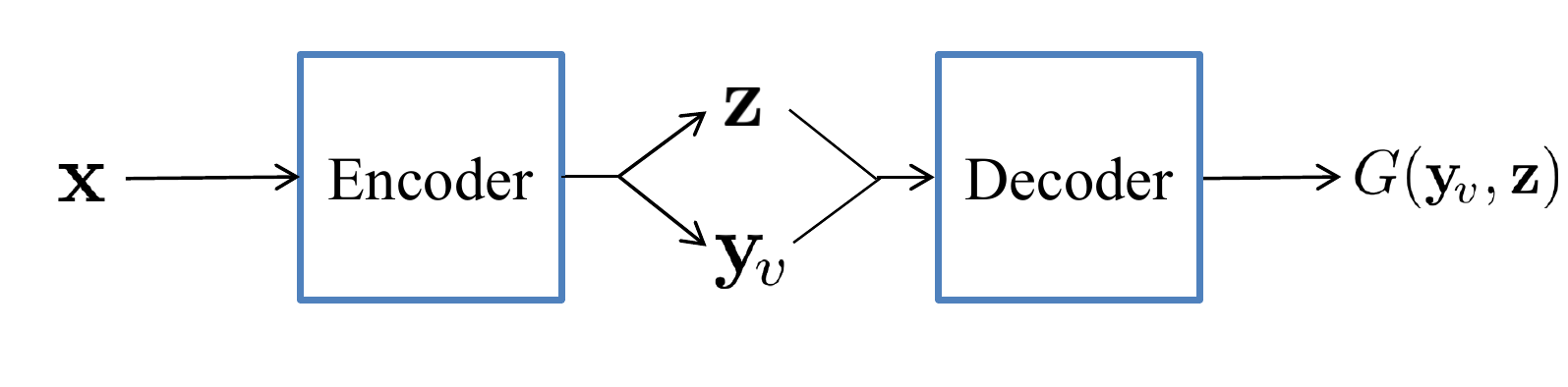}
  \captionsetup{aboveskip=-1em,belowskip=1em,justification=centering}
\caption{Semi-DGPose reconstruction at test time.\label{fig:reconstruction_diag}}
\vspace{-2em}
\end{figure}

\paragraph{\textbf{Reconstruction.}} At test time, only an image $\bfx$ is given as input, and the reconstructed image $G(\poseyv,\bfz)$ is obtained from the Decoder, as illustrated in Fig.~\ref{fig:reconstruction_diag}.
In the reconstruction process, \textit{direct manipulation} of the pose representation $\poseyv$ allows image generations with varying body poses and sizes while the appearance is kept the same.

\begin{figure}[h]
\centering
\includegraphics[scale=.4]{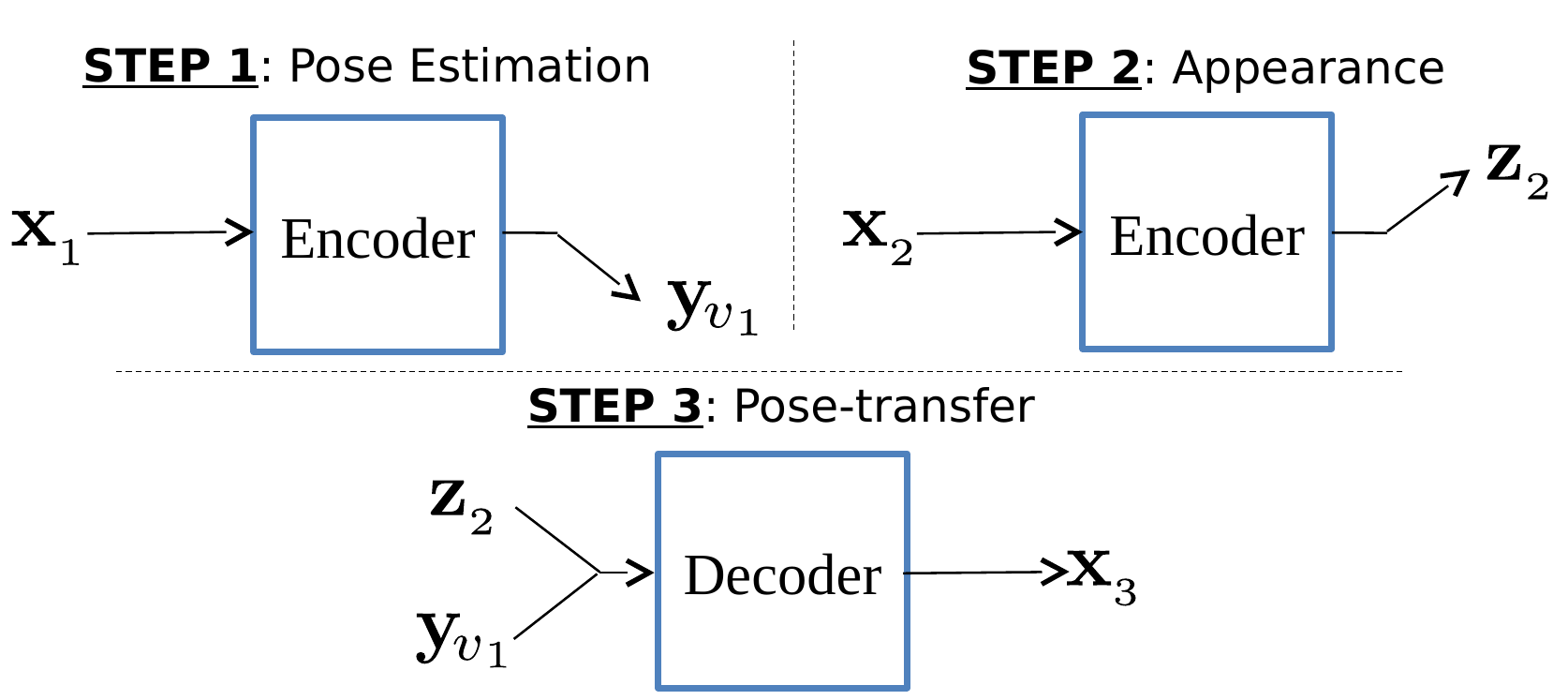}
  \captionsetup{aboveskip=-1em,belowskip=1em,justification=centering}
\caption{Semi-DGPose indirect pose-transfer at test time.\label{fig:indirect_posetransfer_diag}}
\vspace{-1em}
\end{figure}

\paragraph{\textbf{Indirect Pose-transfer.}} Our method allows us to do \textit{indirect pose-transfer} without specific training for such a task.
As illustrated in Fig.~\ref{fig:indirect_posetransfer_diag}, an image $\bfx_{_1}$ is first passed through the Encoder network, from which the target pose $\poseyva$ is estimated and kept.
In the second step, another image $\bfx_{_2}$ is propagated through the Encoder, from which the appearance encoding $\bfz_{_2}$ is kept.
Finally, $\bfz_{_2}$ and $\poseyva$ are jointly propagated through the Decoder, and an image $\bfx_{_3}$ is reconstructed, containing a person in the pose $\poseyva$ estimated from the first image, but with the appearance $\bfz_{_2}$ defined by the second image.
This is a novel application that our approach enables. In contrast to the prior art, our network neither relies on any external pose estimator nor on conditioning labels to perform pose-transfer.

\begin{figure}[h]
\centering
\includegraphics[scale=.4]{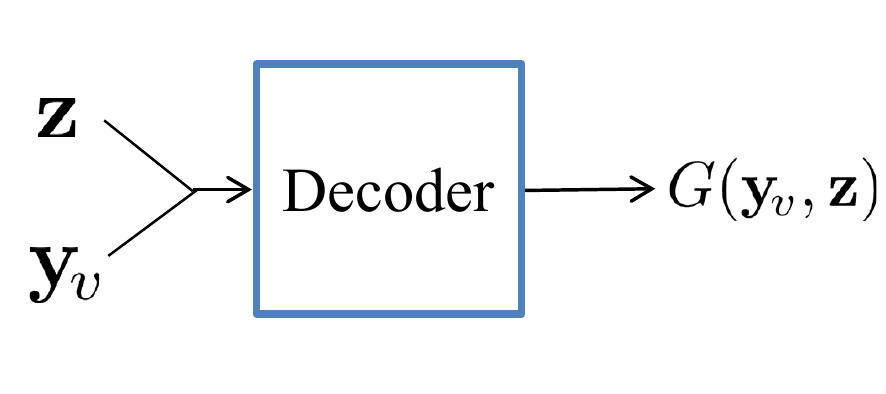}
  \captionsetup{aboveskip=-1em,belowskip=1em,justification=centering}
\caption{Semi-DGPose sampling at test time.\label{fig:sampling_diag}}
\end{figure}

\paragraph{\textbf{Sampling.}} When no image is given as input, we can jointly or separately sample pose $\poseyv$ and appearance $\bfz$ from the posterior distribution.
They may be sampled at the same time, or one may be kept fixed while the other distribution is sampled.
In all cases, the encodings are passed through the Decoder network to generate a corresponding RGB image, as illustrated in Fig.~\ref{fig:sampling_diag}.

\begin{figure}[h]
\centering
\includegraphics[scale=.4]{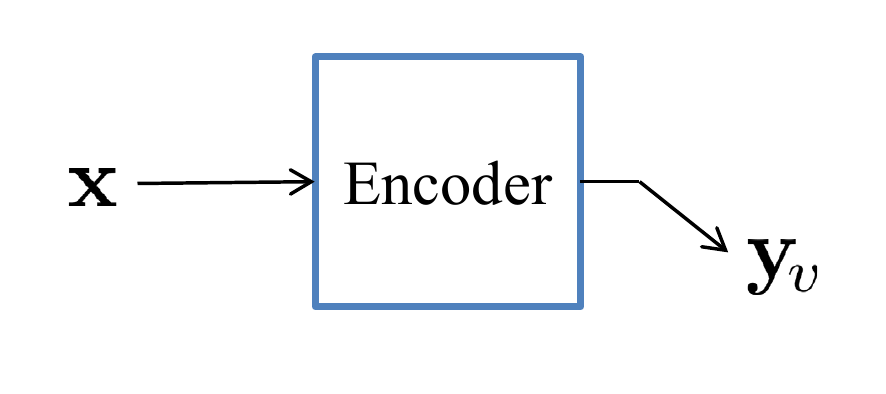}
  \captionsetup{aboveskip=-1em,belowskip=2em,justification=centering}
\caption{Semi-DGPose pose estimation at test time.\label{fig:pose_estimation_diag}}
\end{figure}

\paragraph{\textbf{Pose Estimation.}} One of the main differences between our approach and the prior art is the ability of our model to estimate human-body pose as well.
In this case, as illustrated in Fig.~\ref{fig:pose_estimation_diag}, given an input image $\bfx$, it is possible to perform pose estimation by regressing to the pose representation vector $\poseyv$.
Thus, the appearance encoding $\bfz$ is disregarded, and the Decoder, Mapper, and Discriminator networks are not used.

\section{Experiments and Results}
\label{sec:experiments}

We have performed a large number of experiments to evaluate our models. 
In this section, we present the datasets, metrics, and training hyper-parameters used in our work.
Finally, quantitative and qualitative results show the effectiveness and novelty of our Conditional-DGPose and Semi-DGPose architectures.

\subsection{Human3.6M Dataset}
Human3.6M~\cite{h36m_pami} is a widely used benchmark for human body analysis.
It contains $3.6$ million images acquired by recording 5 female and 6 male actors performing a diverse set of motions and poses corresponding to 15 activities, under 4 different viewpoints.
We followed the standard protocol and used sequences of 2 out of 11 actors as our test set, while the rest of the data was used for training.
We use a subset of 14 (out of 32) body joints represented by their $(x,y)$ 2D image coordinates as our ground-truth data, neglecting minor body parts (\eg fingers).
Due to the high frequency of video acquisition (50Hz), there is a considerable level of practically redundant images.
Thus, out of images from all 4 cameras, we subsample frames in time, producing subsets for training and testing, with $317,989$ and $1,280$ images, respectively.
All the original images have a resolution of $1000 \times 1000$ pixels.

\subsection{ChictopiaPlus Dataset}
ChictopiaPlus~\cite{LassnerPG17} is an extension of the Chictopia dataset~\cite{liang2015deep}.
It augments the original per-pixel annotations for body parts with pose annotation~\cite{insafutdinov2016deepercut}, 3D shape~\cite{loper2015smpl}, and facial segmentation.
In contrast to the Human3.6M dataset, in which each actor always wears the same outfit, it contains $23,011$ training, $2,913$ validation, and $2,873$ testing images of segmented people (without background) dressed in a great variety of clothes.
All the images have an original resolution of $286 \times 286$ pixels.

\subsection{DeepFashion Dataset}
The DeepFashion dataset (In-shop Clothes Retrieval Benchmark)~\cite{liu2016deepfashion} consists of 52,712 images of people in a variety of clothing and poses.
We follow Ma~\etal~\cite{ma2017}, using their joints' annotations obtained with an off-the-shelf pose estimator~\cite{cao2017realtime}, and divide the dataset into training (44,950 images) and testing (6,560 images) subsets.
Images with wrong pose estimations were suppressed and all original images have $256\times256$ pixels. 
Importantly, we aim to learn a complete generative model of people in images, which is significantly more complex, compared to models focusing on a particular task, such as pose-transfer.
For this reason, we use images individually in our training set, instead of employing pairs of images of the same person as in~\cite{ma2017,deformable_gans_for_pose_generation}.

\subsection{Metrics\label{sec:metrics}}
\begin{figure}[ht]
\centering
\begin{tabular}{c}
\tiny\textbf{\makebox[1.5em][l]{}\textsf{ORIGINAL}\makebox[2em][l]{}\textsf{RECONST.}}\hfill\textbf{\textsf{ORIGINAL}\makebox[2em][l]{}\textsf{RECONST.}}\makebox[1em][l]{}\vspace{-1.5em}\\
\subfloat[]{\includegraphics[width=0.35\linewidth]{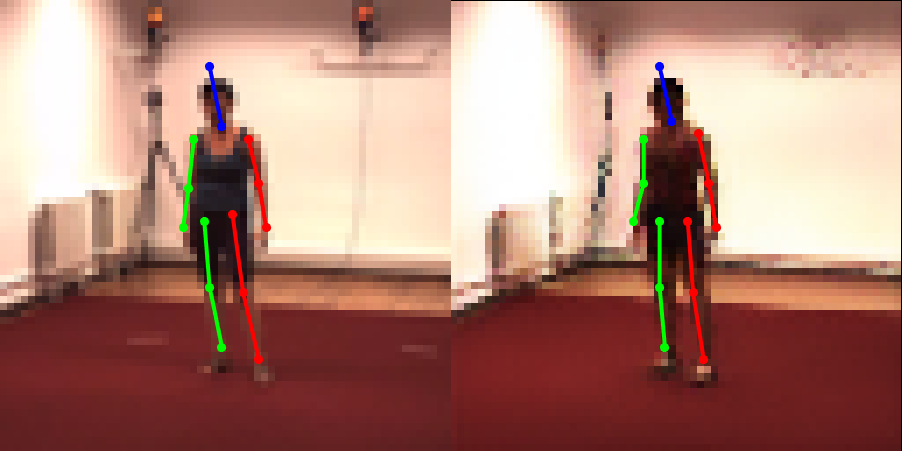}}\hspace{1em}
\subfloat[]{\includegraphics[width=0.35\linewidth]{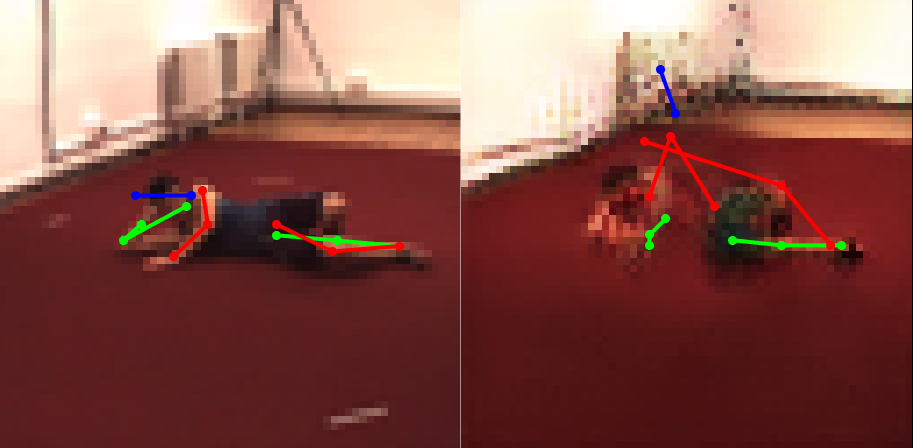}}
\end{tabular}
\caption{\textbf{Accuracy of the reconstructed poses.} Samples illustrating best and worst pose reconstructions on the Human3.6M dataset. 
Each pair of images shows the pose estimation over the original image (left) and the reconstructed image (right).
Lines connect the estimated joints for visualisation purposes. Right limbs, left limbs, and head are shown, respectively, by green, red and blue lines.
(a) It illustrates the best reconstructed poses, with PCK@0.5 = 1.00.
(b) It illustrates the worst reconstructed poses, with PCK@0.5 = 0.00.
All images are $64 \times 64$ pixels.}
\label{fig:metric_accuracy_pose}
\vspace{-2em}
\end{figure}

Quantitative evaluation of generative models is inherently difficult~\cite{Theis2016a}. 
Since our models explicitly represent \textit{appearance} and \textit{body pose} as separate variables, we evaluate their performance w.r.t. three different aspects.
i) \textbf{Image quality} of reconstructions is evaluated using the standard Peak Signal-to-Noise Ratio (PSNR) and Structural Similarity Index (SSIM) metrics~\cite{wang2004image}.
ii) \textbf{Accuracy of the reconstructed poses} is evaluated using a protocol introduced by us as follow. 
To set a common ground for comparing an original test set, with a reconstructed one, we start using a well-established (discriminative) human pose estimator~\cite{newell2016}, and initially estimating all 2D poses in the original test set.
In our protocol, we assume that such estimations are the \textit{ground-truth} poses of the test set.
Subsequently, we apply the same discriminative estimator over the reconstructed test images, produced by the trained generative models.
Finally, we use of the Percentage of Correct Keypoints (PCK) metric~\cite{yang2011articulated}, which computes the percentage of 2D joints correctly located by a pose estimator, given the \textit{ground-truth} and a normalised distance threshold corresponding to the size of the person's torso. 
Thus, we assume that any degradation in the PCK metric is caused by imperfections on the reconstructed images, since a PCK score of 100\% would correspond to having all the estimated joints, in the original and the reconstructed images, at the same locations, up to the distance threshold.
We illustrate this metric in \figref{fig:metric_accuracy_pose}.
iii) \textbf{Accuracy of pose estimation}, obtained by the Semi-DGPose model, is measured using the PCK metric with \textit{real} 2D annotated labels as ground-truths.

\subsection{Training\label{sec:hyper-param}}
All models were trained with mini-batches consisting of 64 images.
We used the Adam optimiser~\cite{kingma2014adam} with an initial learning rate set to $10^{-4}$.
The weight decay regulariser was set to $5\times10^{-4}$.
Network weights were initialised randomly for fully-connected layers and with robust initialisation~\cite{he2015delving} for convolutional and transposed-convolutional layers.
Except when stated differently, for all images and all models, we used a $64\times64$ pixels crop, centring the person of interest.
We did not use any form of data augmentation or preprocessing except for image normalisation to zero mean and unit variance.
All models were implemented in Caffe~\cite{jia2014caffe}, and all experiments ran on an NVIDIA Titan X GPU.


\begin{figure}[hb]
  \centering
  \includegraphics[width=0.75\linewidth]{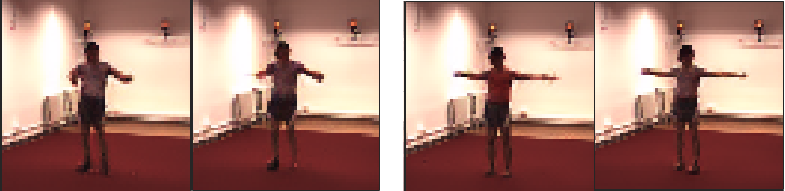}
  \subfloat[]{\hspace{-35em}}
  \subfloat[]{\hspace{-26em}}
  \subfloat[]{\hspace{-15em}}
  \subfloat[]{\hspace{-6em}}
  \caption{\textbf{Reconstructed images, obtained with each one of the four representations of human pose evaluated:} (a) 2D vector, (b) 2D vector extended, (c) heatmaps and (d) heatmaps extended.
  We highlight the difficult for capturing the spatial extent of some body parts, particularly extremities far from the torso, when the vector representations are adopted.
  In this example, the use of joints' heatmaps is already sufficient to improve the reconstruction. 
  However, the extended version (with rigid parts and body) turns the model more robust to more complex poses, since the 14 joints are fairly sparse.} 
  \label{fig:reconst_comparison}
\end{figure}

\subsection{Conditional-DGPose\label{sec:cvae_experiments}}
As mentioned earlier (Sec.~\ref{sec:approach}), the Conditional-DGPose is taken by us as an intermediate step in the investigation towards our Semi-DGPose model.
To better evaluate and understand its capabilities, we start our experiments by validating it qualitatively with the Human3.6M benchmark, since this dataset is composed of images in a controlled environment.
Initially, in Sec.~\ref{sec:exp_pose_rep}, we evaluate different pose representations, with the best performance presented by the heatmap representation.
In Sec.~\ref{sec:exp_cond_h36m}, we show the effectiveness of the Conditional-DPGose architecture, illustrating \textit{reconstruction} and \textit{sampling} tasks.
Besides that, we particularly stress the effects of pose manipulation, by performing \textit{pose-transfer} and \textit{hallucinating} multiple people in a variety of unseen or even unrealistic poses, still on the Human3.6M dataset.
After that, we present qualitative and quantitative results on the ChictopiaPlus dataset~\cite{LassnerPG17}.
The Conditional-DGPose outperforms the closest related comparable baseline, the ClothNetBody~\cite{LassnerPG17}, achieving state-of-the-art results on the ChictopiaPlus.
Finally, qualitative and quantitative experiments on the DeepFashion dataset~\cite{liu2016deepfashion} are shown.
On this dataset, our baseline is the image-to-image translation architecture by Ma \etal~\cite{ma2017}, which is trained on pairs of images showing the same person in different poses.
Although our Conditional-DGPose method tackles a significantly more complex problem, \ie learning a generative model and its latent representation in the high-dimensional image space, instead of mapping one image to another, it presents reasonable results in comparison with the ones from~\cite{ma2017}.

\subsubsection{Pose Representation\label{sec:exp_pose_rep}}

We perform experiments with the two pose representations mentioned in Sec.~\ref{sec:pose_rep} and with their respective extensions.
We executed end-to-end training with the Conditional-DGPose architecture, which converged in approximately 15 epochs.
The qualitative evaluation was performed by the inspection of the reconstructed images, shown in Fig.~\ref{fig:reconst_comparison}.
As can be observed, the vector representations, even the extended one, fail to capture some parts of the body.
This problem is particularly evident concerning the extremities of the limbs.
On the other hand, the additional heatmaps for rigid parts and whole body have shown a positive impact in the reconstructions.
The quantitative measurements, shown in Tab.~\ref{table:representations}, support our qualitative evaluation.
In all experiments, the heatmaps had the same dimension of the images ($64 \times 64$).

\begin{table}[h]
{\footnotesize
\centering
\begin{tabulary}{\textwidth}{@{}lCCC}
\hline
\textbf{Pose representation} & \textbf{L1-Norm}  \\
\hline
2D vector (14 joints) & 14.52  \\
2D vector extended (28 joints)  & 13.91   \\
\hline
Heatmaps (14 joints) & 13.55 \\
Heatmaps extended\\ $\llcorner$ (14 joints + 9 rigid parts + 1 whole body) & \textbf{\underline{13.41}}  \\
\hline
\end{tabulary}
\caption{\footnotesize Average reconstruction errors obtained with the Conditional-DGPose architecture using L1-norm for our validation set.
}
\label{table:representations}
}
\end{table}


\subsubsection{Conditional-DGPose Results on Human3.6M\label{sec:exp_cond_h36m}}

\begin{figure}[h]
\centering
\tiny\textbf{\textsf{\makebox[6em][l]{} JOINTS \makebox[2.5em][l]{} RIGID \makebox[4em][l]{} BODY \hfill ORIGINAL \makebox[2em][l]{} RECONST.\makebox[5em][l]{}}}\\
\vspace{-1.5em}
\subfloat{\includegraphics[width=.8\linewidth]{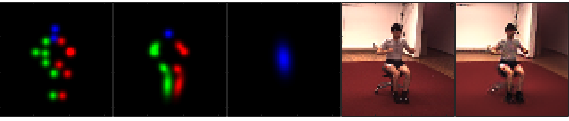}}\hspace{.1em}\\
\vspace{-1em}
\renewcommand{\thesubfigure}{a}
\subfloat{\includegraphics[width=.8\linewidth]{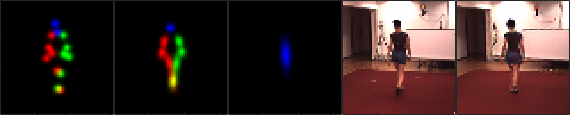}}\hspace{.1em}
\renewcommand{\thesubfigure}{b}
\caption{\textbf{Reconstructions on Human3.6M.} From the left to right columns we have: joints, rigid parts and body heatmaps; original image and finally, the reconstructed image.
In the heatmaps, right parts are shown in green, left parts in red and central parts in blue. Human3.6M images are $64 \times 64$ pixels.}
\label{fig:qualitative_recons}
\end{figure}

\begin{figure}[h]
\subfloat{\includegraphics[width=1.0\linewidth]{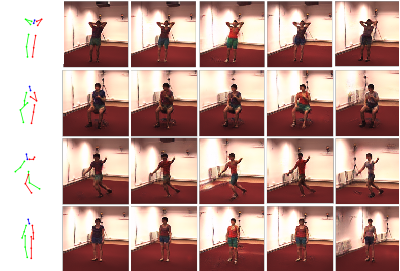}}
\caption{\textbf{Sampling on Human3.6M.} Results obtained by randomly changing pose and appearance independently.}
\label{fig:sampling_h36m}
\end{figure}

Initially, in Fig.~\ref{fig:qualitative_recons}, we show our heatmap pose representation along with reconstructions, to demonstrate that realistic images with accurate poses can be generated.
Furthermore, we illustrate \textit{sampling} in~\figref{fig:sampling_h36m}, in which the separation between pose and appearance is made evident by the independent change of each variable.

\begin{figure*}[ht]
  \centering
  \includegraphics[width=\linewidth]{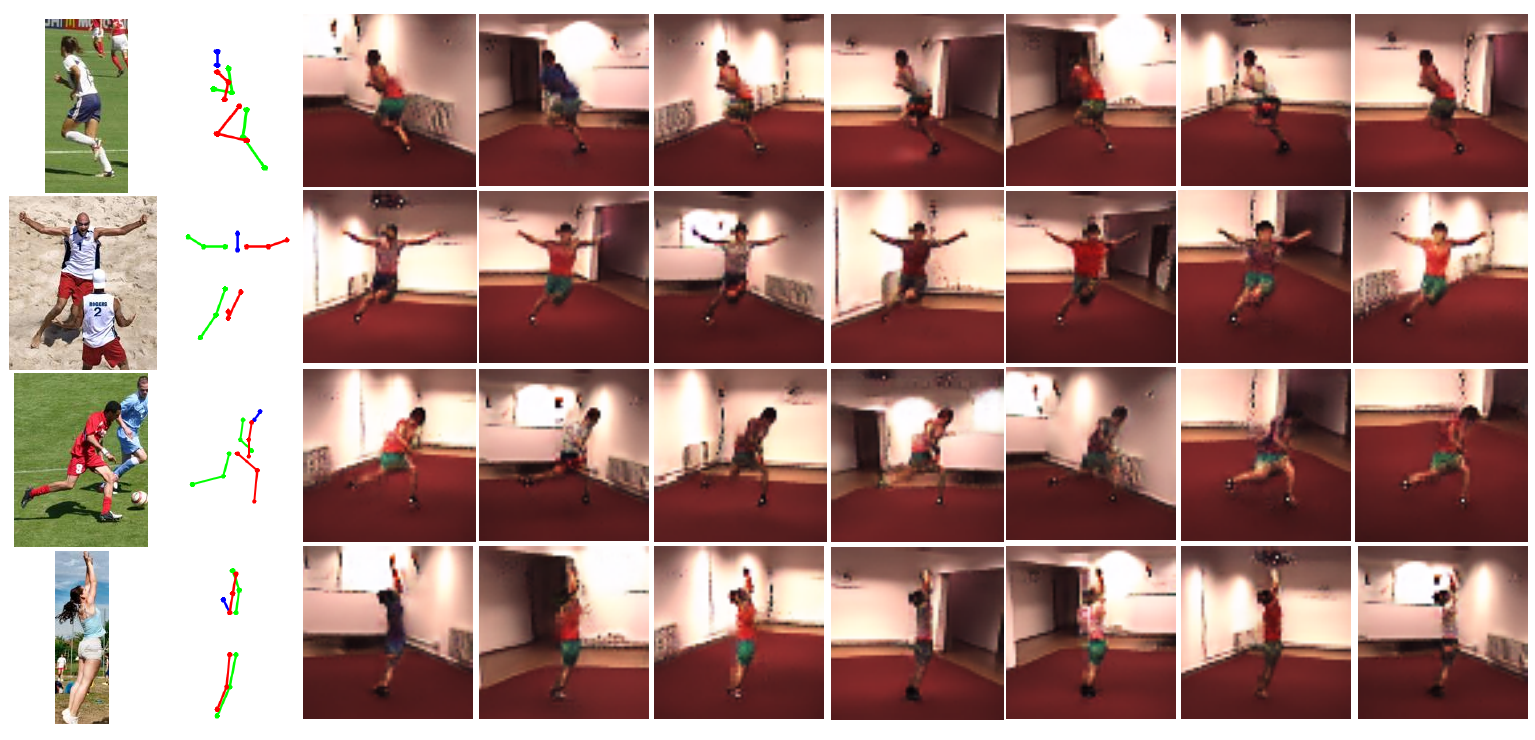}
  \captionsetup{skip=-0.5em}
  \caption{\textbf{Cross-domain pose-transfer on Human3.6M.} Here we illustrate the
  \textit{pose-transfer} capability of our Conditional-DGPose.
  On the leftmost column, we show test images from the LSP
  dataset~\cite{Johnson10}, along with their corresponding ground-truth 2D pose
  annotations, composed of 14 joints.
  These are taken as conditioners (\textit{target-poses}) on our model for the generation of the reconstructions, shown from the third
  to the rightmost column.
  As can be observed, the \textit{target-poses} are transferred to the output
  images, while the latter maintain their original appearances.
  We highlight the fact that neither the LSP images nor their poses were part of
  the training set.}
  \label{fig:pose-transfer}
\end{figure*}

Next, we stress the pose-transfer and compositionality capabilities of the model, pushing it beyond what is usually done in related methods.
Regarding \textit{pose-transfer}, we demonstrate the capability of our model to learn pose and appearance as separate variables which allows direct control over the two at test time.
To this end, we generate images in which we maintain the appearance of the input image, yet the generated person is ``moved'' into the required target pose.
The target pose may be composed manually, extracted from another image with an off-the-shelf pose estimator or provided interactively by a user.
This is illustrated in \figref{fig:pose-transfer}, in which we employ target poses from the LSP dataset~\cite{Johnson10}, that have completely different poses in a drastically different environment compared to our training set.
The quality of the generations shows that our generative model could disentangle pose and appearance and generate images with poses that do not exist in the training data.

\begin{figure}[h]
\centering
\tiny\textbf{\textsf{\makebox[2em][l]{} JOINTS \makebox[4em][l]{} ORIGINAL \makebox[3em][l]{} OUTPUT\makebox[2em][l]{}}}\vspace{-1.25em}\\
\subfloat{\includegraphics[width=0.55\linewidth]{figures/recons_multi3.png}}\hfill\\
\subfloat{\includegraphics[width=0.55\linewidth]{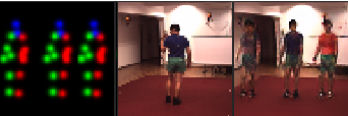}}\hfill\\
\subfloat{\includegraphics[width=0.55\linewidth]{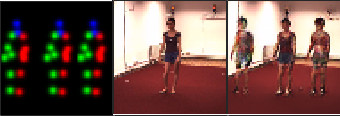}}
\caption{
\textbf{Hallucinating multiple people on Human3.6M.} 
The Conditional-DGPose model was trained with images containing only one person. 
The output images are generated keeping the appearance of the original images but conditioned to the manipulated heatmap pose representation (left).
Heatmaps of rigid parts and whole body are not shown for simplicity.}
\label{fig:multi_people}
\end{figure}

\begin{figure}[h]
\centering
\tiny\textbf{\textsf{\makebox[2em][l]{} JOINTS \makebox[3.5em][l]{} OUTPUT \hfill JOINTS \makebox[3em][l]{} ORIGINAL \makebox[3em][l]{} OUTPUT\makebox[4em][l]{}}}\vspace{-1.25em}\\
\subfloat[\label{fig:three_bodies_joints}]{\includegraphics[width=0.37\linewidth]{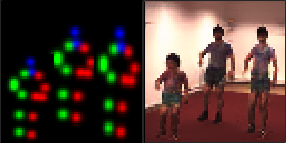}}\hfill
\subfloat[\label{fig:three_heads_new}]{\includegraphics[width=0.55\linewidth]{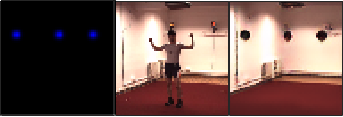}}\hfill
\vspace{1em}
\tiny\textbf{\textsf{\makebox[3.4em][l]{} JOINTS \makebox[3em][l]{} ORIGINAL \makebox[3em][l]{} OUTPUT\makebox[3em][l]{}}}\vspace{-1.25em}\\
\subfloat[\label{fig:moving_head_new}]{\includegraphics[width=0.55\linewidth]{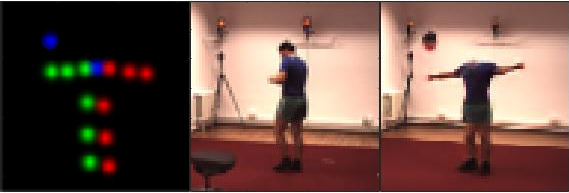}}
\caption{
\textbf{Generating ``unreal'' images on Human3.6M.} We illustrate the versatility of the model extrapolating the generation of
images to unseen scenes. ({\bf a}) Sampled image in which the pose
representation in the centre was manually translated and scaled, producing two additional bodies:
one shorter and chunkier (\textit{left}) and one taller and thinner
(\textit{right}). ({\bf b}) Reconstructed image in which all the body parts were
suppressed, except the head. ({\bf c}) Pose-transfer in which the position of
the head was manually changed, disconnecting it from the rest of the body.
Heatmaps of rigid parts and whole body are not shown for simplicity.}
\label{fig:multi_people_unreal}
\end{figure}

Concerning manipulation, we show in Fig.~\ref{fig:multi_people} how our model can be used to ``compose'' images that have never been seen in the training data.
For instance, we can generate images with multiple people in the same (replicated) pose simply by conditioning on a respective heatmap.
In fact, we can go one step further and generate an image where all people are in the same pose, but one of them is, \eg \emph{shorter} and another \emph{thinner}, as shown in Fig.~\ref{fig:three_bodies_joints}.
In an extreme case, we can even generate ``unreal'' images containing only certain body parts (\eg heads) or disconnecting them from the rest of the body, as in Figs.~\ref{fig:three_heads_new} and~\ref{fig:moving_head_new}, respectively.
Note that the training dataset is composed of only single person images.
Thus the model has never seen an image with multiple people or only some separate body parts. 
This demonstrates that the learned latent space of our model is indeed disentangled. 
To the best of our knowledge, this capability has not been demonstrated by any other work in the literature. 


\begin{figure*}[ht]
\vspace{-0.5em}
\centering
\begin{minipage}{0.3275\linewidth}
\subfloat{
\captionsetup{position=top,labelformat=empty}
\subfloat[\tiny\textbf{\textsf{ORIGINAL}}]{\includegraphics[width=.33\linewidth]{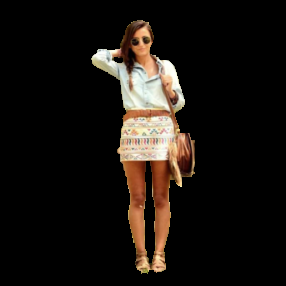}}
\subfloat[\tiny\textbf{\textsf{OURS}}]{\includegraphics[width=.33\linewidth]{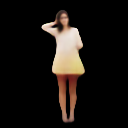}}
\subfloat[\tiny\textbf{\textsf{ClothNet-Body}}]{\includegraphics[width=.33\linewidth]{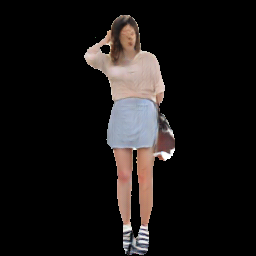}}
}
\end{minipage}
\begin{minipage}{0.3275\linewidth}
\subfloat{
\captionsetup{position=top,labelformat=empty}
\subfloat[\tiny\textbf{\textsf{ORIGINAL}}]{\includegraphics[width=.33\linewidth]{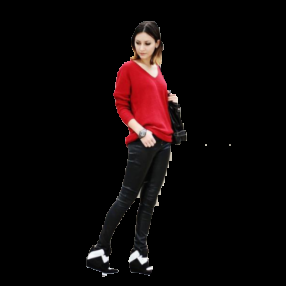}}
\subfloat[\tiny\textbf{\textsf{OURS}}]{\includegraphics[width=.33\linewidth]{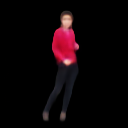}}
\subfloat[\tiny\textbf{\textsf{ClothNet-Body}}]{\includegraphics[width=.33\linewidth]{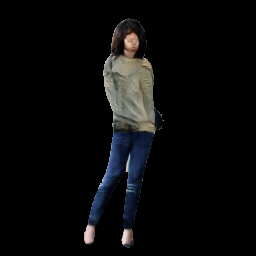}}
}
\end{minipage}
\begin{minipage}{0.3275\linewidth}
\subfloat{
\captionsetup{position=top,labelformat=empty}
\subfloat[\tiny\textbf{\textsf{ORIGINAL}}]{\includegraphics[width=.33\linewidth]{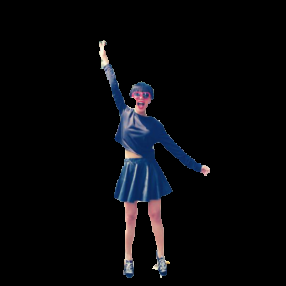}}
\subfloat[\tiny\textbf{\textsf{OURS}}]{\includegraphics[width=.33\linewidth]{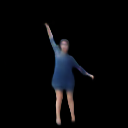}}
\subfloat[\tiny\textbf{\textsf{ClothNet-Body}}]{\includegraphics[width=.33\linewidth]{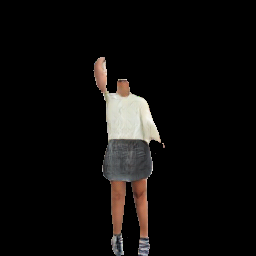}}
}
\end{minipage}\\
\vspace{-2.1em}
\begin{minipage}{0.3275\linewidth}
\subfloat{
\subfloat{\includegraphics[width=.33\linewidth]{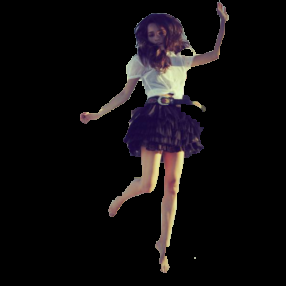}}
\subfloat{\includegraphics[width=.33\linewidth]{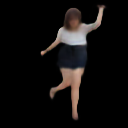}}
\subfloat{\includegraphics[width=.33\linewidth]{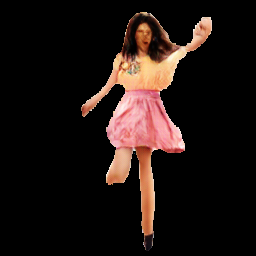}}
}
\end{minipage}
\begin{minipage}{0.3275\linewidth}
\subfloat{
\subfloat{\includegraphics[width=.33\linewidth]{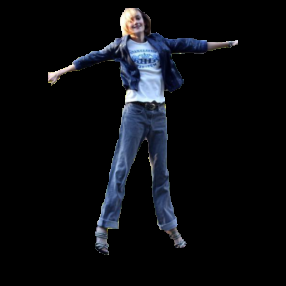}}
\subfloat{\includegraphics[width=.33\linewidth]{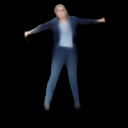}}
\subfloat{\includegraphics[width=.33\linewidth]{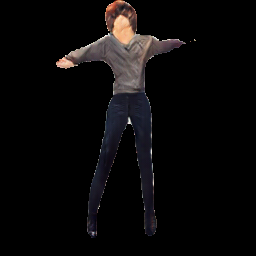}}
}
\end{minipage}
\begin{minipage}{0.3275\linewidth}
\subfloat{
\subfloat{\includegraphics[width=.33\linewidth]{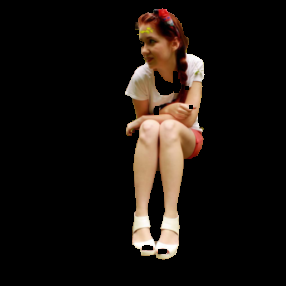}}
\subfloat{\includegraphics[width=.33\linewidth]{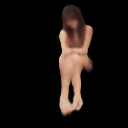}}
\subfloat{\includegraphics[width=.33\linewidth]{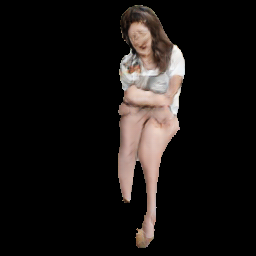}}
}
\end{minipage}
\\
\vspace{-1em}
\begin{minipage}{0.3275\linewidth}
\subfloat{
\subfloat{\includegraphics[width=.33\linewidth]{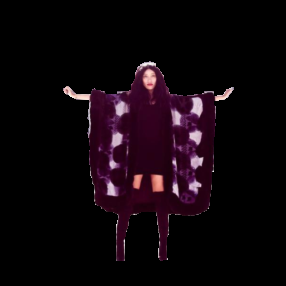}}
\subfloat{\includegraphics[width=.33\linewidth]{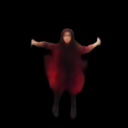}}
\subfloat{\includegraphics[width=.33\linewidth]{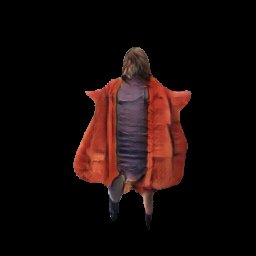}}
}
\end{minipage}
\begin{minipage}{0.3275\linewidth}
\subfloat{
\subfloat{\includegraphics[width=.33\linewidth]{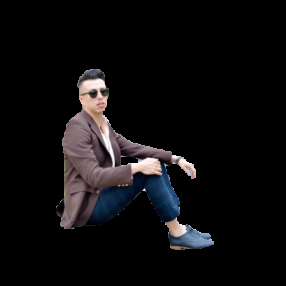}}
\subfloat{\includegraphics[width=.33\linewidth]{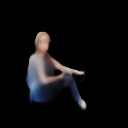}}
\subfloat{\includegraphics[width=.33\linewidth]{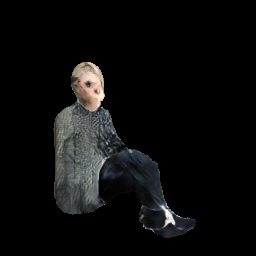}}
}
\end{minipage}
\begin{minipage}{0.3275\linewidth}
\subfloat{
\subfloat{\includegraphics[width=.33\linewidth]{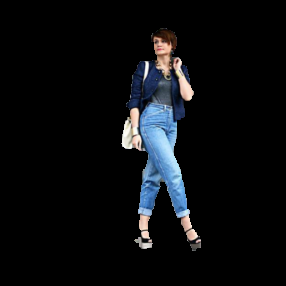}}
\subfloat{\includegraphics[width=.33\linewidth]{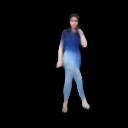}}
\subfloat{\includegraphics[width=.33\linewidth]{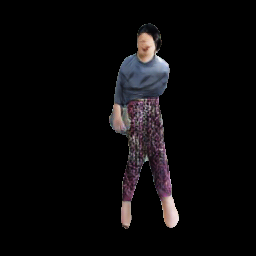}}
}
\end{minipage}
\\
\vspace{-1em}
\begin{minipage}{0.3275\linewidth}
\subfloat{
\subfloat{\includegraphics[width=.33\linewidth]{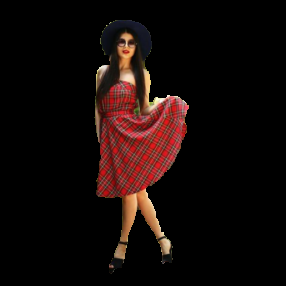}}
\subfloat{\includegraphics[width=.33\linewidth]{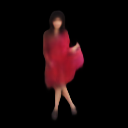}}
\subfloat{\includegraphics[width=.33\linewidth]{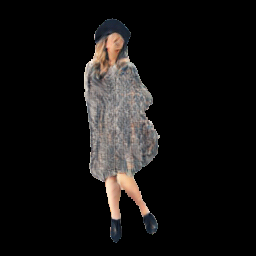}}
}
\end{minipage}
\begin{minipage}{0.3275\linewidth}
\subfloat{
\subfloat{\includegraphics[width=.33\linewidth]{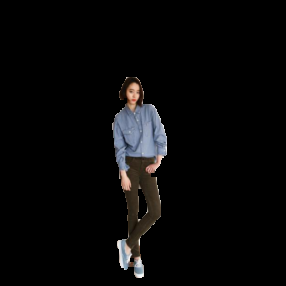}}
\subfloat{\includegraphics[width=.33\linewidth]{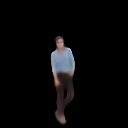}}
\subfloat{\includegraphics[width=.33\linewidth]{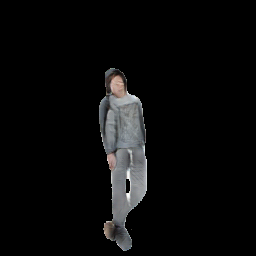}}
}
\end{minipage}
\begin{minipage}{0.3275\linewidth}
\subfloat{
\subfloat{\includegraphics[width=.33\linewidth]{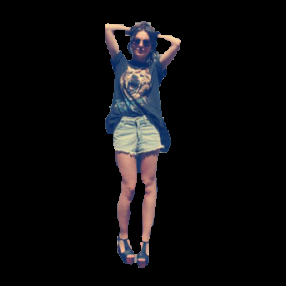}}
\subfloat{\includegraphics[width=.33\linewidth]{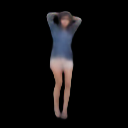}}
\subfloat{\includegraphics[width=.33\linewidth]{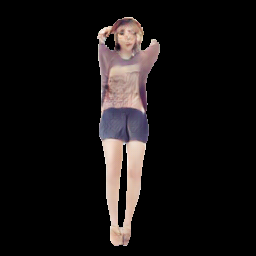}}
}
\end{minipage}
\caption{\textbf{Reconstructions on ChictopiaPlus.} In each trio of images we have, respectively: original image ($256 \times 256$),  Conditional-DGPose and ClothNet-body~\cite{LassnerPG17} reconstructions.
Notice that the images generated by our model are much closer to the originals in terms of appearance (colours).
Moreover, in general, the Conditional-DGPose captures the body parts' locations more accurately, resulting in better pose reconstructions (see Fig.~\ref{fig:pck_chictopia_vs_our}).
Best viewed if zoomed in digital version.}
\label{fig:rec_ours_vs_clothnet}
\end{figure*}

\subsubsection{Conditional-DGPose Results on ChictopiaPlus}
\begin{figure}[h]
\includegraphics[width=0.95\linewidth]{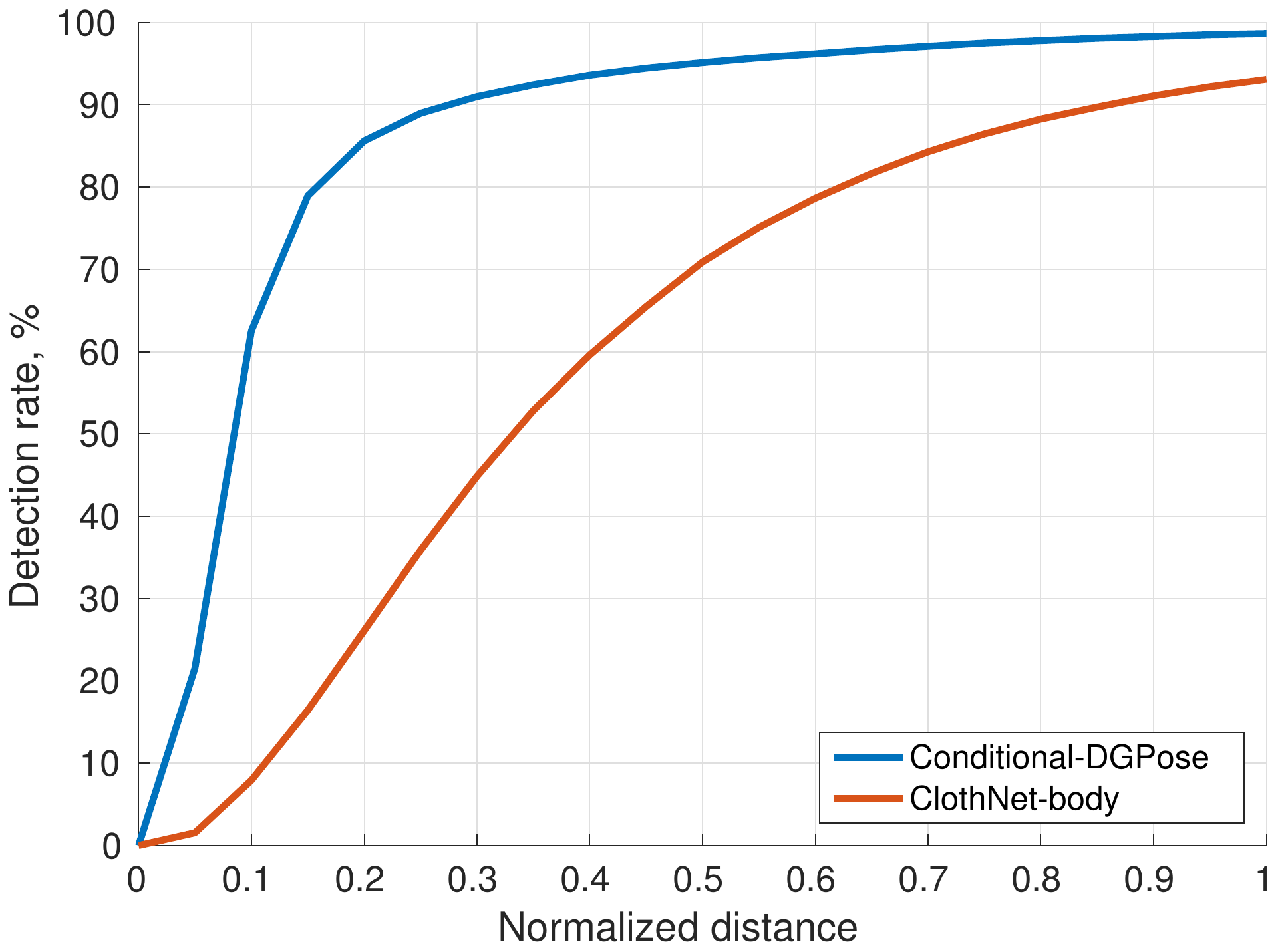}
\caption{\textbf{Accuracy of Poses on ChictopiaPlus.} The PCK scores over reconstructed images of our Conditional-DGPose (\textit{blue}) significantly outperforms the
ClothNet-body~\cite{LassnerPG17} (\textit{red}). Detection rate represents the percentage of joints correctly relocated in the reconstructions.}
\label{fig:pck_chictopia_vs_our}
\end{figure}

We compare our method with Lassner~\etal~\cite{LassnerPG17}, the closest related work from the literature.
We employ the PSNR and the SSIM metrics to evaluate image quality, and the PCK metric to provide a quantitative evaluation of pose reconstructions, as described previously (see Sec.~\ref{sec:metrics}). 
In Tab.~\ref{table:chic_image_quality}, we initially show that our method outperforms the ClothNet-body network~\cite{LassnerPG17} regarding both, the PSNR and the SSIM metrics.
Moreover, our model reports $95.14\%$ of accuracy, with PCK score at $0.5$, and again outperforms~\cite{LassnerPG17} by a large margin, which reports $70.89\%$. 
The overall PCK curve is shown in \figref{fig:pck_chictopia_vs_our}.
Finally, qualitative results are shown in \figref{fig:rec_ours_vs_clothnet}.
Our results demonstrate the good quality of our reconstructions w.r.t. image quality and the human pose.
The better performance, in comparison with~\cite{LassnerPG17}, can be particularly noticed in the extremities of body limbs, which we hypothesise as a benefit of the single stage end-to-end Conditional-DGPose model, in contrast to the multiple stages of training and testing in~\cite{LassnerPG17}.

\begin{table}[h]
\centering
\begin{tabular}{c|c|c} 
 & PSNR & SSIM \\
\hline
Conditional-DGPose & \textbf{\underline{21.33}} & \textbf{\underline{0.88}} \\
\hline
ClothNet-body~\cite{LassnerPG17} & 16.89 & 0.82\\
\end{tabular}
\caption{\textbf{Image Quality on ChictopiaPlus.} Quantitative evaluation w.r.t. image quality, showing that our method outperforms~\cite{LassnerPG17} considering both metrics, the PSNR and the SSIM.}
\label{table:chic_image_quality}
\end{table}

\subsubsection{\label{subsec:cond_dgpose_df}Conditional-DGPose Results on DeepFashion}

Here we show qualitative and quantitative experiments on the DeepFashion dataset~\cite{liu2016deepfashion}.
The baseline on this dataset is the image-to-image pose guided generation ($\textnormal{PG}^{2}$) by Ma \etal~\cite{ma2017}.
Thus, we use their same training and test sets. However, as our model is not an image-to-image translation architecture, we do not use pairs of images for training.
Instead, we use individually 44,950 training images and 6,560 test images.

\begin{figure}[ht]
\includegraphics[width=0.95\linewidth]{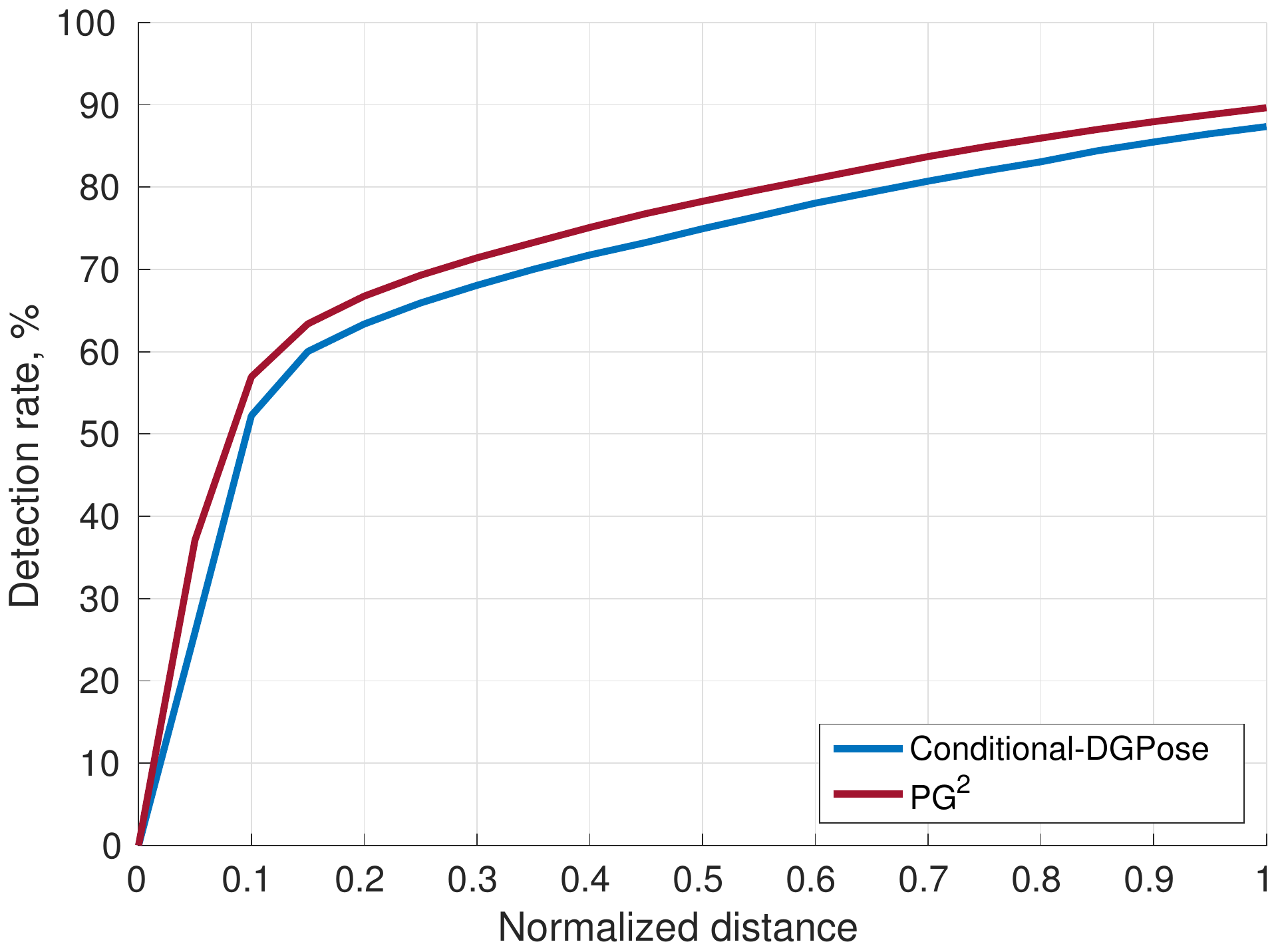}
\caption{\textbf{Accuracy of Poses on DeepFashion.} The PCK scores over reconstructed images of our Conditional-DGPose (\textit{blue}) performs only slightly below the $\text{PG}^2$ network~\cite{ma2017} (\textit{red}),
despite the fact it is tackling a significantly more complex problem than image-to-image translation. 
Detection rate represents the percentage of joints correctly relocated in the reconstructions.}
\label{fig:pck_df_vs_our_cond}
\end{figure}

Again, we employ the PSNR and the SSIM metrics to evaluate image quality, and the PCK metric to provide a quantitative evaluation of pose reconstructions, as described previously (see Sec.\ref{sec:metrics}). 
In Table~\ref{table:df_image_quality}, we initially show that even not being trained on images pairs and tackling the significantly more complex task of learning a generative model, instead of executing image-to-image translation, 
our method achieves scores only slightly below the ones by the $\text{PG}^2$ network on image reconstruction.
A similar observation can be done regarding pose reconstruction, since our model reports $74.94\%$ of accuracy, with PCK score at $0.5$, against $78.27\%$ from Ma \etal~\cite{ma2017}. 
The overall PCK curve is shown in \figref{fig:pck_df_vs_our_cond}.

\begin{figure*}[ht]
\begin{minipage}{0.335\linewidth}
\subfloat{
\subfloat{\includegraphics[width=.33\linewidth]{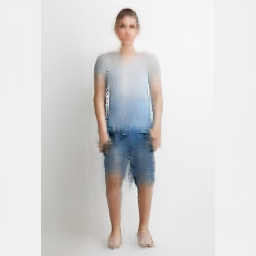}}
\subfloat{\includegraphics[width=.33\linewidth]{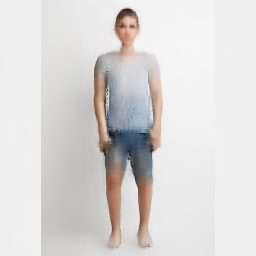}}
\subfloat{\includegraphics[width=.33\linewidth]{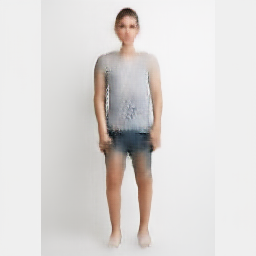}}
\subfloat{\includegraphics[width=.33\linewidth]{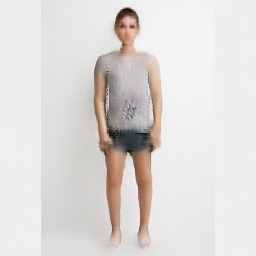}}
\subfloat{\includegraphics[width=.33\linewidth]{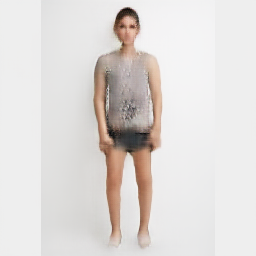}}
\subfloat{\includegraphics[width=.33\linewidth]{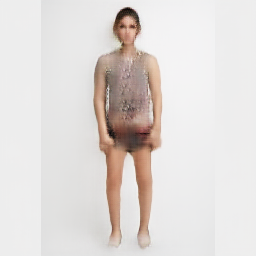}}
\subfloat{\includegraphics[width=.33\linewidth]{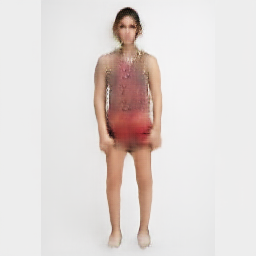}}
\subfloat{\includegraphics[width=.33\linewidth]{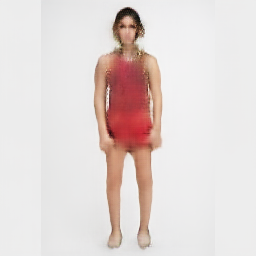}}
\subfloat{\includegraphics[width=.33\linewidth]{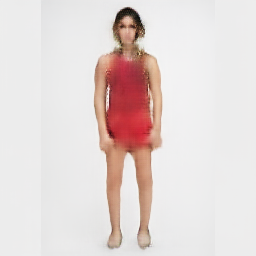}}
}
\end{minipage}
\caption{\textbf{Conditional-DGPose Appearance Manifold.} Illustration of the appearance manifold learned on the DeepFashion dataset.
We smoothly traverse the manifold for a given pose, causing changes in the visual appearance of the person in the image.
No image is used as input, only our heatmap pose representation, evidencing that a truly generative model of images was learned, in which pose and appearance are disentangled.
Best viewed if zoomed in digital version.}
\label{fig:df_manifold_condDGPose}
\end{figure*}
\begin{figure*}[ht]
\vspace{-0.5em}
\centering
\begin{minipage}{0.3275\linewidth}
\subfloat{
\captionsetup{position=top,labelformat=empty}
\subfloat[\tiny\textbf{\textsf{ORIGINAL}}]{\includegraphics[width=.33\linewidth]{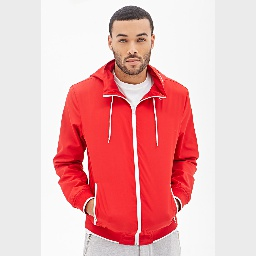}}
\subfloat[\tiny\textbf{\textsf{OUR}}]{\includegraphics[width=.33\linewidth]{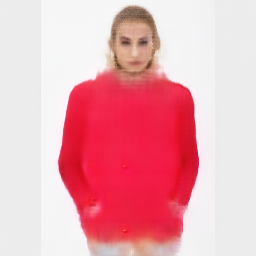}}
\subfloat[\tiny\textbf{\textsf{PG$^2$}}]{\includegraphics[width=.33\linewidth]{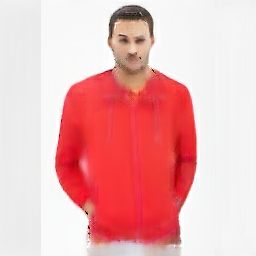}}
}
\end{minipage}
\begin{minipage}{0.3275\linewidth}
\subfloat{
\captionsetup{position=top,labelformat=empty}
\subfloat[\tiny\textbf{\textsf{ORIGINAL}}]{\includegraphics[width=.33\linewidth]{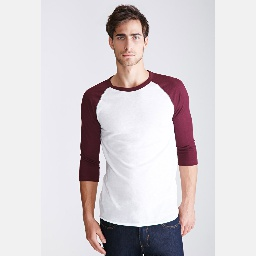}}
\subfloat[\tiny\textbf{\textsf{OUR}}]{\includegraphics[width=.33\linewidth]{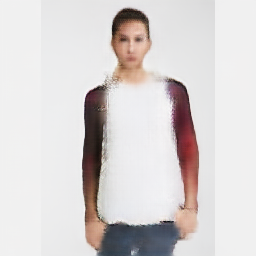}}
\subfloat[\tiny\textbf{\textsf{PG$^2$}}]{\includegraphics[width=.33\linewidth]{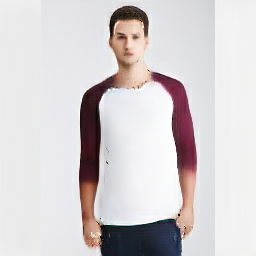}}
}
\end{minipage}
\begin{minipage}{0.3275\linewidth}
\subfloat{
\captionsetup{position=top,labelformat=empty}
\subfloat[\tiny\textbf{\textsf{ORIGINAL}}]{\includegraphics[width=.33\linewidth]{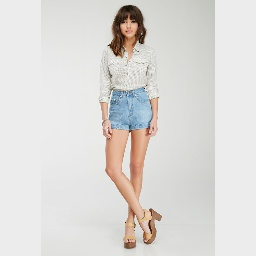}}
\subfloat[\tiny\textbf{\textsf{OUR}}]{\includegraphics[width=.33\linewidth]{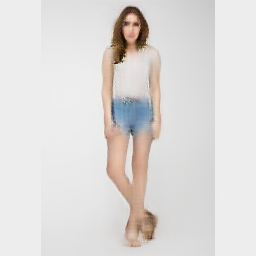}}
\subfloat[\tiny\textbf{\textsf{PG$^2$}}]{\includegraphics[width=.33\linewidth]{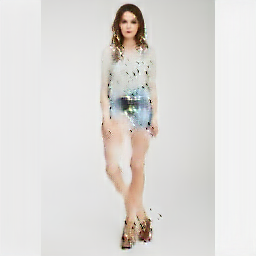}}
}
\end{minipage}\\
\vspace{-2em}
\begin{minipage}{0.3275\linewidth}
\subfloat{
\subfloat{\includegraphics[width=.33\linewidth]{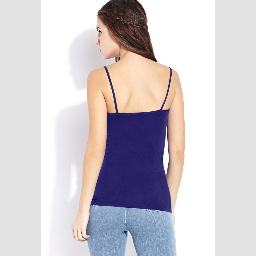}}
\subfloat{\includegraphics[width=.33\linewidth]{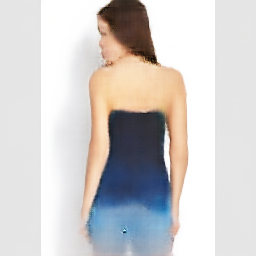}}
\subfloat{\includegraphics[width=.33\linewidth]{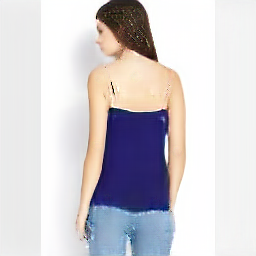}}
}
\end{minipage}
\begin{minipage}{0.3275\linewidth}
\subfloat{
\subfloat{\includegraphics[width=.33\linewidth]{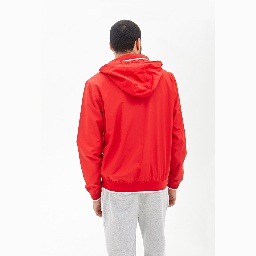}}
\subfloat{\includegraphics[width=.33\linewidth]{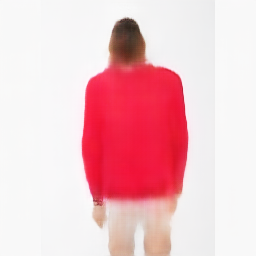}}
\subfloat{\includegraphics[width=.33\linewidth]{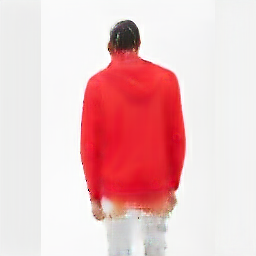}}
}
\end{minipage}
\begin{minipage}{0.3275\linewidth}
\subfloat{
\subfloat{\includegraphics[width=.33\linewidth]{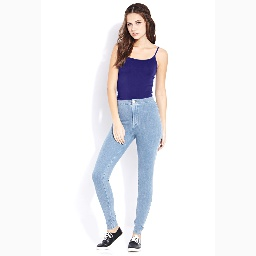}}
\subfloat{\includegraphics[width=.33\linewidth]{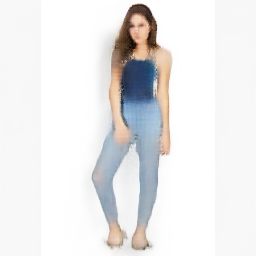}}
\subfloat{\includegraphics[width=.33\linewidth]{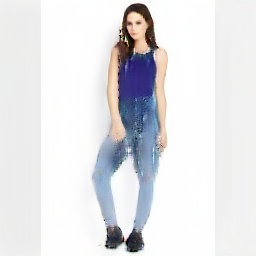}}
}
\end{minipage}\\
\vspace{-1em}
\begin{minipage}{0.3275\linewidth}
\subfloat{
\subfloat{\includegraphics[width=.33\linewidth]{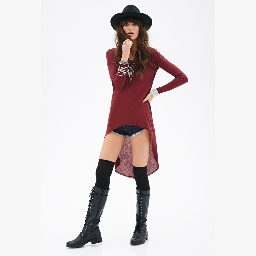}}
\subfloat{\includegraphics[width=.33\linewidth]{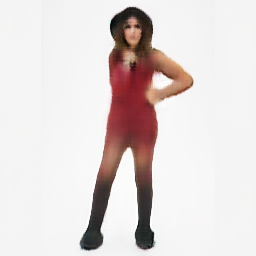}}
\subfloat{\includegraphics[width=.33\linewidth]{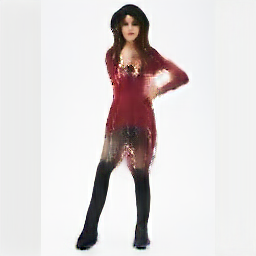}}
}
\end{minipage}
\begin{minipage}{0.3275\linewidth}
\subfloat{
\subfloat{\includegraphics[width=.33\linewidth]{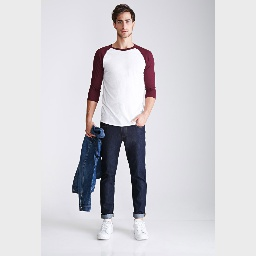}}
\subfloat{\includegraphics[width=.33\linewidth]{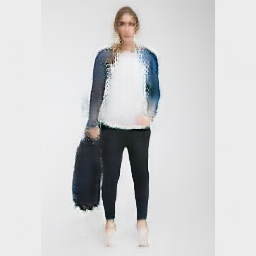}}
\subfloat{\includegraphics[width=.33\linewidth]{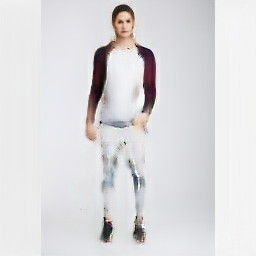}}
}
\end{minipage}
\begin{minipage}{0.3275\linewidth}
\subfloat{
\subfloat{\includegraphics[width=.33\linewidth]{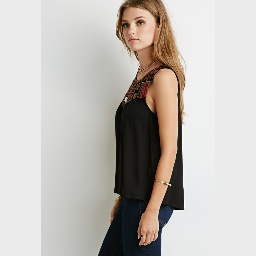}}
\subfloat{\includegraphics[width=.33\linewidth]{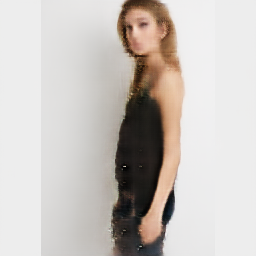}}
\subfloat{\includegraphics[width=.33\linewidth]{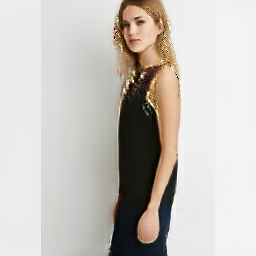}}
}
\end{minipage}\\
\caption{\textbf{Reconstructions on DeepFashion }. In each trio of images, we have, respectively: original image,  Conditional-DGPose and $\textnormal{PG}^{2}$~\cite{ma2017} reconstructions.
All images have $256\times256$ pixels.
Although tackling a more complex task than~\cite{ma2017}, our results are still reasonable.
Best viewed if zoomed in digital version.}
\label{fig:rec_conditional_vs_ma}
\end{figure*}

Concretely, the learning of a full generative model, instead of image-to-image translation, allows for the execution of tasks, such as sampling from the learned latent space, which are just not feasible with architectures purely trained on image pairs.
To illustrate this, in~\figref{fig:df_manifold_condDGPose} we traverse the appearance manifold learned on the DeepFashion dataset.
Using only our heatmap pose representation as input, for a given pose, we smoothly vary the values of the latent appearance representation, generating samples with different visual aspect for the same body posture.
Such kind of direct sampling is not feasible with the $\text{PG}^2$~\cite{ma2017} architecture.

Finally, the Conditional-DGPose performs 3.06\% and 4.82\% worse than the $\text{PG}^2$~\cite{ma2017} regarding, respectively, the PSNR and the SSIM metrics (see Table~\ref{table:df_image_quality}).
Despite that, it produces reasonable results in comparison with the ones from~\cite{ma2017}.
A qualitative evaluation is shown in Fig.~\ref{fig:rec_conditional_vs_ma}.

\begin{table}[ht]
\centering
\begin{tabular}{c|c|c} 
 & PSNR & SSIM \\
\hline
Conditional-DGPose & 18.38 & 0.79 \\
\hline
$\text{PG}^2$~\cite{ma2017} & \textbf{\underline{18.96}} & \textbf{\underline{0.83}}\\
\end{tabular}
\caption{\textbf{Image Quality on DeepFashion.} Quantitative evaluation w.r.t. image quality, showing that our method presents a performance only slightly below the baseline~\cite{ma2017},
considering both metrics, the PSNR and the SSIM, despite the fact it tackles a significantly more complex task than image-to-image translation.}
\label{table:df_image_quality}
\end{table}


\subsection{Semi-DGPose}
\label{sec:semi_experiments}
Here, we initially evaluate our Semi-DGPose model on the Human3.6M~\cite{h36m_pami} dataset. 
The Human3.6M is more suitable than both, the ChictopiaPlus and the DeepFashion, for pose estimation evaluations, since the former has joints' annotations obtained by an accurate motion capture system. 
While the two other datasets are augmented with 2D pose labels obtained using an \textit{off-the-shelf} pose estimator, consequently resulting in more errors in the ground-truth annotations.
We show quantitative and qualitative results, focusing particularly on the pose estimation and the \textit{indirect pose-transfer} capabilities, described later in this section.
Our experiments and results show the effectiveness of the Semi-DGPose method on the Human3.6M.

To show the generality of the model, we present additional results on the DeepFashion dataset.
We now use our Conditional-DGPose architecture and the image-to-image translation network $\text{PG}^2$~\cite{ma2017} as baselines, despite to their relevant differences with the Semi-DGPose.
However, to our knowledge, there are no closer related methods in the literature, \ie that simultaneously pursue the \textit{understanding} and the \textit{generation} of people directly in the image space.
Since our Conditional-DGPose method outperforms the ClothNet-body~\cite{LassnerPG17} architecture, we do not carry out a direct comparison with the latter.

\subsubsection{Semi-DGPose Results on Human3.6M \label{sec:semidgpose_h36m}}
\begin{figure}[ht]
\centering
\begin{tabular}{c}
\renewcommand{\thesubfigure}{a}
\begin{minipage}{\linewidth}
\subfloat[\label{fig:pck_alpha}]{\includegraphics[width=.95\linewidth]{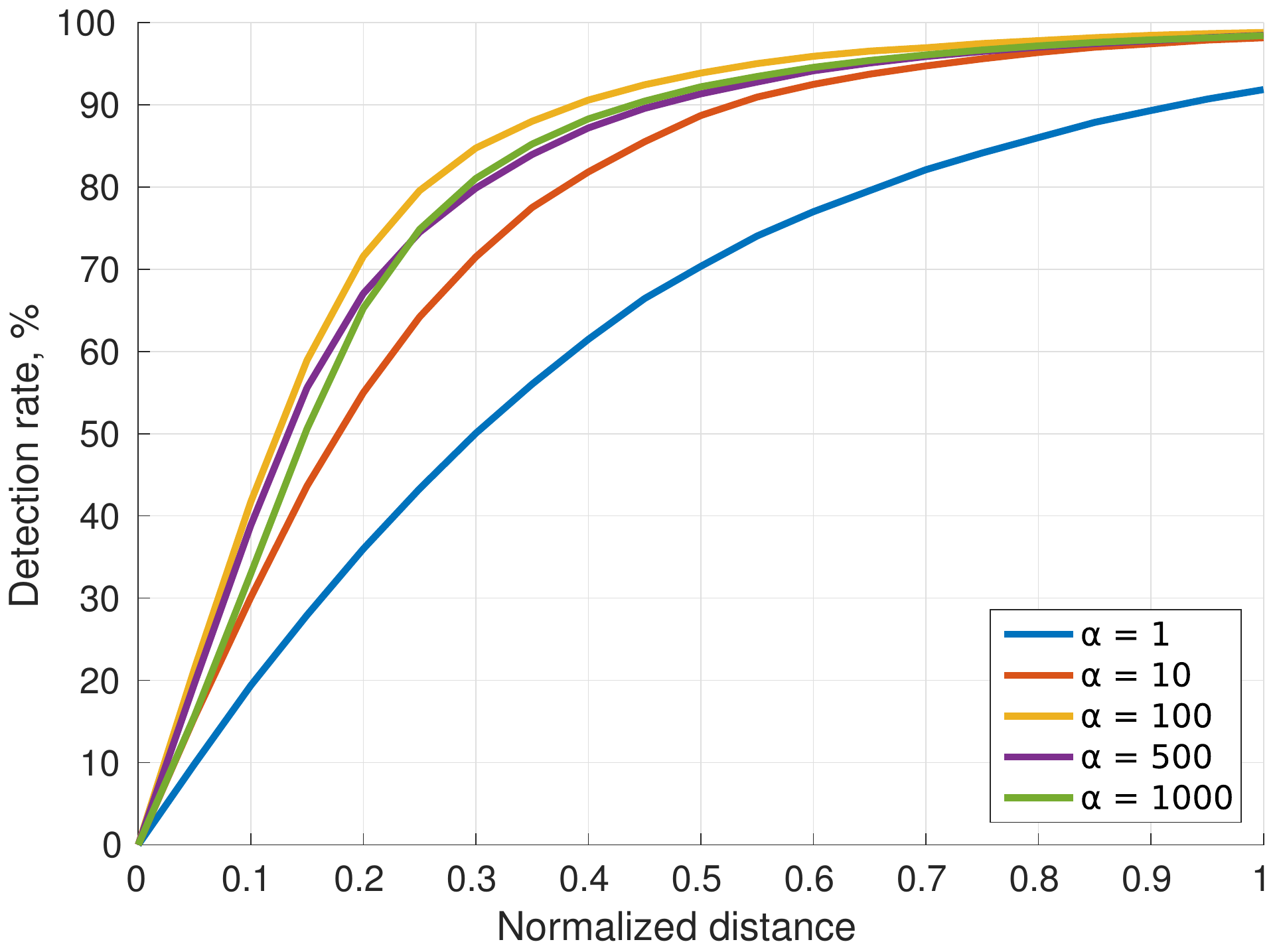}}
\vspace{1em}
\end{minipage}
\\
\renewcommand{\thesubfigure}{b}
\begin{minipage}{\linewidth}
\centering
\subfloat{\includegraphics[width=.9\linewidth]{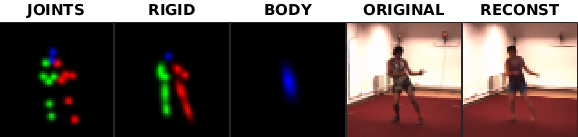}}\\
\subfloat[\label{fig:semi_reconst}]{\includegraphics[width=.9\linewidth]{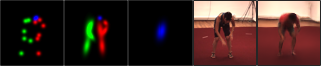}}
\end{minipage}
\end{tabular}
\caption{(a) PCK scores for the cross-validation adjustment of the regression loss weight $\alpha$. (b) Qualitative reconstructions with full supervision.}
\label{fig:semi_cross}
\end{figure}
\begin{figure}[ht]
\begin{minipage}{1.45\linewidth}
\subfloat[]{\includegraphics[width=0.126\linewidth]{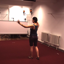}}\hspace{2em}
\subfloat[]{\includegraphics[width=0.126\linewidth]{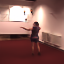}}
\subfloat[]{\includegraphics[width=0.126\linewidth]{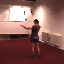}}
\subfloat[]{\includegraphics[width=0.126\linewidth]{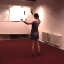}}
\subfloat[]{\includegraphics[width=0.126\linewidth]{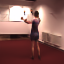}}
\end{minipage}
\caption{\textbf{Direct manipulation.} Original image ({\bf a}), followed by
reconstructions in which the person's height was changed to a percentage of the original, as: ({\bf b}) 80\%, ({\bf c}) 95\%,
({\bf d}) 105\% and ({\bf e}) 120\%. The same procedure may be applied to produce different changes in the body size and aspect ratio.}
\label{fig:direct_manipulation}
\end{figure}
To evaluate the efficacy of our model, we perform a ``relative'' comparison.
In other words, we first train our model with full supervision (\ie all data points are labelled) to evaluate performance in an ideal case and then we train the model with other setups, using labels only for $75\%$, $50\%$, and $25\%$ data points.
Such an evaluation allows us to decouple the efficacy of the model itself and the semi-supervision to see how the gradual decrease in the level of supervision affects the final performance of the method on the same validation set.

\begin{figure*}[ht]
\centering
\subfloat[PCK=92.9\%]{\includegraphics[width=0.24\linewidth]{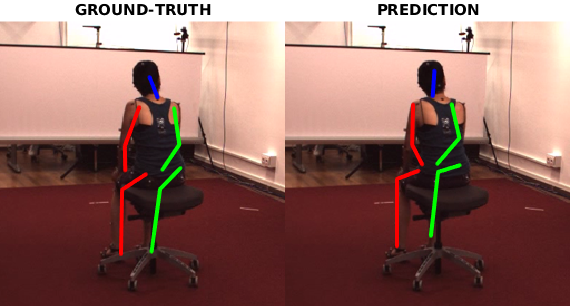}}\hspace{0.2em}
\subfloat[PCK=100.0\%]{\includegraphics[width=0.24\linewidth]{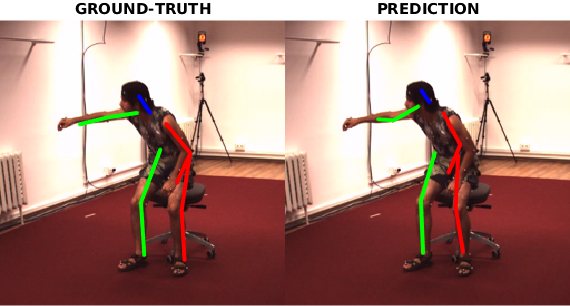}}\hspace{0.2em}
\subfloat[PCK=96.4\%]{\includegraphics[width=0.24\linewidth]{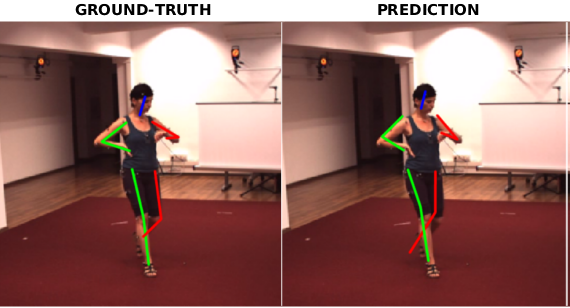}}\hspace{0.2em}
\subfloat[PCK=100.0\%]{\includegraphics[width=0.24\linewidth]{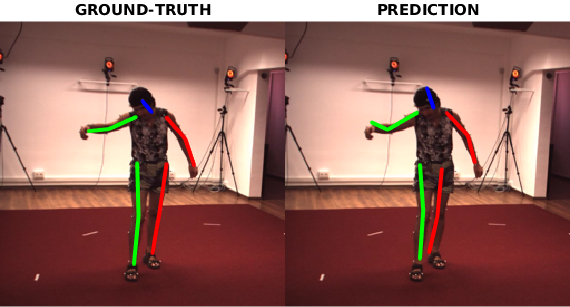}}
\caption{\textbf{Pose Estimation on Human3.6M.} Pairs of ground-truth and predicted joints superimposed on the original images.
Below each pair, we show the PCK score normalised at 0.5 times the torso size, as usual for the PCK metric. 
Such normalised distance explains the high scores despite the existence of minor differences between ground-truth and predicted positions.
Results were obtained with 100\% of supervision during training, and each pair correspond to one of the 4 cameras from the Human3.6M dataset.}
\label{fig:pose_fully_sup}
\end{figure*}

\begin{figure*}[ht]
\centering
\subfloat[\label{fig:semi_sup_rec_orig}]{\includegraphics[width=0.125\linewidth]{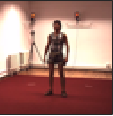}}\hfill
\subfloat[\label{fig:semi_sup_rec_joints}]{\includegraphics[width=0.125\linewidth]{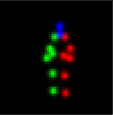}}\hfill
\subfloat[\label{fig:semi_sup_100}]{\includegraphics[width=0.126\linewidth]{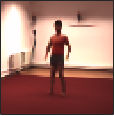}}
\subfloat[\label{fig:semi_sup_75}]{\includegraphics[width=0.127\linewidth]{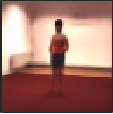}}
\subfloat[\label{fig:semi_sup_50}]{\includegraphics[width=0.127\linewidth]{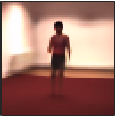}}
\subfloat[\label{fig:semi_sup_25}]{\includegraphics[width=0.125\linewidth]{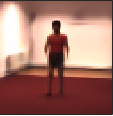}}\hfill
\subfloat[\label{fig:semi_sup_rec_cvae_id94}]{\includegraphics[width=0.128\linewidth]{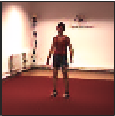}}
\caption{\textbf{Reconstructions on Human3.6M.} ({\bf a}) Original image. ({\bf b}) Heatmap pose representation (rigid parts and body suppressed in the illustration for simplicity), followed by reconstructions with
different levels of supervision:
({\bf c}) 100\%, ({\bf d}) 75\%, ({\bf e}) 50\%, ({\bf f}) 25\%, and ({\bf g})
Conditional-DGPose.}
\label{fig:semi_sup_quali}
\end{figure*}

\begin{figure}[ht]
\begin{minipage}{\linewidth}\subfloat{\includegraphics[width=0.95\linewidth]{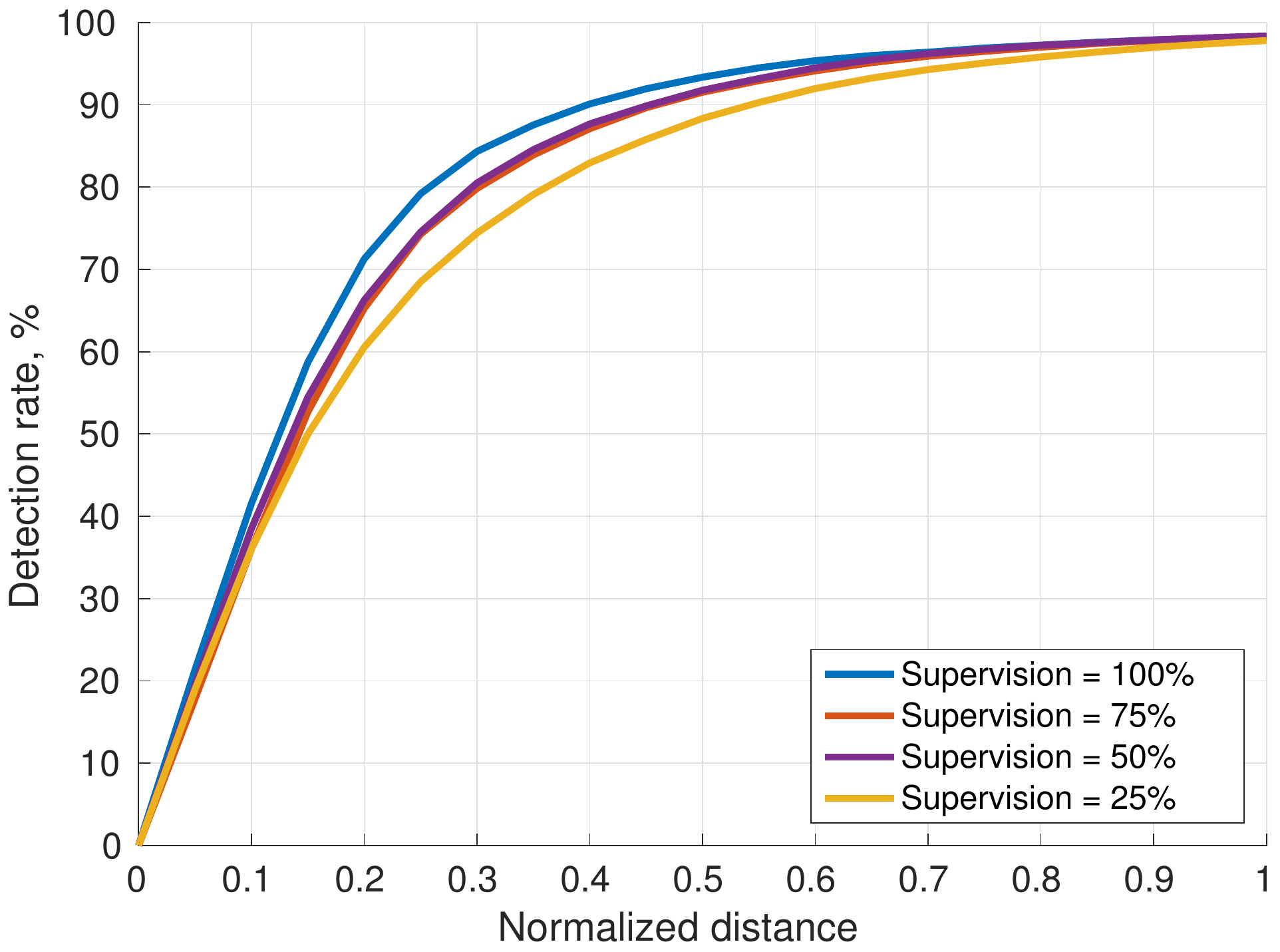}}\end{minipage}
\caption{\textbf{Accuracy of Poses on Human3.6M.} Quantitative evaluations of Semi-DGPose for different levels of supervision using the PCK scores. 
Note that, even with 25\% supervision, our Semi-DGPose obtains 88.35\% PCK score, normalised at 0.5.}
\label{fig:semi_sup_pck}
\end{figure}

With full supervision, we first cross-validated the hyper-parameter $\alpha$ which weights the regression loss (see Eq.~\ref{eq:ss:sup}, in Sec.~\ref{sec:preliminaries}) and found that $\alpha=100$ yields the best results, as shown in Fig.~\ref{fig:pck_alpha}.
Following~\cite{siddharth2016learning}, we keep $\gamma=1$ in all experiments (see Eq.~\ref{eq:ssvae}, in Sec.~\ref{sec:preliminaries}). 
In Fig.~\ref{fig:semi_reconst}, we show reconstructed images along with the heatmap pose representation, which are realistic and comparable with the ones obtained with the Conditional-DGPose (see Fig.~\ref{fig:qualitative_recons}).
\textit{Direct manipulation}, when pose representation is changed during the reconstruction process while appearance is kept the same, is illustrated in Fig.~\ref{fig:direct_manipulation}.
Still with full supervision, we show the pose estimation accuracy for different samples in Fig.~\ref{fig:pose_fully_sup}.
The Semi-DGPose achieves $93.85\%$ PCK score, normalised at $0.5$, in the fully-supervised setup (see Fig.~\ref{fig:semi_sup_pck}).
This pose estimation accuracy is on par with the state-of-the-art pose estimators on unconstrained images~\cite{yang2017learning}. 
However, since the Human3.6M was captured in a controlled environment, a standard (discriminative) pose estimator is expected to perform better.

\begin{figure*}[ht]
\begin{minipage}{\linewidth}
\centering
\subfloat[\label{fig:semi_pose_orig}]{\includegraphics[width=0.125\linewidth]{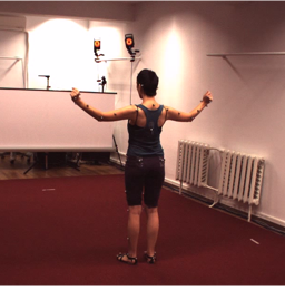}}\hspace{2em}
\subfloat[\label{fig:semi_pose_100}]{\includegraphics[width=0.126\linewidth]{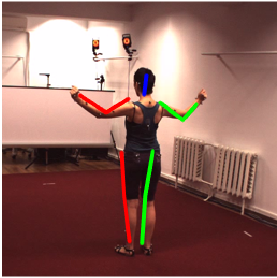}}
\subfloat[\label{fig:semi_pose_75}]{\includegraphics[width=0.125\linewidth]{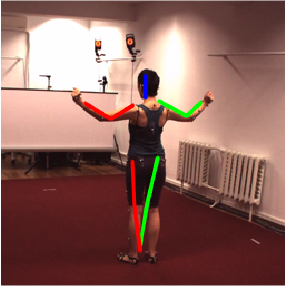}}
\subfloat[\label{fig:semi_pose_50}]{\includegraphics[width=0.125\linewidth]{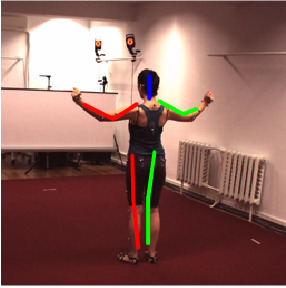}}
\subfloat[\label{fig:semi_pose_25}]{\includegraphics[width=0.126\linewidth]{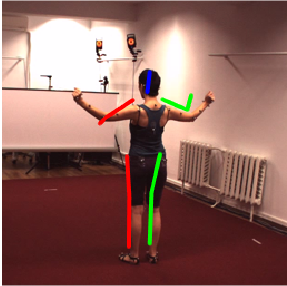}}
\end{minipage}
\caption{\textbf{Qualitative results of semi-supervised pose estimation.} Original image ({\bf a}), followed by
predictions, over the original image, with: ({\bf b}) 100\%, ({\bf c}) 75\%,
({\bf d}) 50\% and ({\bf e}) 25\% of supervision.
The figure aims to illustrate how the decrease in supervision affects pose estimation.
The results are similar, yet it is possible to observe some important discrepancies.
For instance, due to the shortage of labelled training data, the pose estimation result in ({\bf e}) is worse than the one shown in ({\bf b}), particularly regarding the location of arms' extremities.}
\label{fig:semi_sup_pose_estimation}
\end{figure*}

Subsequently, we evaluate it across different levels of supervision, with the PSNR and SSIM metrics and show results in Tab.~\ref{tab:semi_sup_psnr_ssim}.
In Fig.~\ref{fig:semi_sup_quali}, we show reconstructed images obtained with such different levels.
It allows us to observe how image quality is affected when we gradually reduce the availability of labels.
Furthermore, we also evaluate the pose estimation accuracy with semi-supervision. 
The overall PCK curves corresponding to each percentage of supervision in the training set is shown in Fig.~\ref{fig:semi_sup_pck}. 
Note that, even with only 25\% of labels available, our model still obtains 88.35\% PCK score, normalised at 0.5, showing the effectiveness of the semi-supervised approach.
Qualitative samples are shown in Figure~\ref{fig:semi_sup_pose_estimation}.
Again, aiming to illustrate how the gradual decrease of supervision in the training set affects the quality of pose estimation on the test images.

\begin{table}[ht]
\centering
\subfloat{
\begin{tabular}{c|c|c} 
Level of supervision & PSNR & SSIM \\
\hline
100\% & 22.27 & 0.89\\
\hline
75\% & 21.49 & 0.87\\
\hline
50\% & 21.36 & 0.86\\
\hline
25\% & 20.06 & 0.83\\
\end{tabular}
}
\caption{\label{tab:semi_sup_psnr_ssim} \textbf{Image Quality on Human3.6M.} Quantitative evaluations of the Semi-DGPose with different levels of supervision using the PSNR and SSIM metrics.}
\end{table}

\begin{figure}[h]
\centering
\subfloat{\includegraphics[width=.95\linewidth]{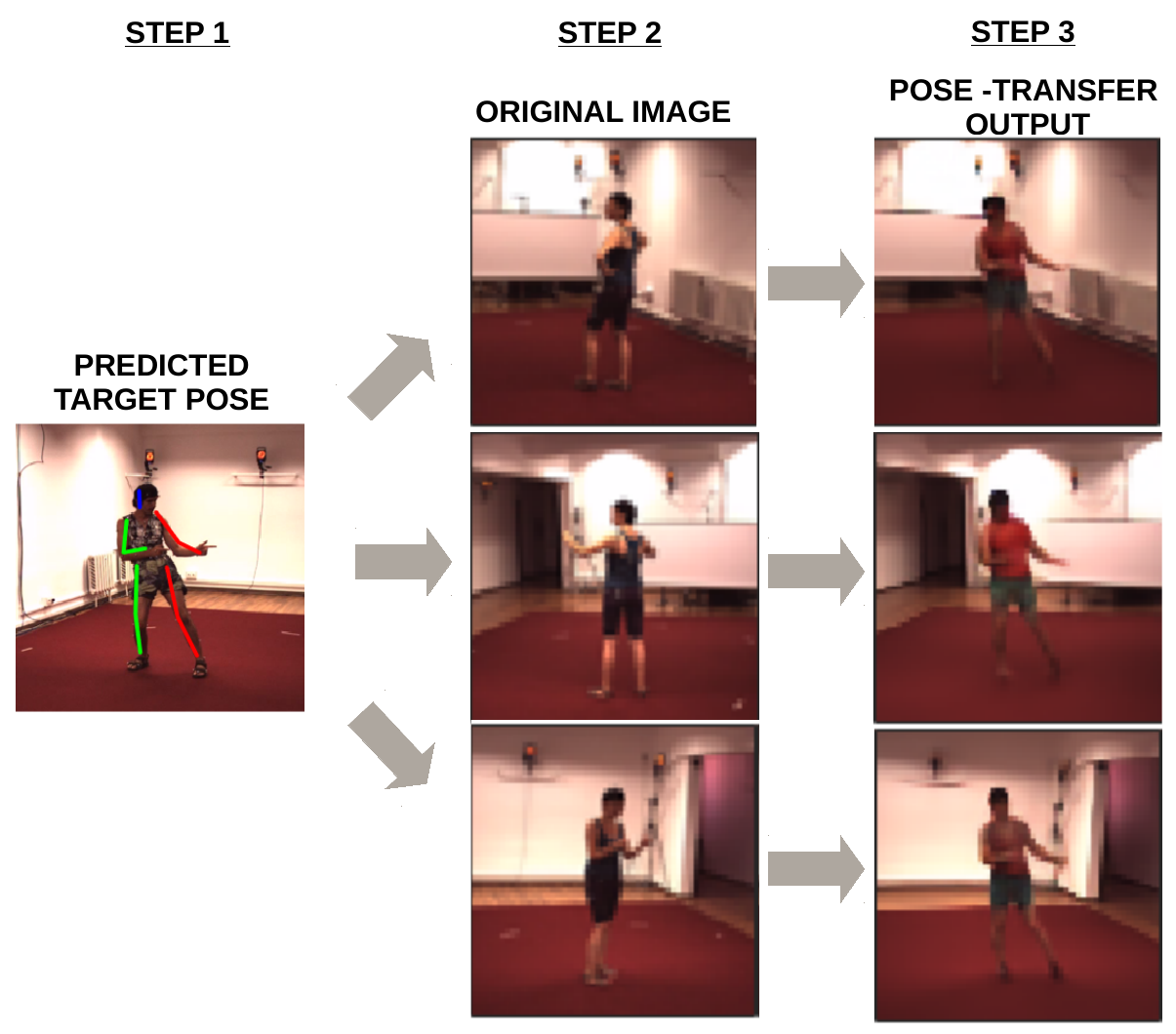}}
\caption{\textbf{\textit{Indirect pose-transfer} on Human3.6M.} {\bf Step 1}: the latent target pose representation $\poseyva$ is estimated (Encoder).
{\bf Step 2}: the image from which the latent \textit{appearance} $\bfz_{_2}$ is estimated (Encoder).
{\bf Step 3}: the output image generated as a combination of $\poseyva$ and $\bfz_{_2}$ (Decoder).
The people's outfits in the output images are approximated to the ones in the original images. However, restricted by the low diversity of outfits observed in Human3.6M training data.
Note that, to highlight the separation of appearance and pose, we chose the image on Step 1 to be from camera 2, while the original images are from cameras, 1, 3 and 4, respectively.
As can be seen, the background scene is totally defined by the original images.
}
\label{fig:semi_sup_pose_transfer}
\end{figure}

\begin{figure}[ht]
\begin{minipage}{.965\linewidth}
\begin{tabular}{cc}
\multicolumn{2}{c}{
\textbf{\scriptsize\makebox[6em][l]{}\textsf{JOINTS}\hfill\textsf{RIGID}\hfill\textsf{BODY}\makebox[5.5em][l]{}}
\vspace{-1.2em}
}
\\
\multicolumn{2}{c}{
\renewcommand{\thesubfigure}{a}
\subfloat[]{\includegraphics[trim={0 .05cm 6.05cm .05cm},clip,width=0.725\linewidth]{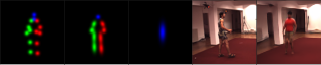}}
\vspace{1em}
}
\\
\renewcommand{\thesubfigure}{b}
\begin{minipage}{0.5\linewidth}
\raggedright
\textbf{\scriptsize\makebox[1em][l]{}\textsf{ORIGINAL}\hfill\textsf{OUTPUT}\makebox[2.1em][l]{}}
\subfloat{\includegraphics[trim={9.05cm .05cm 0 .05cm},clip,width=.925\linewidth]{figures/semi_100_22_new.png}} \vspace{-1em}\\
\subfloat[\label{fig:semi_reconst_100}]{\includegraphics[trim={9.05cm .05cm 0 .05cm},clip,width=.925\linewidth]{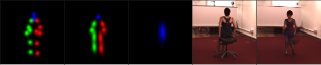}}
\end{minipage} &
\renewcommand{\thesubfigure}{c}
\begin{minipage}{0.5\linewidth}
\raggedright
\textbf{\scriptsize\makebox[1em][l]{}\textsf{ORIGINAL}\hfill\textsf{OUTPUT}\makebox[2.1em][l]{}}
\subfloat{\includegraphics[trim={9.05cm .05cm 0 .05cm},clip,width=.925\linewidth]{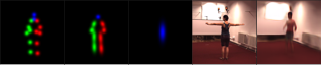}} \vspace{-1em}\\
\subfloat[\label{fig:semi_reconst_75}]{\includegraphics[trim={9.05cm .05cm 0 .05cm},clip,width=.925\linewidth]{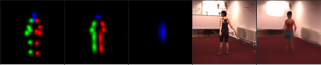}}
\end{minipage}
\\
\renewcommand{\thesubfigure}{d}
\begin{minipage}{0.5\linewidth}
\raggedright
\vspace{1em}
\textbf{\scriptsize\makebox[1em][l]{}\textsf{ORIGINAL}\hfill\textsf{OUTPUT}\makebox[2.1em][l]{}}
\subfloat{\includegraphics[trim={9.05cm .05cm 0 .05cm},clip,width=0.925\linewidth]{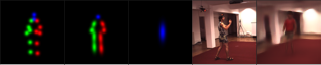}} \vspace{-1em}\\
\subfloat[\label{fig:semi_reconst_50}]{\includegraphics[trim={9.05cm .05cm 0 .05cm},clip,width=0.925\linewidth]{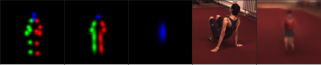}}
\end{minipage} &
\renewcommand{\thesubfigure}{e}
\begin{minipage}{0.5\linewidth}
\raggedright
\vspace{1em}
\textbf{\scriptsize\makebox[1em][l]{}\textsf{ORIGINAL}\hfill\textsf{OUTPUT}\makebox[2.1em][l]{}}
\subfloat{\includegraphics[trim={9.05cm .05cm 0 .05cm},clip,width=0.925\linewidth]{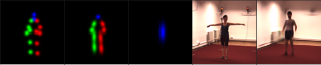}} \vspace{-1em}\\
\subfloat[\label{fig:semi_reconst_25}]{\includegraphics[trim={9.05cm .05cm 0 .05cm},clip,width=0.925\linewidth]{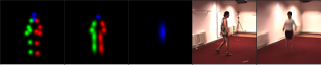}}
\end{minipage}
\end{tabular}
\end{minipage}
\caption{\label{fig:semi_level_sup}
\textbf{Indirect Pose-transfer on Human3.6M.}
Qualitative results with different levels of supervision. 
(a) Heatmap representation of the target pose (i.e. after being processed by the Mapper module) used for all the subsequent results. 
Such results show pairs of original images and pose-transfer outputs obtained with the following levels of supervision: (b) 100\%, (c) 75\%, (d) 50\%, and (e) 25\%.
In the pose-transfer outputs, appearance comes from the original images while the body posture is defined by the target pose.
}
\vspace{-1em}
\end{figure}

Concerning \textit{indirect pose-transfer}, as both latent variables corresponding to pose and appearance can be inferred by the model's Encoder (recognition network) at test time, 
latent variables extracted from different images can be combined in a subsequent step, and employed together as inputs for the Decoder (generative network).
The result of that is a generated image combining appearance and body pose, extracted from two different images.
The process is done in three phases, as illustrated in Fig.~\ref{fig:semi_sup_pose_transfer}.
Firstly, the latent pose representation $\poseyva$ is estimated from the first input image through the Encoder.
Secondly, the latent \textit{appearance} representation $\bfz_{_2}$ is estimated from a second image, also through the Encoder.
Lastly, $\poseyva$ and $\bfz_{_2}$ are propagated through the Decoder, and a new image is generated, combining body pose and appearance, respectively, from the first and second \textit{encoded} images.
We evaluate qualitatively the effects of semi-supervision over the indirect pose-transfer in Fig.~\ref{fig:semi_level_sup}.

\subsubsection{Semi-DGPose Results on DeepFashion}

\begin{figure}[h]
\includegraphics[width=0.95\linewidth]{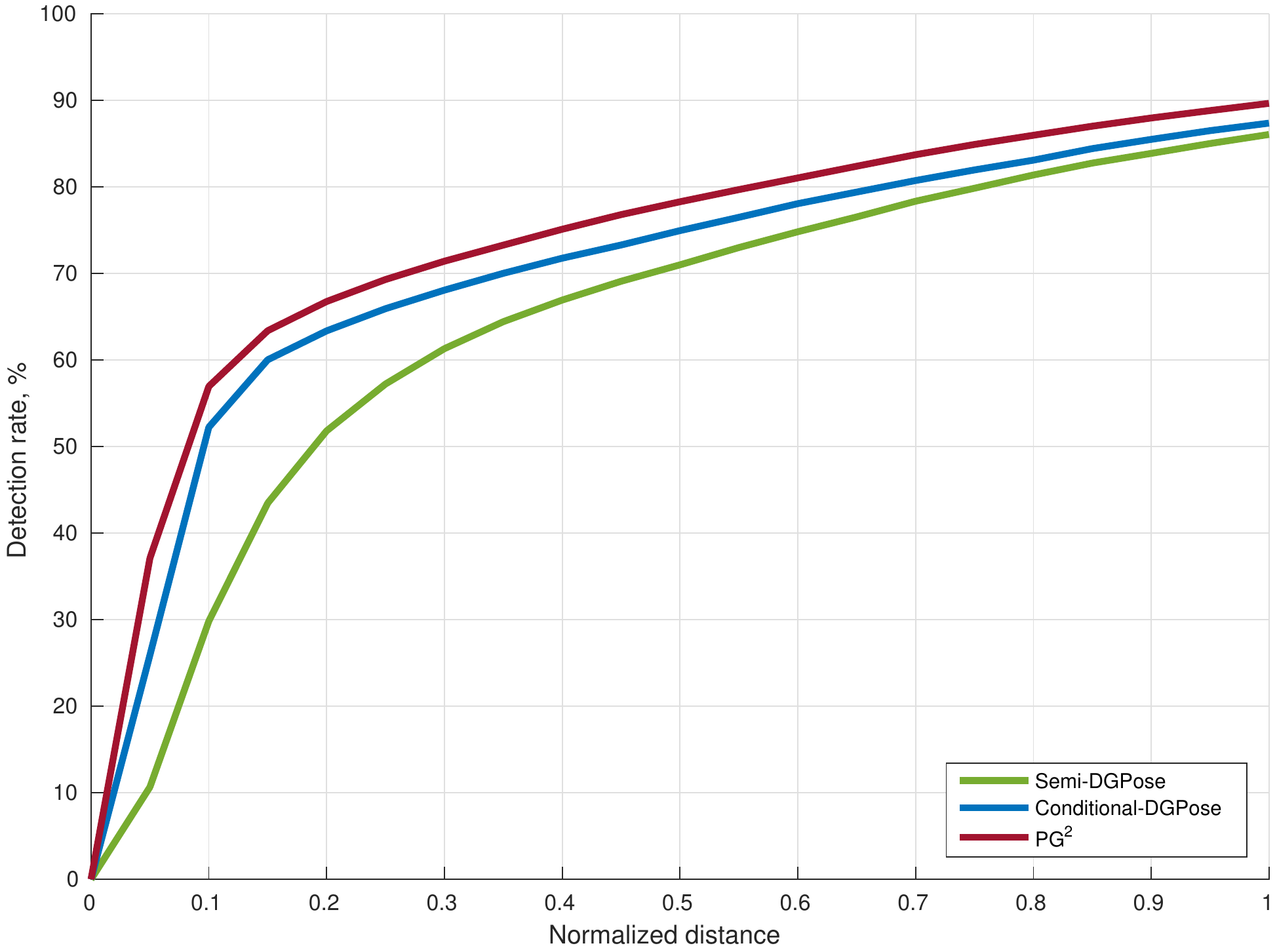}
\caption{\textbf{Accuracy of Poses on DeepFashion.} 
Quantitative evaluation of Semi-DGPose PCK scores over reconstructed poses. 
The Semi-DGPose (\textit{green}) shows less accurate results, however, in contrast to the Conditional-DGPose (\textit{blue}) and the $\text{PG}^2$ network~\cite{ma2017} (\textit{red}), 
it performs a significantly more complex task, simultaneously executing pose estimation and allowing for semi-supervised learning.
Detection rate represents the percentage of joints correctly relocated in the reconstructions.}
\label{fig:pck_df_vs_our_semi}
\end{figure}

To show the generality of the Semi-DGPose, model we present additional results on the DeepFashion dataset, using our Conditional-DGPose architecture and the image-to-image translation network $\text{PG}^2$~\cite{ma2017} as baselines.
The same hyper-parameters reported previously were used in training.
In Tab.~\ref{table:df_image_quality_semi_dgpose}, we compare the image quality of reconstructions, while in Fig.~\ref{fig:pck_df_vs_our_semi} we show the comparison concerning the quality of pose reconstructions.
Although the Semi-DGPose presents less accurate results, it is important to highlight that it is also tackling the pose estimation task, which is not performed by either one of the other two methods, \ie the Conditional-DGPose and the $\text{PG}^2$.
To pursue, simultaneously, the \textit{understanding}, \ie estimation of pose and appearance in the latent space, and the \textit{generation} of people directly in images, shows to be indeed a significantly more complex task.
Nevertheless, the justification for seeking such a challenging goal, as mentioned before, mainly lie on its important capability of allowing for semi-supervised learning, that is not present in the comparable methods.
\begin{table}[h]
\centering
\begin{tabular}{c|c|c} 
 & PSNR & SSIM \\
\hline
Semi-DGPose & 16.84 & 0.76 \\ 
\hline
Conditional-DGPose & 18.38 & 0.79 \\
\hline
$\text{PG}^2$~\cite{ma2017} & \textbf{\underline{18.96}} & \textbf{\underline{0.83}}\\
\end{tabular}
\caption{\textbf{Image Quality on DeepFashion.} Quantitative evaluation of Semi-DGPose using PSNR and SSIM measures comparing the image quality of reconstructions.
The Semi-DGPose shows less accurate results, yet in contrast to the other methods, it performs a significantly more complex task, simultaneously executing pose estimation, and also allowing for semi-supervised learning.
}
\label{table:df_image_quality_semi_dgpose}
\end{table}

\begin{figure}[ht]
\begin{tabular}{cc}
\begin{minipage}{0.46\linewidth}
\tiny\textbf{\textsf{\makebox[2.5em][l]{} ORIGINAL \makebox[2em][l]{} RECONSTRUCTION\makebox[1em][l]{}}}\vspace{-1em}\\
\subfloat{\includegraphics[trim={8.95cm .05cm 0 .05cm},clip,width=\linewidth]{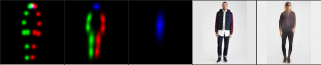}} \vspace{-1em}\\
\subfloat{\includegraphics[trim={8.95cm .05cm 0 .05cm},clip,width=\linewidth]{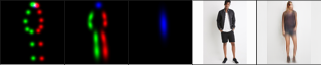}}
\end{minipage} &
\begin{minipage}{0.46\linewidth}
\tiny\textbf{\textsf{\makebox[2.5em][l]{} ORIGINAL \makebox[2em][l]{} RECONSTRUCTION\makebox[1em][l]{}}}\vspace{-1em}\\
\subfloat{\includegraphics[trim={8.95cm .05cm 0 .05cm},clip,width=\linewidth]{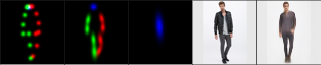}} \vspace{-1em}\\
\subfloat{\includegraphics[trim={8.95cm .05cm 0 .05cm},clip,width=\linewidth]{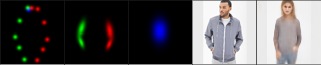}}
\end{minipage}
\vspace{.2em}\\
\begin{minipage}{0.46\linewidth}
\subfloat{\includegraphics[trim={8.95cm .05cm 0 .05cm},clip,width=\linewidth]{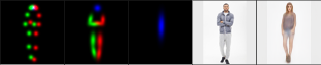}} \vspace{-1em}\\
\subfloat{\includegraphics[trim={8.95cm .05cm 0 .05cm},clip,width=\linewidth]{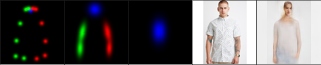}}
\end{minipage} &
\begin{minipage}{0.46\linewidth}
\subfloat{\includegraphics[trim={8.95cm .05cm 0 .05cm},clip,width=\linewidth]{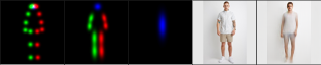}} \vspace{-1em}\\
\subfloat{\includegraphics[trim={8.95cm .05cm 0 .05cm},clip,width=\linewidth]{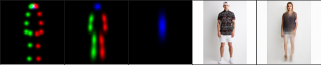}}
\end{minipage}
\end{tabular}
\caption{\textbf{Reconstructions on DeepFashion.} 
The only input of the Semi-DGPose is the original image.
At test time, as pose is estimated in the latent space, discrepancies between the original and reconstructed poses are more frequently observed, in comparison with the Conditional-DGPose.
Best viewed if zoomed in digital version.}
\label{fig:semi_df}
\vspace{-1em}
\end{figure}

In Fig.~\ref{fig:semi_df}, we show comparisons between input and reconstructed images.
In some of the samples, we can observe small differences between the original and the reconstructed body postures, mainly regarding the positions of the limbs. 
This illustrates the higher complexity involved in simultaneously estimating pose and appearance in our latent space.
For instance, inaccurate predictions of pose, performed by the Encoder, may have effects into the final reconstructed appearance, and vice-versa, when the latent representations are mapped back to the image space, by the Decoder.  
Such interdependency does not exist when pose is a given observable variable, as in the case of the conditional models or image-to-image translation networks.

\begin{figure*}[ht]
\centering
\begin{minipage}{0.475\linewidth}
\subfloat{
\captionsetup{position=top,labelformat=empty}
\subfloat[\tiny\textbf{\textsf{ORIGINAL}}]{\includegraphics[scale=.20775]{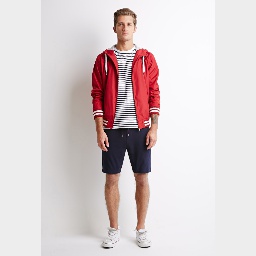}}
\subfloat[\tiny\textbf{\textsf{TARGET POSE}}]{\includegraphics[scale=.207]{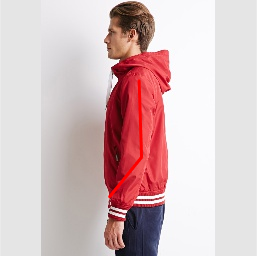}}
\subfloat[\tiny\textbf{\textsf{OURS}}]{\includegraphics[scale=.83]{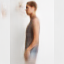}}
\subfloat[\tiny\textbf{\textsf{PG$^2$}}]{\includegraphics[scale=.20775]{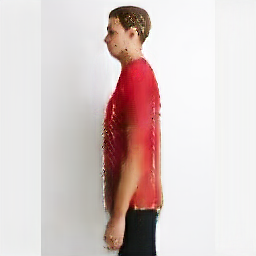}}
}
\vspace{-2.1em}
\end{minipage}
\begin{minipage}{0.475\linewidth}
\subfloat{
\captionsetup{position=top,labelformat=empty}
\subfloat[\tiny\textbf{\textsf{ORIGINAL}}]{\includegraphics[scale=.20775]{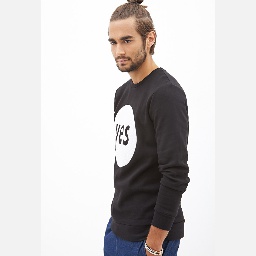}}
\subfloat[\tiny\textbf{\textsf{TARGET POSE}}]{\includegraphics[scale=.2025]{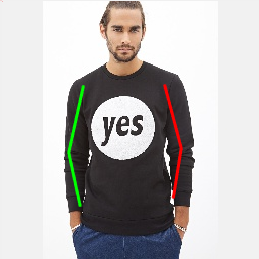}}
\subfloat[\tiny\textbf{\textsf{OURS}}]{\includegraphics[scale=.83]{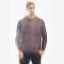}}
\subfloat[\tiny\textbf{\textsf{PG$^2$}}]{\includegraphics[scale=.20775]{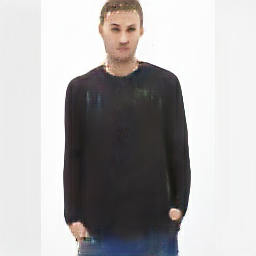}}
}
\vspace{-2.1em}
\end{minipage}
\begin{minipage}{0.475\linewidth}
\subfloat{
\subfloat{\includegraphics[scale=.20775]{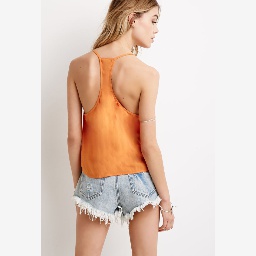}}
\subfloat{\includegraphics[scale=.2025]{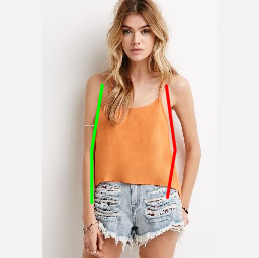}}
\subfloat{\includegraphics[scale=.83]{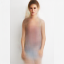}}
\subfloat{\includegraphics[scale=.20775]{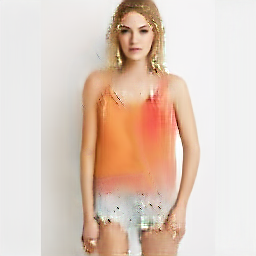}}
}
\vspace{-1em}
\end{minipage}
\begin{minipage}{0.475\linewidth}
\subfloat{
\subfloat{\includegraphics[scale=.20775]{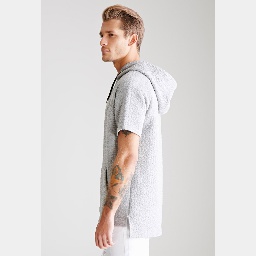}}
\subfloat{\includegraphics[scale=.2025]{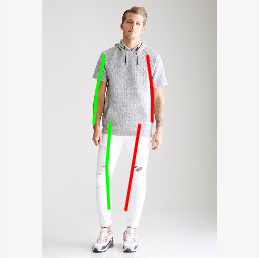}}
\subfloat{\includegraphics[scale=.83]{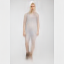}}
\subfloat{\includegraphics[scale=.20775]{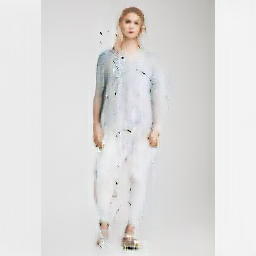}}
}
\vspace{-1em}
\end{minipage}
\begin{minipage}{0.475\linewidth}
\subfloat{
\subfloat{\includegraphics[scale=.20775]{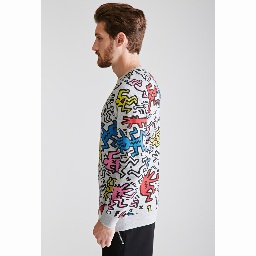}}
\subfloat{\includegraphics[scale=.2025]{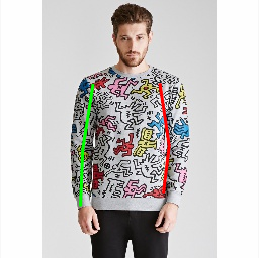}}
\subfloat{\includegraphics[scale=.83]{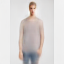}}
\subfloat{\includegraphics[scale=.20775]{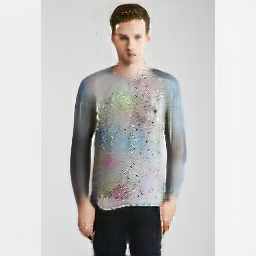}}
}
\vspace{-1em}
\end{minipage}
\begin{minipage}{0.475\linewidth}
\subfloat{
\subfloat{\includegraphics[scale=.20775]{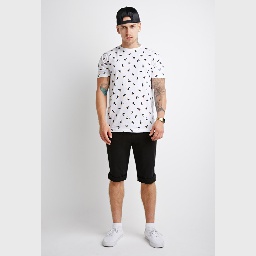}}
\subfloat{\includegraphics[scale=.2025]{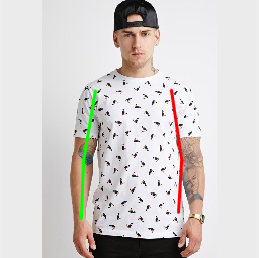}}
\subfloat{\includegraphics[scale=.83]{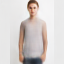}}
\subfloat{\includegraphics[scale=.20775]{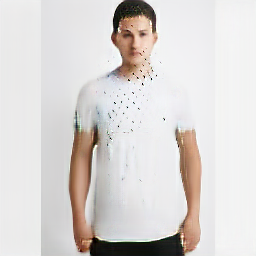}}
}
\vspace{-1em}
\end{minipage}
\begin{minipage}{0.475\linewidth}
\subfloat{
\subfloat{\includegraphics[scale=.20775]{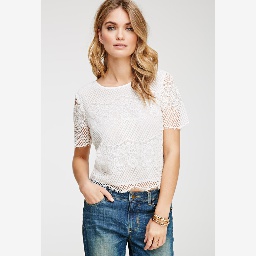}}
\subfloat{\includegraphics[scale=.2025]{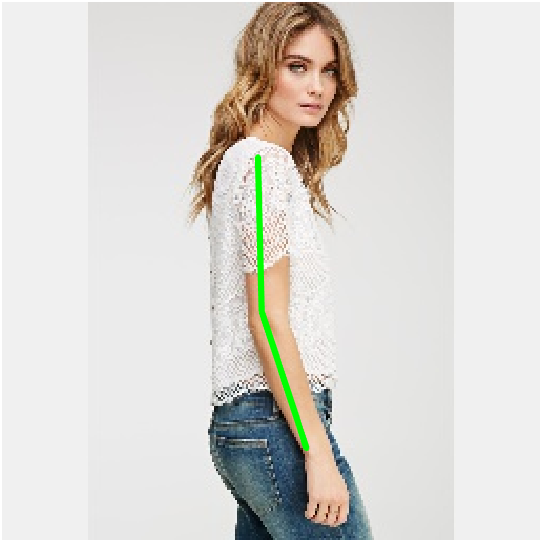}}
\subfloat{\includegraphics[scale=.83]{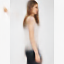}}
\subfloat{\includegraphics[scale=.20775]{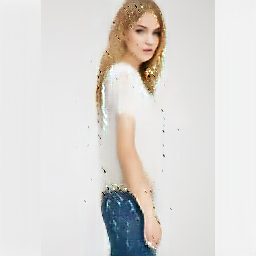}}
}
\vspace{-1em}
\end{minipage}
\begin{minipage}{0.475\linewidth}
\subfloat{
\subfloat{\includegraphics[scale=.20775]{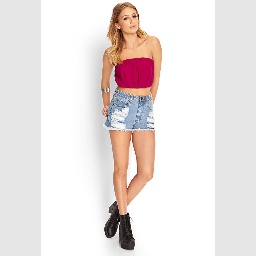}}
\subfloat{\includegraphics[scale=.2025]{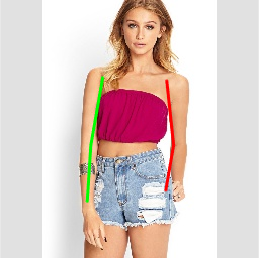}}
\subfloat{\includegraphics[scale=.83]{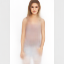}}
\subfloat{\includegraphics[scale=.20775]{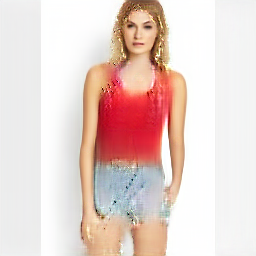}}
}
\vspace{-1em}
\end{minipage}
\begin{minipage}{0.475\linewidth}
\subfloat{
\subfloat{\includegraphics[scale=.20775]{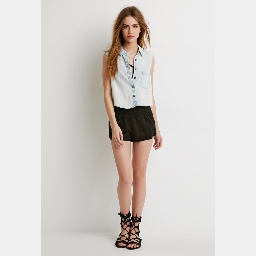}}
\subfloat{\includegraphics[scale=.2025]{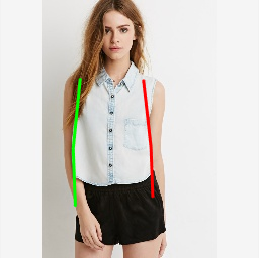}}
\subfloat{\includegraphics[scale=.83]{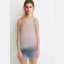}}
\subfloat{\includegraphics[scale=.20775]{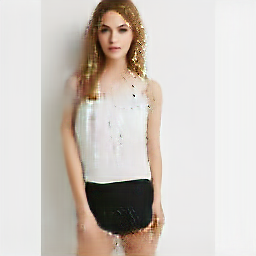}}
}
\vspace{-1em}
\end{minipage}
\begin{minipage}{0.475\linewidth}
\subfloat{
\subfloat{\includegraphics[scale=.20775]{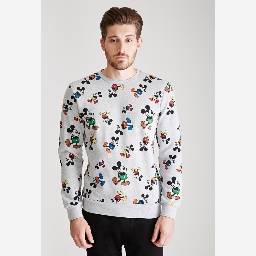}}
\subfloat{\includegraphics[scale=.2025]{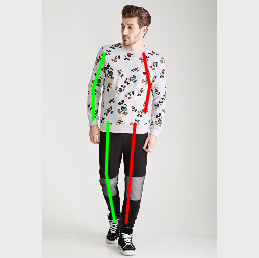}}
\subfloat{\includegraphics[scale=.83]{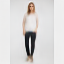}}
\subfloat{\includegraphics[scale=.20775]{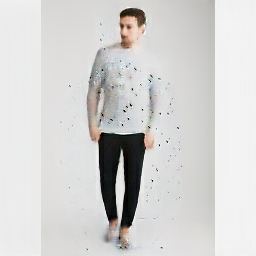}}
}
\vspace{-1em}
\end{minipage}
\begin{minipage}{0.475\linewidth}
\subfloat{
\subfloat{\includegraphics[scale=.20775]{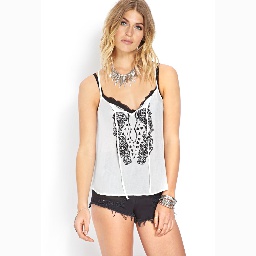}}
\subfloat{\includegraphics[scale=.205]{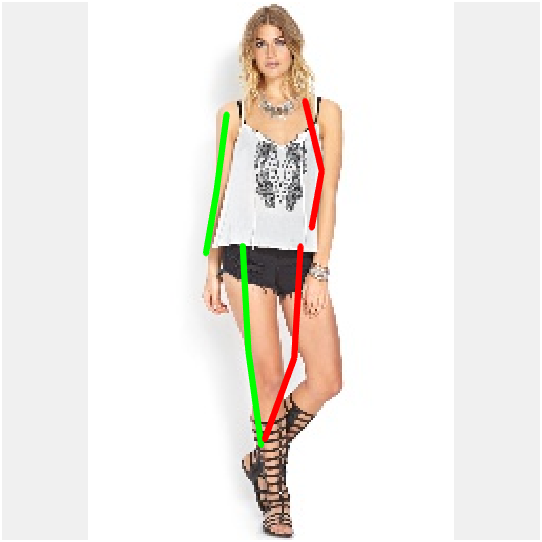}}
\subfloat{\includegraphics[scale=.83]{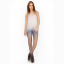}}
\subfloat{\includegraphics[scale=.20775]{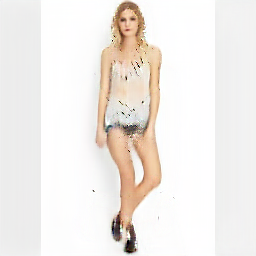}}
}
\end{minipage}
\begin{minipage}{0.475\linewidth}
\subfloat{
\subfloat{\includegraphics[scale=.20775]{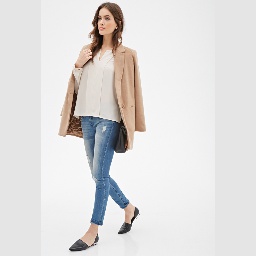}}
\subfloat{\includegraphics[scale=.205]{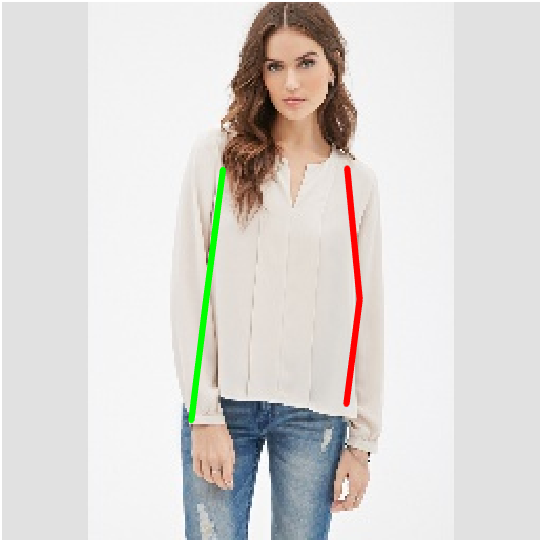}}
\subfloat{\includegraphics[scale=.83]{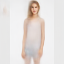}}
\subfloat{\includegraphics[scale=.20775]{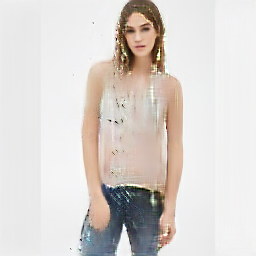}}
}
\end{minipage}
\caption{\textbf{Indirect pose-transfer in DeepFashion dataset}. 
In each set of images, we have, respectively: the original image, the target image with the superimposed target pose predicted by the Semi-DGPose,
the pose-transfer output from the Semi-DGPose and the pose-transfer output from $\textnormal{PG}^{2}$~\cite{ma2017}.
Although tackling a more complex task than~\cite{ma2017}, which includes the prediction of pose, our results are still reasonable.}
\label{fig:ind_pose_transf_semidgpose}
\vspace{-1em}
\end{figure*}

Finally, we highlight \textit{indirect pose-transfer} in the DeepFashion dataset, which is a distinctive capability of the Semi-DGPose, in comparison to related methods.
In \figref{fig:ind_pose_transf_semidgpose}, we compare the indirect pose-transfer results, from our single-stage structured generative model, the Semi-DGPose, with the results from the image-to-image translation baseline, the $\text{PG}^2$ network~\cite{ma2017}.
It is important to notice that our Semi-DGPose model was not trained specifically for pose-transfer, \ie it was not trained on pairs of images.
On the other hand, the $\text{PG}^2$ architecture is trained on pairs of images of the same person, in different poses, scales or point of views (first two images of each set in~\figref{fig:ind_pose_transf_semidgpose}).
Moreover, in the Semi-DGPose the body pose is estimated by the Encoder network (illustrated in every second image of each set in~\figref{fig:ind_pose_transf_semidgpose}), along with appearance, while in the $\text{PG}^2$ pose is given as an observable variable to the model.
Despite such critical competitive disadvantages, we can observe that the Semi-DGPose produce reasonable results in comparison to the ones from $\text{PG}^2$.
Lastly, it is crucial to call attention for the capabilities of our Semi-DGPose approach such as, interpretability of the latent space, pose estimation, sampling and semi-supervised learning, which are not jointly present in the $\text{PG}^2$ or in the related work from the literature.
These features justify our approach for learning a deep generative model of people in images and, to our knowledge, significantly differentiate the Semi-DGPose model from prior art.

\subsection{Limitations of the Models\label{sec:limitations}}

Here, we discuss two important limitations common to the Conditional-DGPose and the Semi-DGPose.
The first refers to the modelling of appearance in both models.
As we mention in Sec.~\ref{sec:intro}, our latent representation of appearance encodes all the visual information in the images (e.g. clothing, skin colours, hairstyles, and background) except for the body pose of the subjects.
However, such a strategy does not separate the individual visual characteristics in the latent representation.
In Fig.~\ref{fig:df_manifold_condDGPose} (Sec.~\ref{subsec:cond_dgpose_df}), we can observe that as the appearance manifold is traversed, the visual features gradually change altogether. 
A disentangled representation for appearance itself would be needed for allowing control over specific visual features.
Another aspect concerning appearance regards limitations to approximate clothing ``seen'' few times or ``unseen'' during training.
Interestingly, the extrapolation capabilities shown for unseen poses (see Fig.~\ref{fig:multi_people_unreal} in Sec.~\ref{sec:exp_cond_h36m}) is not observed for appearance.
For example, in the Human3.6M dataset, the low diversity of subjects’ outfits may eventually prevent the clothing in the reconstructed images to be precisely equal to the ones in the original images, as can be observed in Fig.~\ref{fig:semi_sup_pose_transfer} (Sec.~\ref{sec:semidgpose_h36m}).  
Other works in the literature refer to this same problem concerning the Human3.6M dataset, e.g. Rhodin et al.~\cite{rhodin2018unsupervised}.

The second relevant limitation refers to our pose representation.
Aiming to investigate and explore the capabilities of simple body representations, we have worked only with 2D pose in our models.
Such option turns our approaches more general since they are not dependent on 3D information (e.g. 3D models, camera calibration, or multi-view images).
It allows, for example, their application on ordinary monocular images.
Moreover, this strategy is also less susceptible to body shape variations in comparison to segmentations mask or 3D body meshes, which might not be directly transferable from one person to another. 
However, such simplicity creates some limitations. 
An important one concerns the lack of depth information in the body model.
Despite the reasonable results obtained with single people in relatively ``well-behaved'' poses, the models might face difficulties in the presence of stronger self-occlusions associated with particular body postures.
In the absence of depth, it is hard to infer, for instance, which one of two overlapping limbs is closer to the camera.
Without such explicit information in the body representation, the correct reconstruction might present flaws.

To analyse such issues here, which are present in our both models, we have employed the Conditional-DGPose, trained on the Human3.6M dataset, to perform cross-domain pose-transfer over single images from short video sequences.
Employing a sequence of frames allow us to observe how the performance of the model changes according to the concurrent presence of self-occlusion and different poses.
In the current experiments, we have used short videos from the JHMDB dataset~\cite{Jhuang2013}.
Each ``in-the-wild'' video depicts a single person performing one activity.
The dataset provides 2D pose annotations per frame for all videos.
Such annotations are used as inputs for the Conditional-DGPose cross-domain pose-transfer.
We crop the images maintaining the subjects centralised.

\begin{figure*}[ht]
\centering
\begin{minipage}{\linewidth}
\subfloat{
\captionsetup{position=top,labelformat=empty}
\subfloat[\tiny\textbf{\textsf{FRAME 1}}]{\includegraphics[trim={0cm 0cm 0cm 0.4cm},clip,scale=.35]{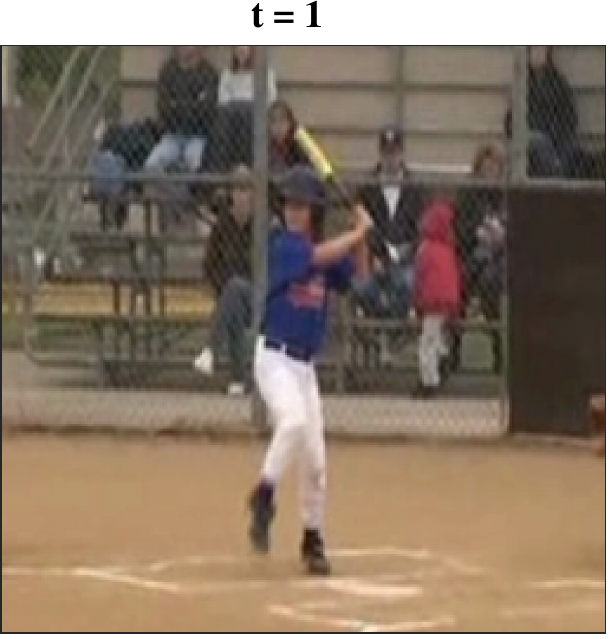}}
\subfloat[\tiny\textbf{\textsf{FRAME 5}}]{\includegraphics[trim={0cm 0cm 0cm 0.4cm},clip,scale=.43]{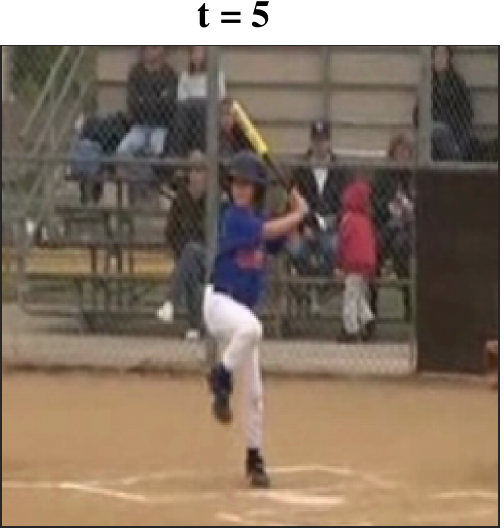}}
\subfloat[\tiny\textbf{\textsf{FRAME 9}}]{\includegraphics[trim={0cm 0cm 0cm 0.5cm},clip,scale=.35]{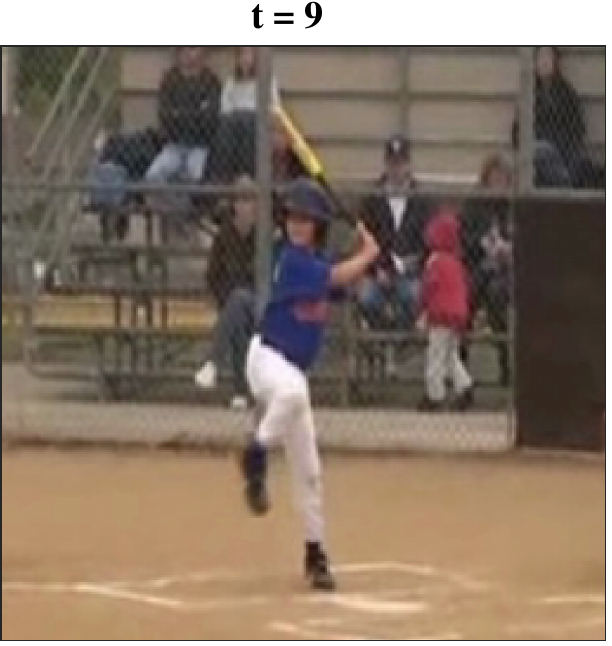}}
\subfloat[\tiny\textbf{\textsf{FRAME 13}}]{\includegraphics[trim={0cm 0cm 0cm 0.6cm},clip,scale=.35]{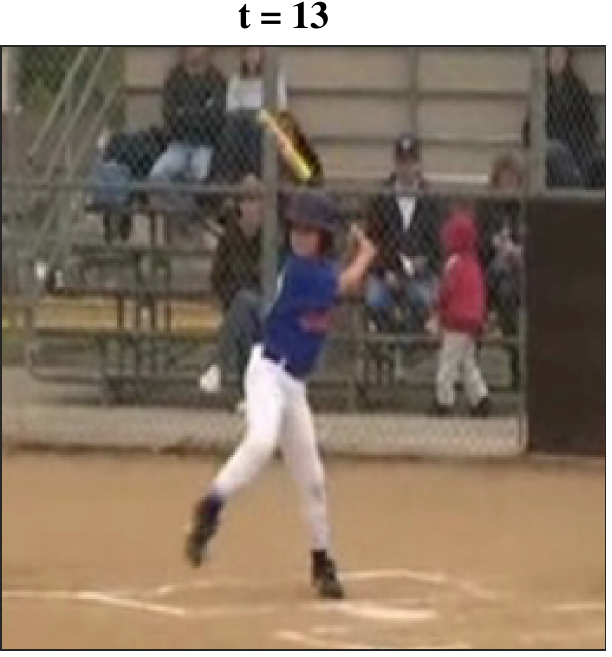}}
\subfloat[\tiny\textbf{\textsf{FRAME 17}}]{\includegraphics[trim={0cm 0cm 0cm 0.55cm},clip,scale=.35]{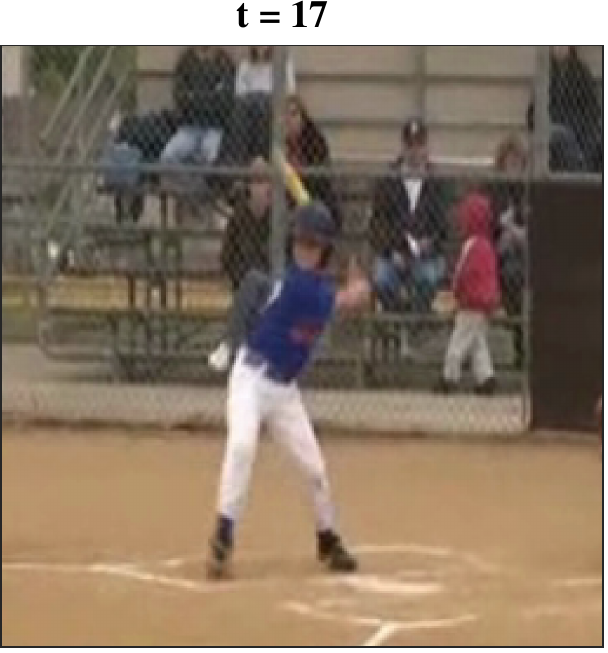}}
\subfloat[\tiny\textbf{\textsf{FRAME 21}}]{\includegraphics[trim={0cm 0cm 0cm 0.6cm},clip,scale=.35]{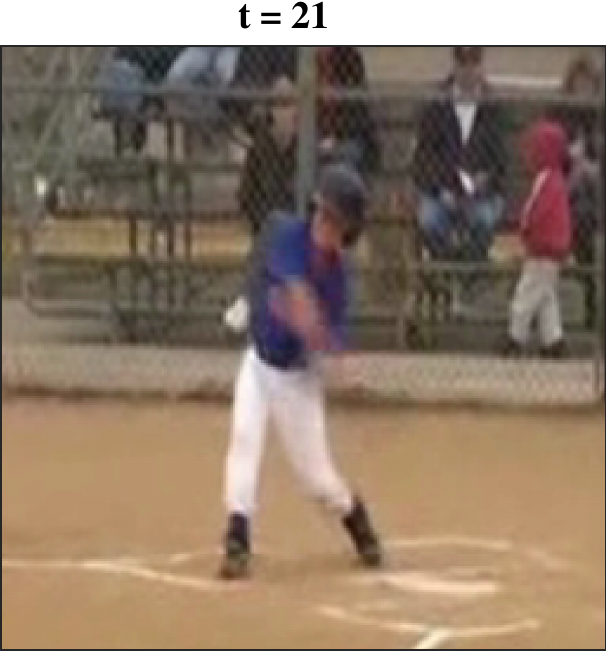}}
\subfloat[\tiny\textbf{\textsf{FRAME 25}}]{\includegraphics[trim={0cm 0cm 0cm 0.55cm},clip,scale=.35]{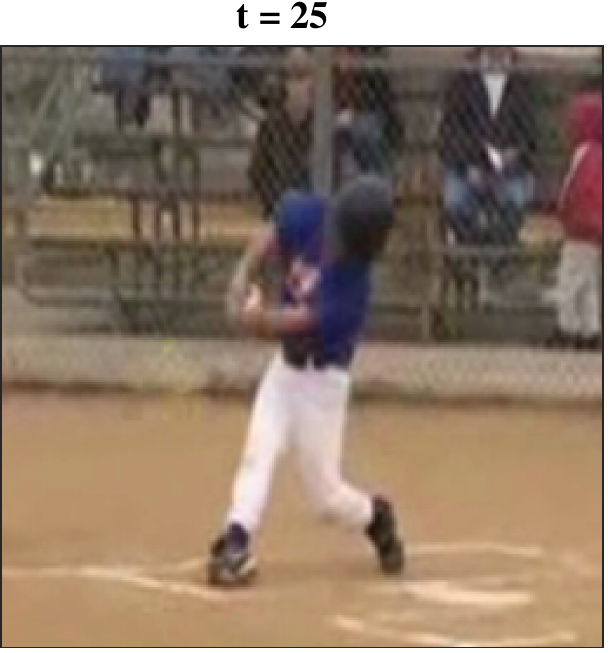}}
\subfloat[\tiny\textbf{\textsf{FRAME 29}}]{\includegraphics[trim={0cm 0cm 0cm 0.55cm},clip,scale=.35]{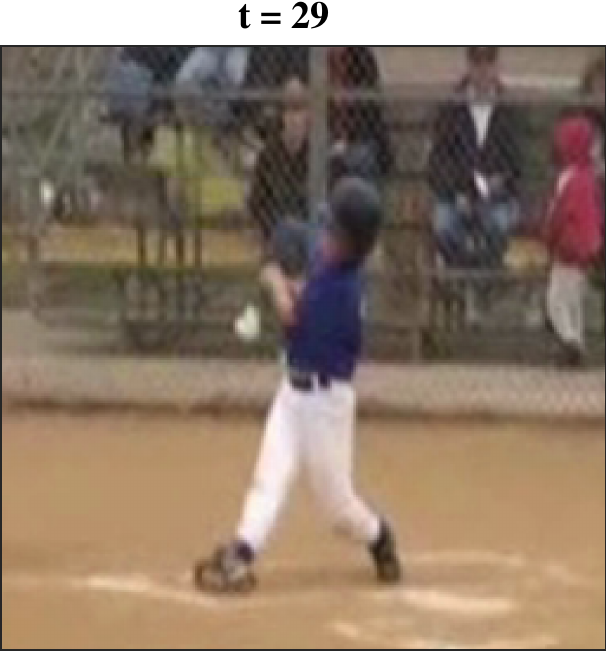}}
}
\vspace{-2em}
\end{minipage}
\vspace{-1.5em}
\begin{minipage}{\linewidth}
\centering
\renewcommand{\thesubfigure}{a}
\subfloat[\label{fig:baseball}]{
\subfloat{\includegraphics[trim={0cm 0cm 0cm 0.4cm},clip,scale=.35]{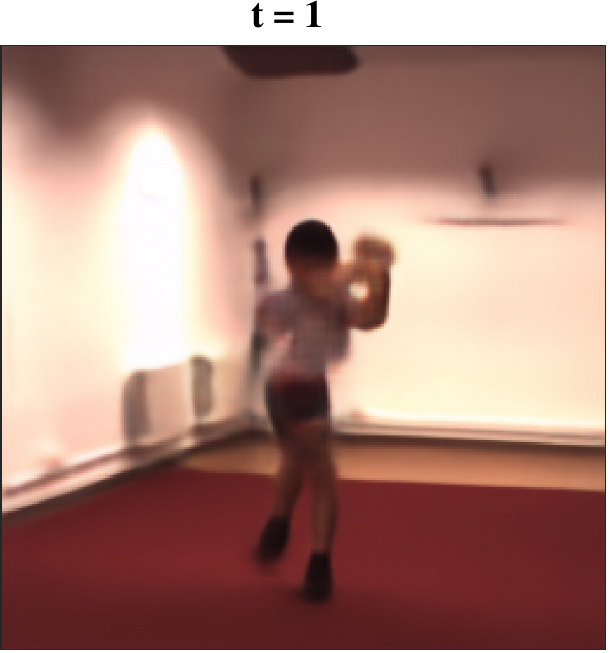}}
\subfloat{\includegraphics[trim={0cm 0cm 0cm 0.4cm},clip,scale=.35]{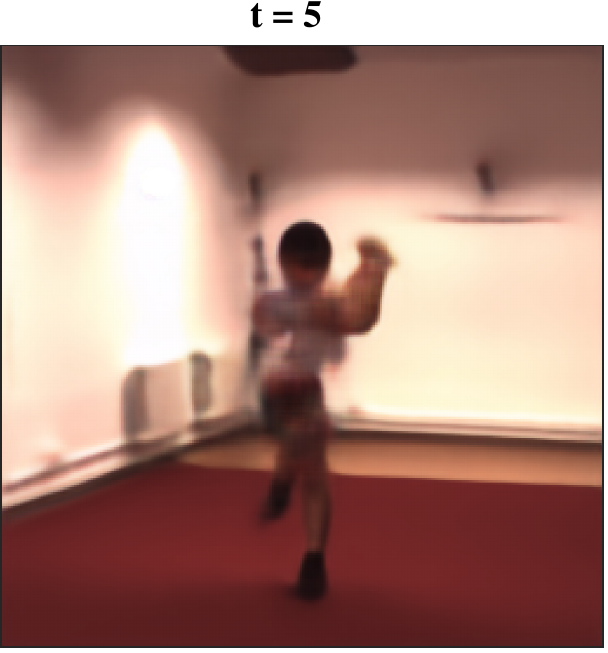}}
\subfloat{\includegraphics[trim={0cm 0cm 0cm 0.4cm},clip,scale=.35]{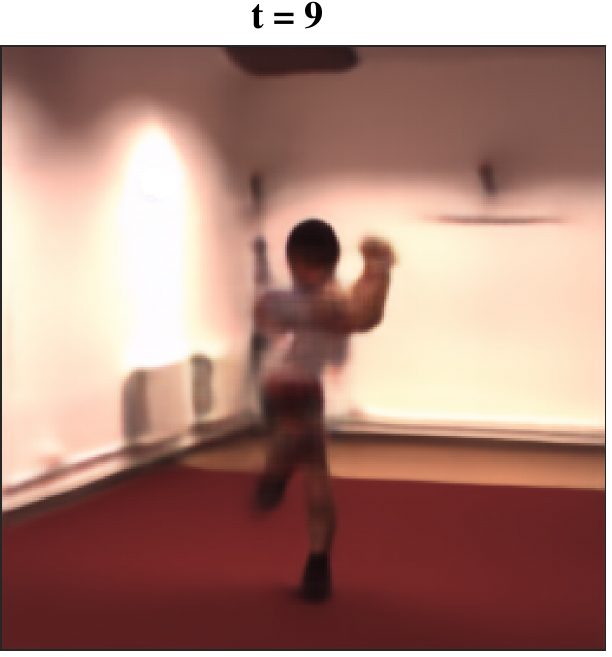}}
\subfloat{\includegraphics[trim={0cm 0cm 0cm 0.4cm},clip,scale=.35]{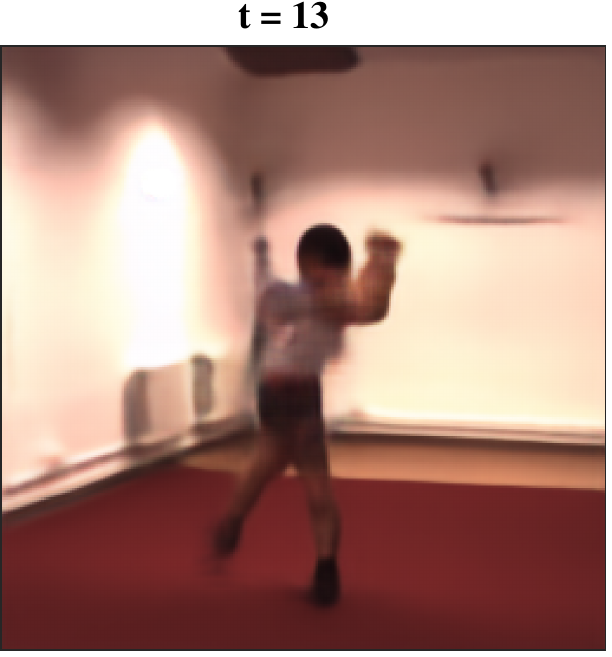}}
\subfloat{\includegraphics[trim={0cm 0cm 0cm 0.4cm},clip,scale=.35]{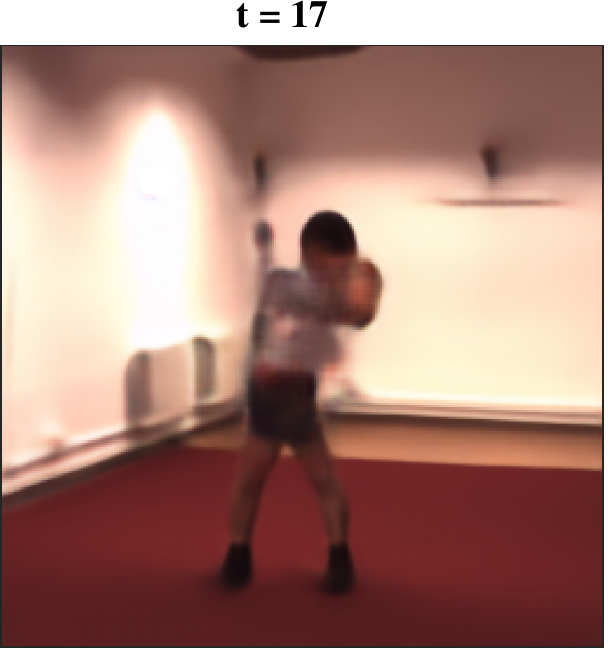}}
\subfloat{\includegraphics[trim={0cm 0cm 0cm 0.4cm},clip,scale=.35]{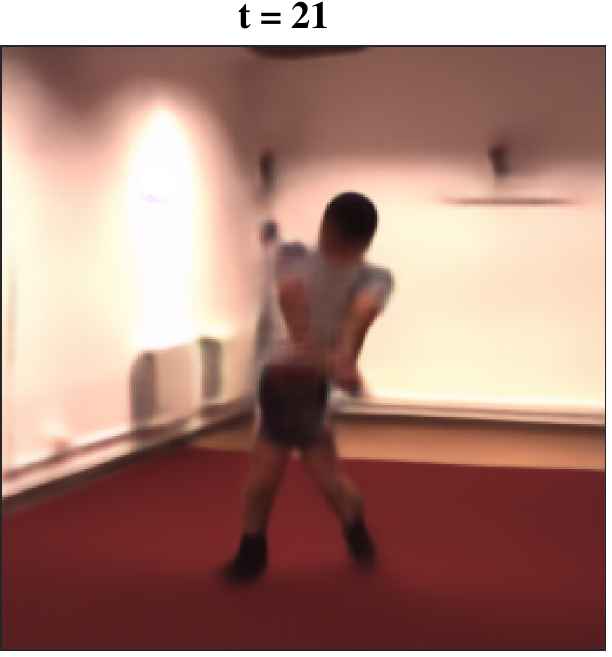}}
\subfloat{\includegraphics[trim={0cm 0cm 0cm 0.4cm},clip,scale=.35]{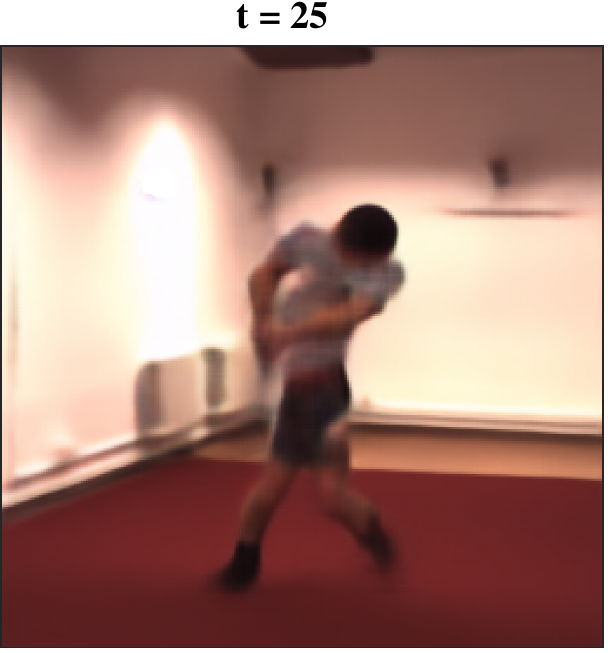}}
\subfloat{\includegraphics[trim={0cm 0cm 0cm 0.4cm},clip,scale=.35]{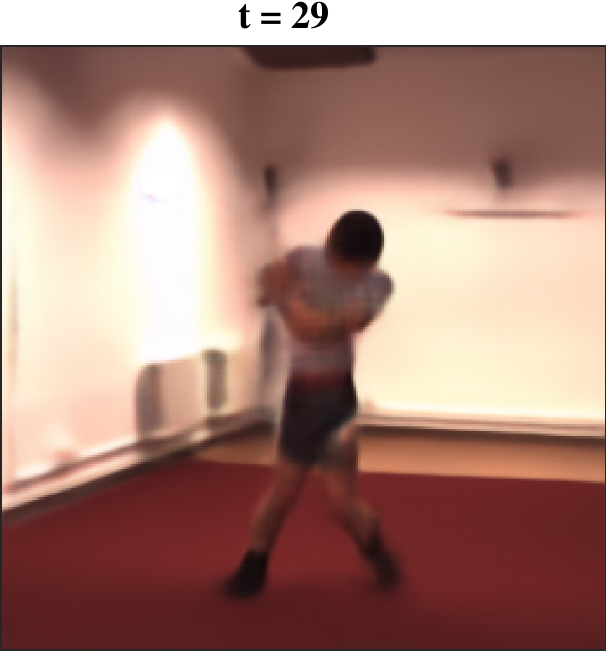}}
}
\vspace{2em}
\end{minipage}

\begin{minipage}{\linewidth}
\subfloat{
\captionsetup{position=top,labelformat=empty}
\subfloat[\tiny\textbf{\textsf{FRAME 1}}]{\includegraphics[trim={0cm 0cm 0cm 0.4cm},clip,scale=.35]{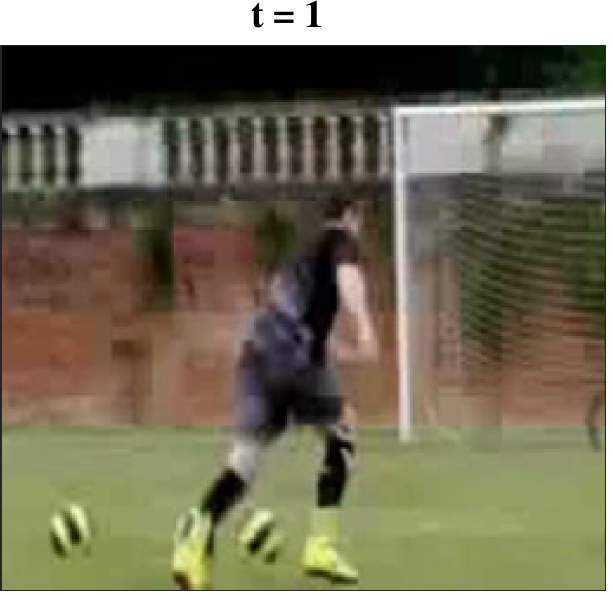}}
\subfloat[\tiny\textbf{\textsf{FRAME 5}}]{\includegraphics[trim={0cm 0cm 0cm 0.4cm},clip,scale=.35]{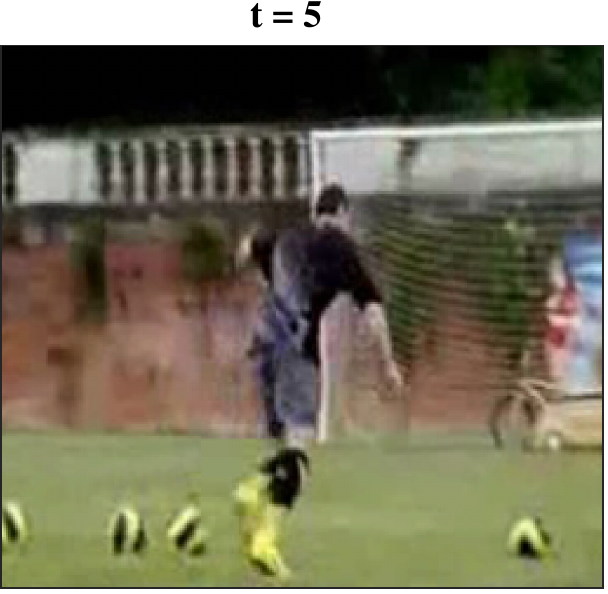}}
\subfloat[\tiny\textbf{\textsf{FRAME 9}}]{\includegraphics[trim={0cm 0cm 0cm 0.4cm},clip,scale=.35]{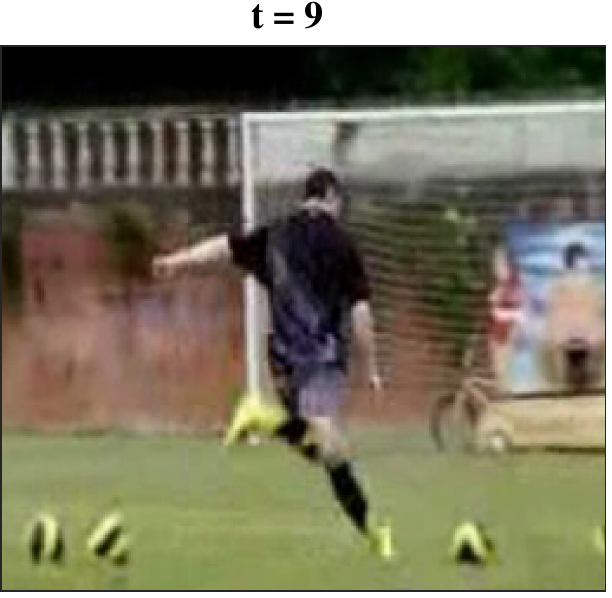}}
\subfloat[\tiny\textbf{\textsf{FRAME 13}}]{\includegraphics[trim={0cm 0cm 0cm 0.4cm},clip,scale=.35]{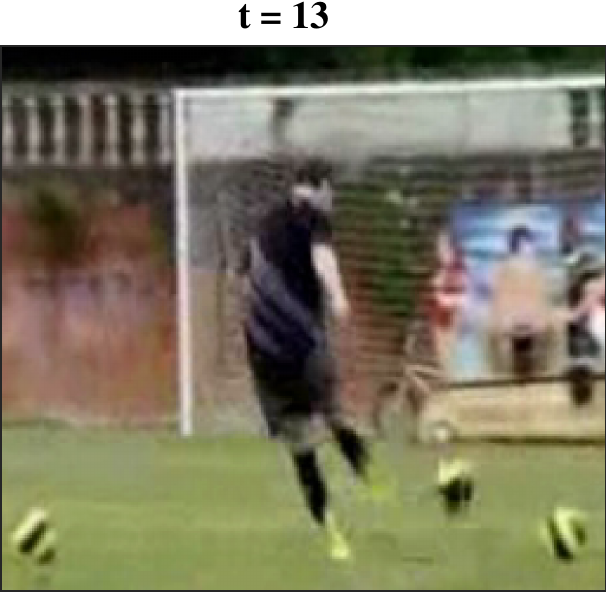}}
\subfloat[\tiny\textbf{\textsf{FRAME 17}}]{\includegraphics[trim={0cm 0cm 0cm 0.4cm},clip,scale=.35]{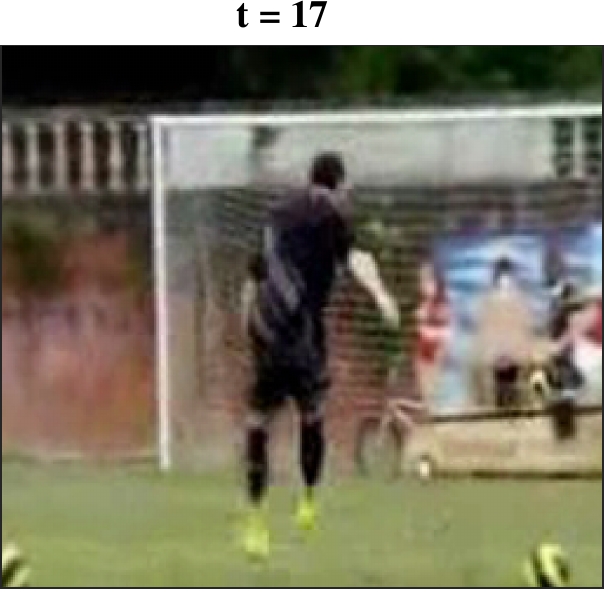}}
\subfloat[\tiny\textbf{\textsf{FRAME 21}}]{\includegraphics[trim={0cm 0cm 0cm 0.4cm},clip,scale=.35]{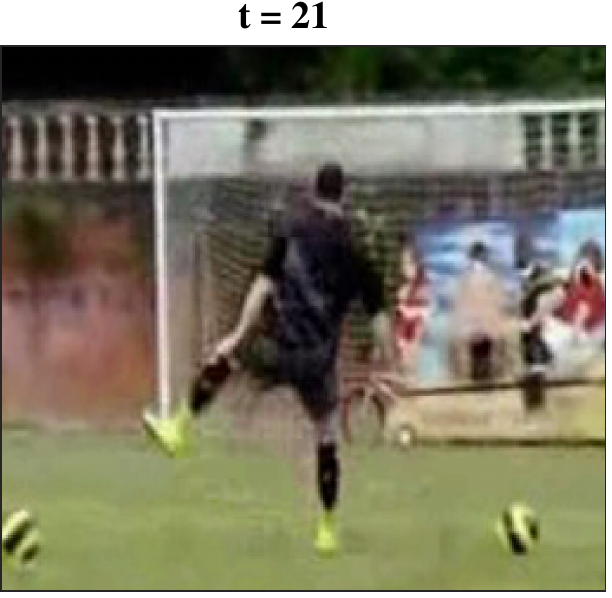}}
\subfloat[\tiny\textbf{\textsf{FRAME 25}}]{\includegraphics[trim={0cm 0cm 0cm 0.4cm},clip,scale=.35]{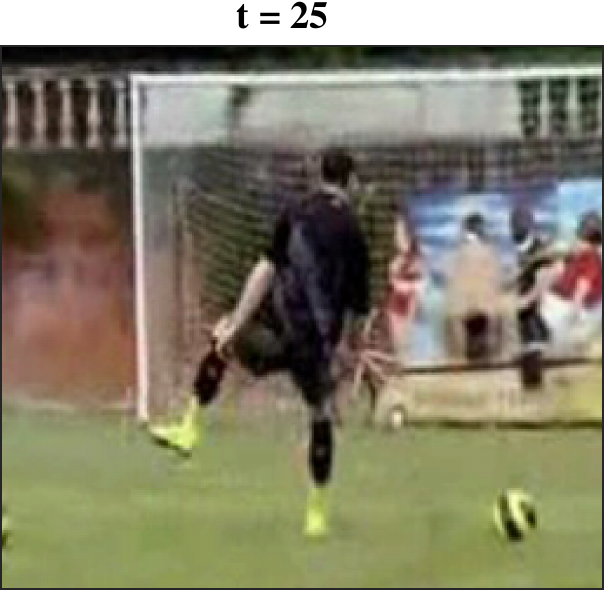}}
\subfloat[\tiny\textbf{\textsf{FRAME 29}}]{\includegraphics[trim={0cm 0cm 0cm 0.4cm},clip,scale=.35]{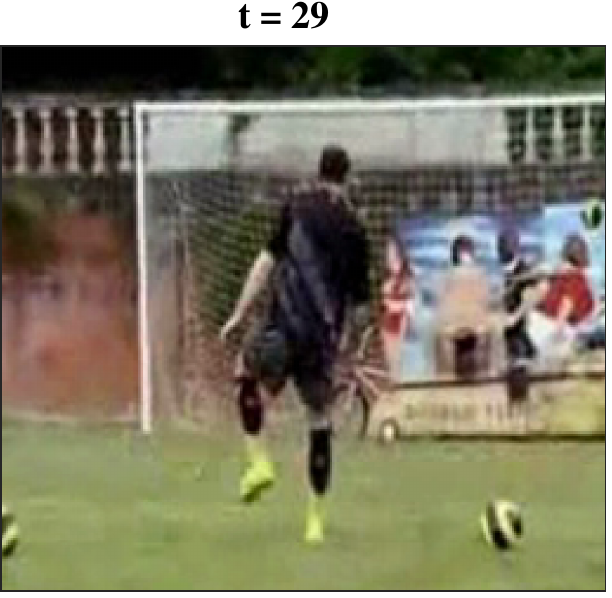}}
}
\vspace{-2em}
\end{minipage}
\renewcommand{\thesubfigure}{b}
\begin{minipage}{\linewidth}
\centering
\subfloat[\label{fig:kick}]{
\subfloat{\includegraphics[trim={0cm 0cm 0cm 0.4cm},clip,scale=.35]{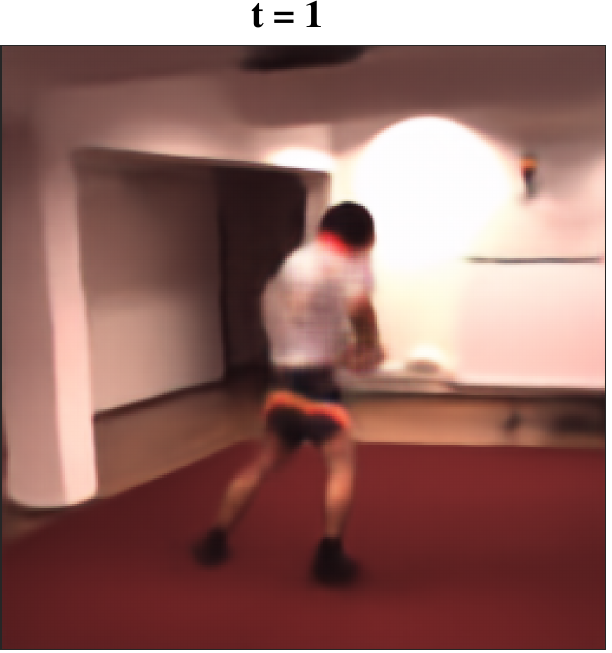}}
\subfloat{\includegraphics[trim={0cm 0cm 0cm 0.4cm},clip,scale=.35]{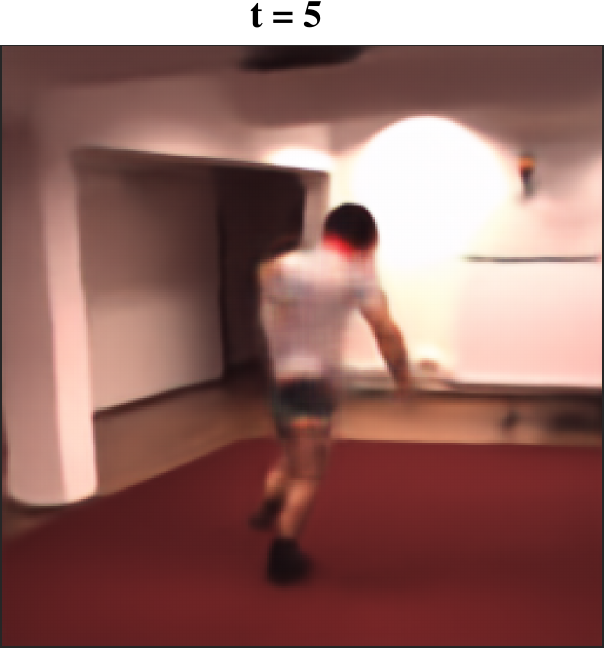}}
\subfloat{\includegraphics[trim={0cm 0cm 0cm 0.4cm},clip,scale=.35]{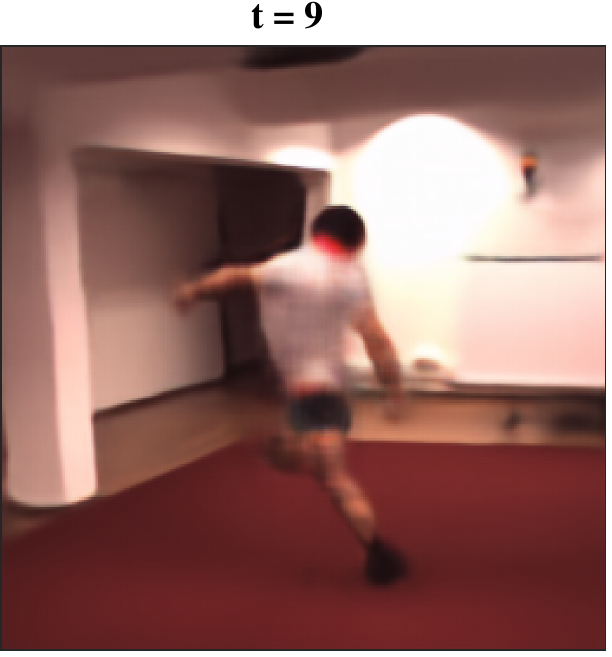}}
\subfloat{\includegraphics[trim={0cm 0cm 0cm 0.4cm},clip,scale=.35]{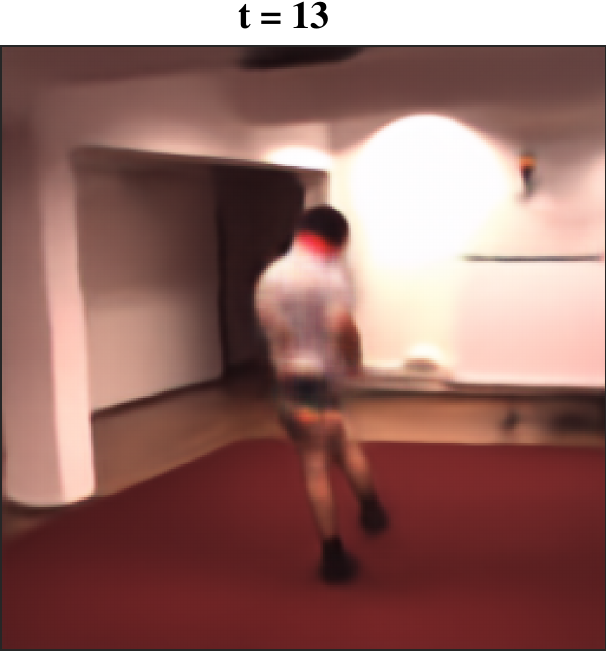}}
\subfloat{\includegraphics[trim={0cm 0cm 0cm 0.4cm},clip,scale=.35]{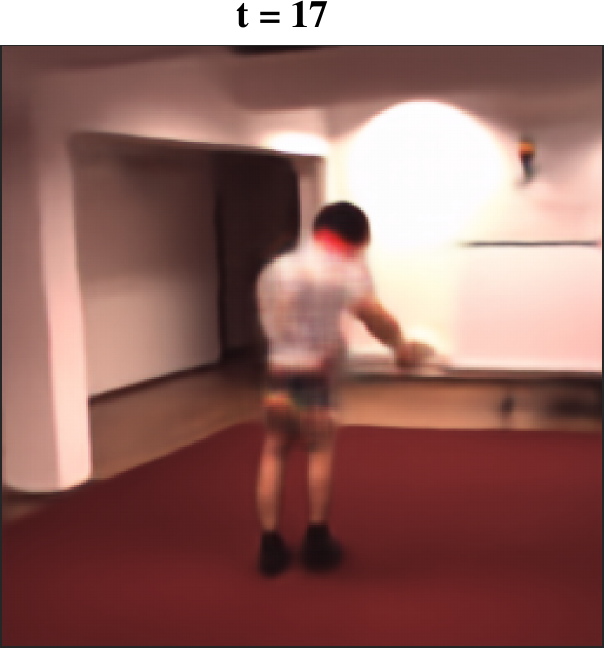}}
\subfloat{\includegraphics[trim={0cm 0cm 0cm 0.4cm},clip,scale=.35]{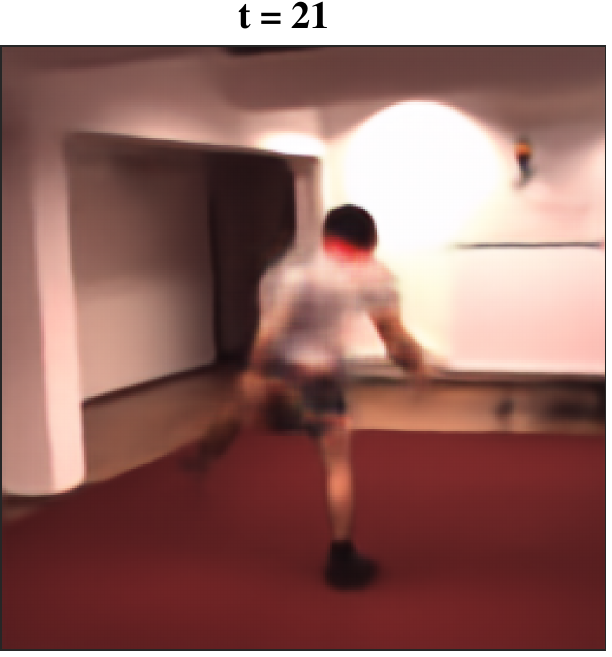}}
\subfloat{\includegraphics[trim={0cm 0cm 0cm 0.4cm},clip,scale=.35]{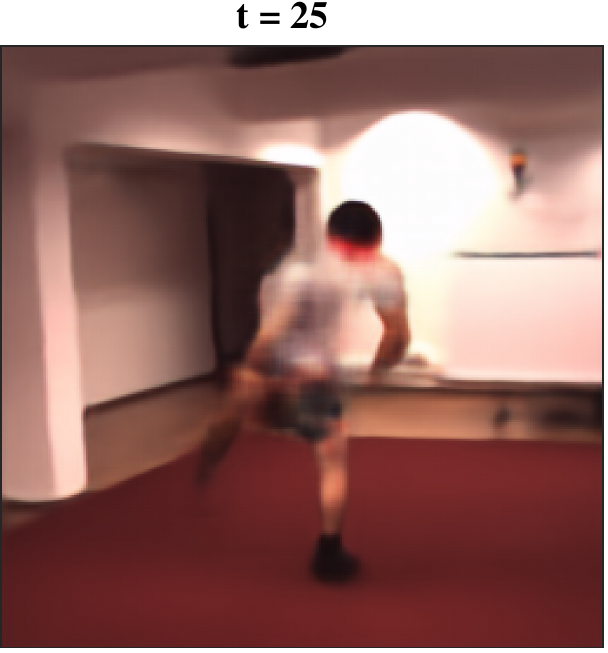}}
\subfloat{\includegraphics[trim={0cm 0cm 0cm 0.4cm},clip,scale=.35]{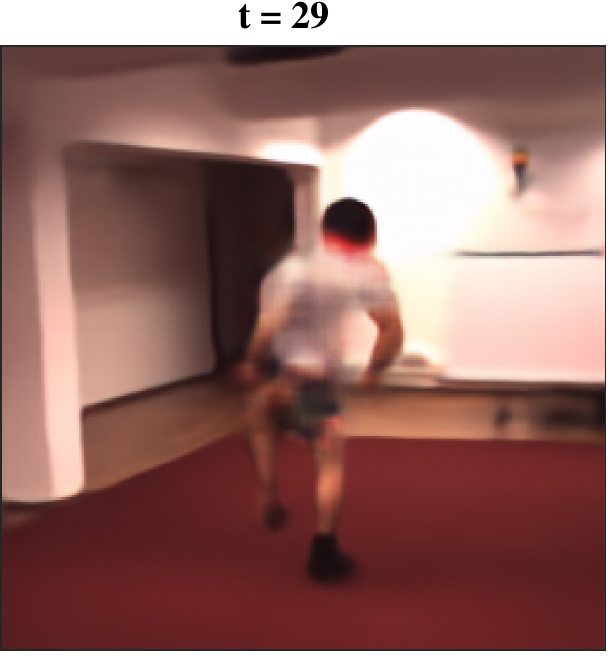}}
}
\end{minipage}
\caption{
Cross-domain pose-transfer over single images from short video sequences from the JHMDB dataset~\cite{Jhuang2013}.
(a) A sequence of frames shows a boy batting a ball while playing baseball (top row) and the correspondent pose-transfer outputs (bottom row).
Mainly due to self-occlusion, some limbs appear blended.
(b) A football player is kicking a ball towards the goal (top row) and the correspondent pose-transfer outputs (bottom row).
Frames 5, 9, and 25 present important issues due to particular postures and self-occlusion of limbs.
Best viewed if zoomed in digital version.}
\label{fig:limitations}
\vspace{-1em}
\end{figure*}

In Fig.~\ref{fig:baseball}, a sequence of frames shows a boy batting a ball while playing baseball (top row) and the correspondent pose-transfer outputs (bottom row).
Although the reenacted frames present the gist of the original sequence, already it is possible to observe that overlapping arms and legs appear to be blended in some of the output images (e.g. frames 1 and 5), making evident the problem we have mentioned earlier.
Fig.~\ref{fig:kick} (top row) depicts a football player kicking a ball towards the goal.
We call attention for frame 5, in which the self-occluded arm of the original subject turns the upper body of the reconstructed person wider.
In frame 9, the concurrent overlapping legs and the unusual pose contribute for an ambiguous posture of the person in the output image, which might be facing forwards or backwards.
The particular body pose in frame 25 provokes the misalignment of head, torso and arms in of the body in the output.
Finally, even without a task-specific training, we believe that the use of a 3D body representation, which would explicitly encode depth, may be beneficial to mitigate the main issues mentioned above.

\section{Conclusions \label{sec:conclusions}}

In this paper, we have presented a comprehensive deep generative model framework for human pose analysis in images.
Our models are based on a principled VAEGAN approach and allow the disentanglement of body posture and visual appearance, aiming for the independent manipulation of such factors.
With our conditional-VAEGAN model, the Conditional-DGPose, differently from previous art, we have taken such manipulation to extreme cases, \eg by performing cross-domain \textit{pose-transfer} and by hallucinating multiple people in a variety of unseen or even unrealistic poses.
Moreover, we have achieved state-of-the-art results on image reconstruction conditioned on pose, \textit{outperforming} the closest related comparable baseline~\cite{LassnerPG17}.
With a single-stage structured semi-supervised VAEGAN architecture, the Semi-DGPose, we pursued the joint \emph{understanding} and \emph{generation} of people in images, not only mapping images to partially interpretable latent representations but also mapping these representations back to the image space.
Importantly, such an approach simultaneously allows for reconstruction, direct manipulation, sampling, pose estimation, indirect pose-transfer, and semi-supervised learning.
These joint capabilities differentiate the Semi-DGPose from other methods in the literature and demonstrate a real-world application of structured deep generative models with the highly relevant potential of being less dependable of fully-labelled data.
We have systematically evaluated our methods on well-known benchmarks, the Human3.6M, the ChictopiaPlus, and the DeepFashion datasets, comparing our results with the closest related baseline methods in the literature~\cite{LassnerPG17,ma2017}.
Such results and comparisons highlight the novelty and effectiveness of our approaches and its capabilities, despite the significant challenge posed by our aimed goal.
We believe that we have shown and reinforced the relevance of employing an interpretable and structured latent space, which allows for semi-supervised learning, as well as the importance of tackling the problem with single-stage end-to-end architectures.


\begin{acknowledgements}
This work was supported by the ERC grant ERC-2012-AdG 321162-HELIOS, EPSRC
grant Seebibyte EP/M013774/1 and EPSRC/MURI grant EP/N019474/1. We would also
like to acknowledge the Royal Academy of Engineering and FiveAI.
Rodrigo de Bem is a CAPES Foundation scholarship holder (Process no: 99999.013296/2013-02, Ministry of Education, Brazil).
\end{acknowledgements}

\bibliographystyle{spmpsci}      
\bibliography{ijcv_dgpose}   

\newpage

\appendix
\setcounter{table}{0}
\renewcommand{\thetable}{A\arabic{table}}
\section{DGPose Architectures Details\label{sec:dgpose_arch_details}}
Here, we provide implementation details of our both architectures considering the following inputs:
images $\bfx$ (batch\_size=64, channels=3, height=64, width=64) and heatmaps $\poseyh$ (batch\_size=64, channels=24, height=64, width=64).
Regarding the heatmap labels, the channels correspond to: \textbf{i)} 14 joints (head top, neck, right shoulder, right elbow, right wrist, right hip, right knee, right ankle,
left shoulder, left elbow, left wrist, left hip, left knee, left ankle); \textbf{ii)} 9 rigid parts (head, right upper arm, right lower arm, right upper leg, right lower leg, left upper arm, left lower arm, left upper leg,
left lower leg); \textbf{iii)} 1 whole body.
Finally, in Tabs.~\ref{table:dgpose_arch} and~\ref{table:pose_model_semi_sup_arch}, we show the full definition of both, the Conditional-DGPose and the Semi-DGPose, respectively.

\begin{table}[ht]
{\scriptsize
\centering
\begin{tabulary}{\textwidth}{ll}
\hline
\multicolumn{2}{c}{\textbf{RESIDUAL Layer}}\\
\hline
\multicolumn{2}{l}{\textbf{Input:} \textit{previous\_layer\_output}}\\
\hline
\textbf{Layer} & \textbf{Definition}\\
\hline
1 & CONV-(N512, K3, S1, P1), BN, ReLU\\
2 & CONV-(N512, K3, S2, P1), BN\\
3 & SUM(CONV-2, \textit{previous\_layer\_output})\\
\hline
\end{tabulary}
\caption{\footnotesize Architecture of the residual block employed in the DGPose encoder.}
\label{table:residual_arch}
}
\end{table}

\begin{table}[ht]
{\scriptsize
\centering
\begin{tabularx}{\linewidth}{@{}ll@{}}
\hline
\multicolumn{2}{c}{\textbf{Encoder}}\\
\hline
\multicolumn{2}{l}{\textbf{Input:} $\bfx$(batch\_size=64, channels=3, height=64, width=64);}\\
\multicolumn{2}{l}{$\poseyh$(batch\_size=64, channels=24, height=64, width=64)}\\
\hline
\textbf{Layer} & \textbf{Definition}\\
\hline
1 & CONCAT($\bfx$, $\poseyh$)\\
2 & CONV-(N64, K7, S2, P1), LeakyReLU(0.01)\\
3 & CONV-(N128, K3, S2, P1), BN, ReLU\\
4 & CONV-(N256, K3, S2, P1), BN, ReLU\\
5 & CONV-(N512, K3, S2, P1), BN, ReLU\\
6 & CONV-(N512, K3, S2, P1), BN, ReLU\\
7 & CONV-(N512, K3, S2, P1), BN, ReLU\\
8 & RESIDUAL-(N512, K3, S1, P1)\\
9 & RESIDUAL-(N512, K3, S1, P1)\\
10 & RESIDUAL-(N512, K3, S1, P1)\\
11 & RESIDUAL-(N512, K3, S1, P1), SIGMOID\\
$\mu_{\mathbf{z}}$ & FC-(N100)\\
$\sigma_{\mathbf{z}}$ & FC-(N100)\\
\hline
\multicolumn{2}{c}{\textbf{Prior}}\\
\hline
\multicolumn{2}{l}{\textbf{Input:} $\poseyh$(batch\_size=64, channels=24, height=64, width=64)}\\
\hline
\textbf{Layer} & \textbf{Definition}\\
\hline
1 & CONV-(N128, K4, S2, P1), LeakyReLU(0.2)\\
2 & CONV-(N256, K4, S2, P1), BN, LeakyReLU(0.2)\\
3 & CONV-(N512, K4, S2, P1), BN, LeakyReLU(0.2)\\
4 & CONV-(N1024, K4, S2, P1), BN, LeakyReLU(0.2)\\
5 & CONV-(N100, K4, S1, P0), SIGMOID\\
$\mu_{prior}$ & FC-(N100)\\
$\sigma_{prior}$ & FC-(N100)\\
\hline
\multicolumn{2}{c}{\textbf{Decoder}}\\
\hline
\multicolumn{2}{l}{\textbf{Input:} $\mathbf{z}$(batch\_size=64, channels=100)}\\
\hline
\textbf{Layer} & \textbf{Definition}  \\
\hline
1 & RESHAPE\tiny{(batch\_size=64, channels=100, height=1, width=1)}\\
2 & DECONV-(N512, K4, S1, P0), BN, LeakyReLU(0.2)\\
3 & DECONV-(N256, K4, S2, P1), BN, LeakyReLU(0.2)\\
4 & DECONV-(N128, K4, S2, P1), BN, LeakyReLU(0.2)\\
5 & DECONV-(N64, K4, S2, P1), BN, LeakyReLU(0.2)\\
6 & DECONV-(N128, K4, S2, P1), BN, LeakyReLU(0.2)\\
7 & CONCAT(DECONV-6, $\poseyh$)\\
8 & CONV-(N512, K5, S1, P2), BN, LeakyReLU(0.2)\\
9 & CONV-(N256, K5, S1, P2), BN, LeakyReLU(0.2)\\
10 & CONV-(N128, K5, S1, P2), BN, LeakyReLU(0.2)\\
11 & CONV-(N128, K5, S1, P2), BN, LeakyReLU(0.2)\\
$\text{G}(\mathbf{y}_{\negthinspace h},\mathbf{z})$ & CONV-(N3, K5, S1, P2), TANH\\
\hline
\multicolumn{2}{c}{\textbf{Discriminator}}\\
\hline
\multicolumn{2}{l}{\textbf{Input:} $\text{G}(\mathbf{y}_{\negthinspace h},\mathbf{z})$(batch\_size=64, channels=3, height=64, width=64);}\\
\multicolumn{2}{l}{$\bfx$(batch\_size=64, channels=3, height=64, width=64)}\\
\hline
\textbf{Layer} & \textbf{Definition}  \\
\hline
1 & CONV-(N64, K4, S2, P1), LeakyReLU(0.2)\\
2 & CONV-(N128, K4, S2, P1), BN, LeakyReLU(0.2)\\
3 & CONV-(N256, K4, S2, P1), BN, LeakyReLU(0.2)\\
4 & CONV-(N512, K4, S2, P1), BN, LeakyReLU(0.2)\\
5 & CONV-(N1, K4, S1, P0), SIGMOID\\
\hline
\end{tabularx}
\caption{\footnotesize Conditional-DGPose architecture for $64 \times 64$ input images.
We use the following abbreviations: N for the number of kernels/neurons, K for kernel size, S for stride and P for zero padding.
Concerning the layers, CONCAT means concatenation layer, CONV means convolutional layer, BN means batch normalization layer with running average coefficient $\beta=0.9$ and learnable affine transformation,
DECONV means transpose convolutional layer, FC means fully connected layer, SUM corresponds to element-wise sum layer and RESIDUAL denotes a residual block, detailed at Table~\ref{table:residual_arch}.
The additional layers can be clearly understood.
Finally, particular parameters for specific layers are defined between parenthesis after the layers' names.}
\label{table:dgpose_arch}
}
\end{table}

\begin{table}[ht]
{\scriptsize
\centering
\begin{tabularx}{\linewidth}{@{}ll@{}}
\hline
\multicolumn{2}{c}{\textbf{Encoder}}\\
\hline
\multicolumn{2}{l}{\textbf{Input:} $\bfx$(batch\_size=64, channels=3, height=64, width=64)}\\
\hline
\textbf{Layer} & \textbf{Definition}\\
\hline
1 & CONV-(N64, K7, S2, P1), LeakyReLU(0.01)\\
2 & CONV-(N128, K3, S2, P1), BN, ReLU\\
3 & CONV-(N256, K3, S2, P1), BN, ReLU\\
4 & CONV-(N512, K3, S2, P1), BN, ReLU\\
5 & CONV-(N512, K3, S2, P1), BN, ReLU\\
6 & CONV-(N512, K3, S2, P1), BN, ReLU\\
7 & RESIDUAL-(N512, K3, S1, P1)\\
8 & RESIDUAL-(N512, K3, S1, P1)\\
9 & RESIDUAL-(N512, K3, S1, P1)\\
10 & RESIDUAL-(N512, K3, S1, P1), SIGMOID\\
$\mu_{\mathbf{z}}$ & FC-(N100)\\
$\sigma_{\mathbf{z}}$ & FC-(N100)\\
$\mu_{\mathbf{y}_{\negthinspace v}}$ & FC-(N48)\\
$\sigma_{\mathbf{y}_{\negthinspace v}}$ & FC-(N48)\\
\hline
\multicolumn{2}{c}{\textbf{Mapper}}\\
\hline
\multicolumn{2}{l}{\textbf{Input:} $\poseyv$(batch\_size=64, channels=48)}\\
\hline
\textbf{Layer} & \textbf{Definition}\\
\hline
1 & RESHAPE\tiny{(batch\_size=64, channels=48, height=1, width=1)}\\
2 & DECONV-(N512, K4, S1, P0), BN, LeakyReLU(0.2)\\
3 & DECONV-(N256, K4, S2, P1), BN, LeakyReLU(0.2)\\
4 & DECONV-(N128, K4, S2, P1), BN, LeakyReLU(0.2)\\
5 & DECONV-(N64, K4, S2, P1), BN, LeakyReLU(0.2)\\
$\poseyh$ & DECONV-(N24, K4, S2, P1), SIGMOID\\
\hline
\multicolumn{2}{c}{\textbf{Decoder}}\\
\hline
\multicolumn{2}{l}{\textbf{Input:} $\mathbf{z}$(batch\_size=64, channels=100);}\\
\multicolumn{2}{l}{$\poseyv$(batch\_size=64, channels=48);}\\
\multicolumn{2}{l}{$\poseyh$(batch\_size=64, channels=24, height=64, width=64)}\\
\hline
\textbf{Layer} & \textbf{Definition}  \\
\hline
1 & CONCAT($\mathbf{z}$, $\poseyv$)\\
2 & RESHAPE\tiny{(batch\_size=64, channels=148, height=1, width=1)}\\
3 & DECONV-(N512, K4, S1, P0), BN, LeakyReLU(0.2)\\
4 & DECONV-(N256, K4, S2, P1), BN, LeakyReLU(0.2)\\
5 & DECONV-(N128, K4, S2, P1), BN, LeakyReLU(0.2)\\
6 & DECONV-(N64, K4, S2, P1), BN, LeakyReLU(0.2)\\
7 & DECONV-(N128, K4, S2, P1), BN, LeakyReLU(0.2)\\
8 & CONCAT(DECONV-6, $\poseyh$)\\
9 & CONV-(N512, K5, S1, P2), BN, LeakyReLU(0.2)\\
10 & CONV-(N256, K5, S1, P2), BN, LeakyReLU(0.2)\\
11 & CONV-(N128, K5, S1, P2), BN, LeakyReLU(0.2)\\
12 & CONV-(N128, K5, S1, P2), BN, LeakyReLU(0.2)\\
$\text{G}(\mathbf{y}_{\negthinspace v},\mathbf{z})$ & CONV-(N3, K5, S1, P2), TANH\\
\hline
\multicolumn{2}{c}{\textbf{Discriminator}}\\
\hline
\multicolumn{2}{l}{\textbf{Input:} $\text{G}(\mathbf{y}_{\negthinspace v},\mathbf{z})$(batch\_size=64, channels=3, height=64, width=64);}\\
\multicolumn{2}{l}{$\mathbf{x}$(batch\_size=64, channels=3, height=64, width=64)}\\
\hline
\textbf{Layer} & \textbf{Definition}  \\
\hline
1 & CONV-(N64, K4, S2, P1), LeakyReLU(0.2)\\
2 & CONV-(N128, K4, S2, P1), BN, LeakyReLU(0.2)\\
3 & CONV-(N256, K4, S2, P1), BN, LeakyReLU(0.2)\\
4 & CONV-(N512, K4, S2, P1), BN, LeakyReLU(0.2)\\
5 & CONV-(N1, K4, S1, P0), SIGMOID\\
\hline
\end{tabularx}
\caption{\footnotesize Semi-DGPose architecture for $64 \times 64$ input images.
We use the following abbreviations: N for the number of kernels/neurons, K for kernel size, S for stride and P for zero padding.
Concerning the layers, CONCAT means concatenation layer, CONV means convolutional layer, BN means batch normalization layer with running average coefficient $\beta=0.9$ and learnable affine transformation,
DECONV means transpose convolutional layer, FC means fully connected layer, SUM corresponds to element-wise sum layer and RESIDUAL denotes a residual block, detailed at Table~\ref{table:residual_arch}.
The additional layers can be clearly understood.
Finally, particular parameters for specific layers are defined between parenthesis after the layers' names.}
\label{table:pose_model_semi_sup_arch}
}
\end{table}


\end{document}